\pdfoutput=1
\documentclass[12pt]{report}
\usepackage[numbers]{natbib}

\usepackage{amsmath,amsfonts,amssymb,amsthm, color}
\usepackage{algorithm,algorithmicx}
\usepackage{epsfig,wrapfig,epsf,subfig}
\usepackage[outercaption]{sidecap} 
\usepackage{times}
\usepackage{fancyvrb}
\usepackage{enumerate}
\usepackage{authblk}
\usepackage{tikz}
\usepackage{bm}
\usetikzlibrary{arrows,shapes}
\usetikzlibrary{patterns}
\usepackage{graphicx}
\usepackage{fullpage}
\usepackage{ltxtable}
\usepackage{nicefrac}
\usepackage{multirow}
\usepackage{setspace}
\usepackage[colorlinks=true, linkcolor=black, bookmarks=true]{hyperref}
\usepackage[noend]{algpseudocode}
\usepackage[utf8]{inputenc} 
\usepackage[T1]{fontenc}    
\usepackage{url}            
\usepackage{booktabs}       
\usepackage{microtype}      
\usepackage{pbox}
\usepackage{natbib}
\renewcommand{\ref}{\hyperref}

\makeatletter
\newtheorem*{rep@theorem}{\rep@title}
\newcommand{\newreptheorem}[2]{%
\newenvironment{rep#1}[1]{%
 \def\rep@title{#2 \ref{##1}}%
 \begin{rep@theorem}}%
 {\end{rep@theorem}}}
\makeatother

\newreptheorem{theorem}{Theorem}
\newreptheorem{lemma}{Lemma}
\newreptheorem{definition}{Definition}

\theoremstyle{definition}

\newcounter{saveenumi}

\algrenewcommand\algorithmicrequire{\textbf{Parameters:}}
\algrenewcommand\algorithmicensure{\textbf{Initialization:}}
\newcommand{\etal}{{\it et al}}

\usepackage[left=1.65in, right=1.65in, top=1in, bottom=1in]{geometry}
\usepackage{makeidx}

\makeindex

\usepackage{times}
\usepackage{epsfig}
\usepackage{graphicx}

\usepackage{bm}
\usepackage{makecell}
\usepackage{multirow}

\usepackage{color}
\usepackage{rotating}
\usepackage{overpic}
\usepackage{xcolor}
\usepackage{colortbl}
\usepackage{hhline}

\usepackage{booktabs}

\usepackage{caption}

\usepackage{amsmath,amssymb,amsfonts,gensymb}

\usepackage{tikz}
\usetikzlibrary{decorations.pathmorphing}
\usetikzlibrary{intersections}
\usetikzlibrary{positioning}
\usetikzlibrary{calc}
\usetikzlibrary{fit}
\usepackage[skins]{tcolorbox}

\usepackage{geometry}

\usepackage{adjustbox}
\usepackage{array}
\def\rot{\rotatebox}
\usepackage{tabu}

\usepackage{upgreek}
\usepackage{bm}
\usepackage{comment}

\usepackage{algorithm}
\usepackage{subfig}

\makeatletter
\newcommand{\bfgreek}[1]{\bm{\@nameuse{#1}}}
\newcommand{\bfgreekco}[2]{\bm{\textcolor{#2}{\@nameuse{#1}}}}
\makeatother

\def\Hline{\Xhline{2\arrayrulewidth}}     

\definecolor{mediumgreen}{RGB}{13,204,128}

\newcommand{\eg}[1]{e.g.}
\newcommand{\val}[1]{\texttt{val}}
\newcommand{\test}[1]{\texttt{test}}
\newcommand{\train}[1]{\texttt{train}}
\newcommand{\aug}[1]{\texttt{train-aug}}

\DeclareMathOperator*{\argmax}{argmax}
\DeclareMathOperator*{\argmin}{argmin}

\numberwithin{equation}{section}

   \def\url#1{}

\newcommand{\thesisTitle}{LEARNING RICH REPRESENTATIONS FOR STRUCTURED VISUAL PREDICTION TASKS}
\newcommand{\thesisAuthor}{Mohammadreza Mostajabi}
\newcommand{\thesisInstitute}{Toyota Technological Institute at Chicago}
\newcommand{\thesisDate}{June, 2019}
\newcommand{\thesisAdvisor}{Dr. Gregory Shakhnarovich}
\newcommand{\committeeFirst}{Dr. Karen Livescu}
\newcommand{\committeeSecond}{Dr. Michael Maire}

\newcommand\numberthis{\addtocounter{equation}{1}\tag{\theequation}}
\newcommand{\thesisAbstract}{
We describe an approach to learning rich representations for images, that enables simple and effective predictors in a range of vision tasks involving spatially structured maps. Examples of tasks where one can leverage our approach include semantic segmentation and depth estimation. Our key idea is to map small image elements (pixels or superpixels) to feature representations extracted from a sequence of nested regions of increasing spatial extent. These regions are obtained by "zooming out" from the pixel/superpixel all the way to scene-level resolution, and hence we call these zoom-out features. Applied to semantic segmentation and other structured prediction tasks, our approach exploits statistical structure in the image and in the label space without setting up explicit structured prediction mechanisms, and thus avoids complex and expensive inference. Instead image elements are classified by a feedforward multilayer network with skip-layer connections spanning the zoom-out levels.

We describe extensive experiments showing the effectiveness of our simple architecture design. When used in conjunction with modern neural architectures such as ResNet,  DenseNet and NASNet (to which it is complementary) our approach achieves competitive accuracy on segmentation benchmarks. In addition, we propose an approach for learning category-level semantic segmentation
purely from image-level classification tags indicating presence of categories. It exploits
localization cues that emerge from training a modified zoom-out architecture tailored for
classification tasks, to drive a weakly supervised process that automatically labels a sparse,
diverse training set of points likely to belong to classes of interest. 
Finally, we introduce data-driven regularization functions for the supervised training of CNNs. Our innovation takes the form of a regularizer derived by learning an autoencoder over the set of annotations. This approach further complements our zoom-out representation, leveraging an improved representation of label space to inform our extraction of features from images.
}

\usepackage{float}

\begin{document}

\newgeometry{left=1.65in, right=1.65in, top=1.65in, bottom=1.65in}

\begin{center}

\Large{\bf \thesisTitle} \\[1cm]
\large{by} \\[1cm]
\large{\thesisAuthor} \\[2cm]
\large{A thesis submitted\\ in partial fulfillment of the requirements for\\the degree of}\\[0.8cm]
\large{Doctor of Philosophy in Computer Science}\\[0.8cm]
\large{at the}\\[0.8cm]
\large{\MakeUppercase \thesisInstitute}\\[0.8cm]
\large{\thesisDate}\\[2cm]
\large{Thesis Committee: \\ {\thesisAdvisor \;(Thesis advisor)},\\ {\committeeFirst},\\ {\committeeSecond}}	

\end{center}

\thispagestyle{empty}
\newgeometry{left=0in, right=0in, top=0in, bottom=0in}

%
%
%
%
%
%

\begin{figure}
 \centering 
 \includegraphics{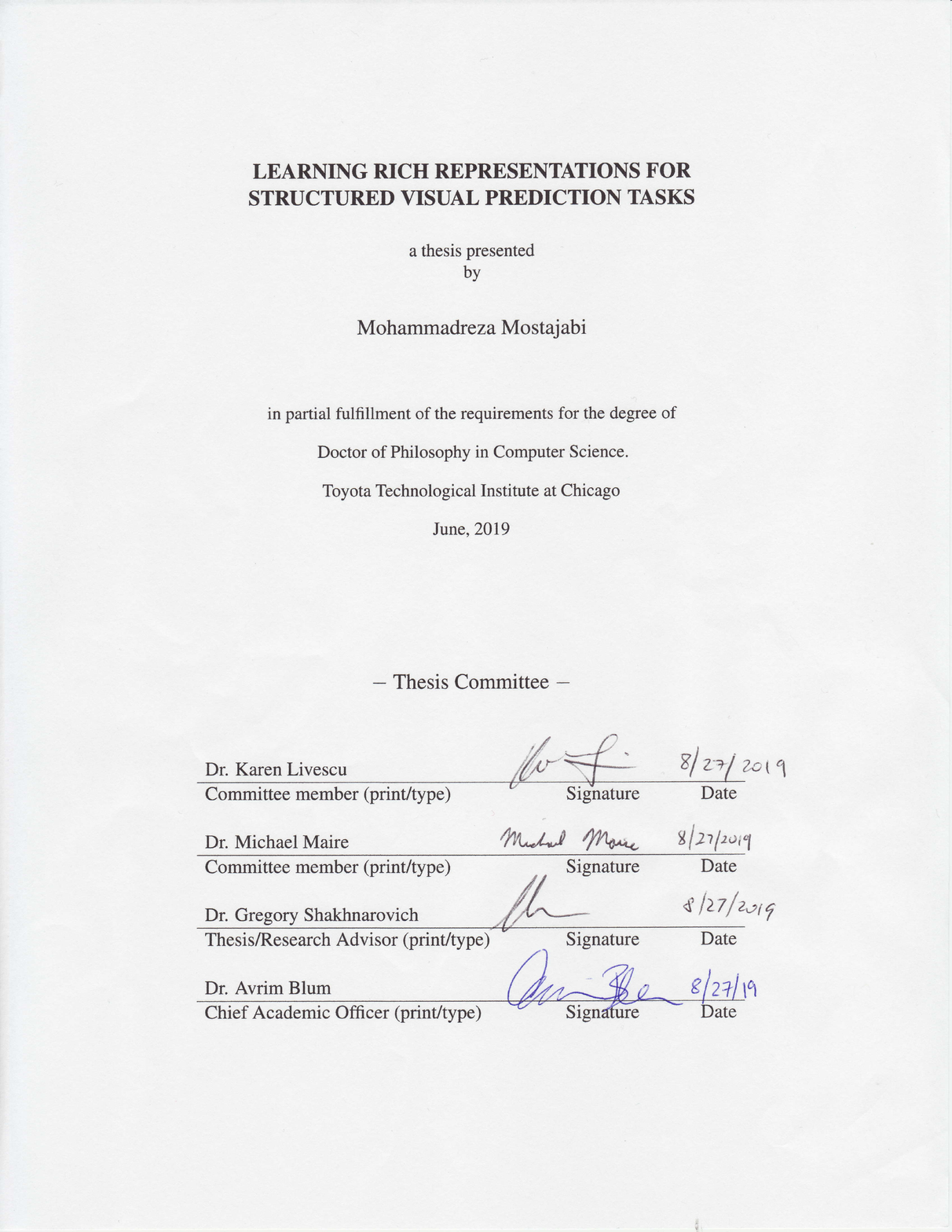}
\end{figure}

\newgeometry{left=1.2in, right=1.2in, top=1in, bottom=1.25in}

\begin{center}

\Large{\bf \thesisTitle} \\[0.2cm]
\large{by} \\[0.2cm]
\Large{\thesisAuthor}

\end{center}

\section*{Abstract}
\thesisAbstract\par

\noindent  Thesis Advisor: Gregory Shakhnarovich

\thispagestyle{empty}
\newpage

\section*{Acknowledgments}

I would like to thank my research advisor Greg Shakhnarovich. He has been a great supporter during my whole PhD. I had the opportunity to work on what I was interested in and I learned a lot from his deep insight in computer vision and machine learning. I would also like to thank Michael Maire for his research collaboration which was a new chapter in my PhD. I benefited from his novel perspective in the field and productive discussions with him. A big thanks to Karen Livescu for the insightful discussions that we had and her precise and constructive comments on my thesis.

I would also thank the faculty and staff at TTIC for making TTIC a remarkable research institute. I would like to especially thank David McAllester, Madhur Tulsiani, Matthew Walter, 	
Julia Chuzhoy, Chrissy Novak, Adam Bohlander and Mary Marre.

My special thanks goes to my collaborators and fellow students at Toyota Technological Institute at
Chicago and University of Chicago, for the fruitful discussions, their supports and all the enjoyable moments we had together. I’d
like to especially thank Steven Basart, Igor Vasiljevic, Qingming Tang, Nicholas Kolkin, Lifu Tu, Hai Wang, Somaye Hashemifar, Sudarshan Babu, Ankita Pasad, Shane Settle, Andrea Daniele, Behnam Tavakoli, Shubhendu Trivedi, Haris Angelidakis, 	
Mrinalkanti Ghosh,	
Takeshi Onishi, Payman Yadollahpour, Ruotian Luo, Falcon Dai and Mantas Mazeika.

My deepest gratitude goes to my family for their endless love and support during all these years. I'd like to thank my parents for supporting my decisions throughout all these years. I owe my success to them. I'd like to thank Elahe for her warm support. Words cannot express how grateful I am to have you by my side.

\newgeometry{left=1.2in, right=1.2in, top=1in, bottom=1.25in}

{ \tableofcontents }
{\listoffigures}
{ \listoftables}
%

\doublespacing 
\chapter{Introduction}  \label{chap:mainintro}

Structured prediction in which ground-truth labels themselves exhibit complex internal structure encompasses many of the most important tasks in computer vision. Semantic image segmentation and depth estimation are excellent examples of such tasks. Semantic image segmentation is of particular interest because it requires modeling everything from low-level local patterns such as texture features to complex interactions between neighboring and distant image elements. It is important because it entails aspects of scene understanding such as image classification and object localization. Furthermore, it has important applications ranging from medical image diagnosis to guiding robots' interaction with the world.

Learning rich data representations is an essential block of modern computer vision. In this thesis, we focus on techniques for learning rich representations for structured prediction tasks. In the following it will be described what we mean by rich representations in the context of semantic segmentation and depth estimation. Consider semantic segmentation in which the goal is to label every pixel in the image with a category label. There are many types of information and clues that can be leveraged. Local features play important roles in segmenting and recognizing fine details and object parts. As an example, the texture of a wheel's spoke is a rich local feature which can be used to segment and recognize a bicycle wheel. However, in many cases, objects are highly occluded and only small parts of the objects are visible. As such, exploiting only local features will not be sufficient to correctly identify partially visible objects. In these cases, one can exploit semantic relations and contextual information such as object co-occurrence as a prior on scene composition. A rich representation for this task is able to encode both local level features and high level contextual information.

Depth estimation is another challenging structured prediction task. The goal in monocular depth estimation is to predict the depth value of each pixel, given a single RGB image as input. Consider an image of an outdoor scene containing grass. The texture of grass becomes less and less apparent as we go farther into the distance. As in semantic segmentation, local texture features are informative. Object occlusion is another source of information. A table is closer to the camera than a chair if the table occludes the chair. Again, a rich representation for monocular depth estimation is a representation which is able to encode both local level features and high level scene information such as object occlusion.

 In natural images, each image element has some connection to other image elements. As an example, consider a pixel that is part of an image of the human body. It is very likely that neighboring pixels are part of the same body due to the continuity of real world objects. Objects also have semantic co-occurence relationships. Observing the ocean supports the existence of a boat rather than a car, while observing roads supports the opposite. Despite the attention semantic segmentation has received, it remains challenging, largely due to complex interactions between local, neighboring and distant image elements and the importance of global context. In order to capture existing structure and contextual information in natural scenes, a common approach has been to use a random field or structured support vector machine~\cite{kohli_cvpr08,shotton_ijcv09,koltun2011efficient}. Such models incorporate local evidence in unary potentials, while interactions between label assignments are captured by pairwise and possibly higher-order potentials. This includes various hierarchical CRFs ~\cite{shotton_ijcv09,ladicky_iccv09,lempitsky2011pylon,yadollahpour2013discriminative}. These approaches introduce their own severe challenges, among them the intractable nature of inference and learning in many “interesting” models~\cite{koltun2011efficient}.
 
 Conventional models prior to this thesis used hand-crafted features as image descriptors and forced explicit constraints on label structure~\cite{batra_eccv12,lempitsky2011pylon,boix_ijcv12,mostajabi2014robust}. In this dissertation our goal is to depart from conventional models and focus on modeling label structure and learning rich representations for structured prediction tasks in an end-to-end fashion.

 \section{Contributions}

Our main contributions are summarized as:

\begin{description}

   \item[$\bullet$] \textit{Learning rich representations with a simple architecture.} Using our zoom-out architecture we learn rich representations at different scales, unlike previous state-of-the-art methods which were based on hand-crafted features. Zoom-out exploits statistical and contextual structure in the label space and dependencies between image elements at different scales, without explicitly encoding them in a complex model. This results in a much simpler and faster model at inference time.  This is achieved by extracting and learning features at different levels and scales which gives zoom-out architecture an ability to encode local level features as well as high level contextual information. This is a necessity for most of the structured prediction tasks.  
   
  \item[$\bullet$] \textit{Powerful deep CNN-based architecture for semantic segmentation.} We are the first ones to show that segmentation, just like image classification, detection and other recognition tasks, can benefit from the advent of deep convolutional networks.
  
  \item[$\bullet$] \textit{An end-to-end framework for structured visual prediction tasks.} In Chapter~\ref{chap:introzoom} we show the application of the zoom-out architecture for image segmentation. In Section~\ref{sec:zoomconc} we address some of the follow-up work that utilized zoom-out in different applications. In Chapter~\ref{chap:labelemb} we apply zoom-out to depth estimation.
  
     \item[$\bullet$] \textit{Diverse sampling for weakly supervised segmentation.} Our diverse sampling method for providing point-wise semantic segmentation supervision from image-level classification tags has almost no hyperparameters and significantly improves results over picking random, dense or top $k$ points with highest scores. 
      
  \item[$\bullet$] \textit{Constructing and learning custom regularization functions.} We propose a two-phase training procedure for regularizing deep networks by modeling and predicting label structure. Our experiments, in the context of semantic segmentation
demonstrate that our regularization strategy leads to consistent accuracy boosts over baselines, both when training
from scratch, or in combination with ImageNet pretraining.
         
     \item[$\bullet$] \textit{Unified architecture for predicting segmentation and depth.} We show that it is possible to predict depth directly from semantic segmentation maps. Based on this observation we propose a two-part architecture for joint prediction of depth and semantic
segmentation which uses semantic segmentation as an intermediate representation for depth estimation.

\end{description}

\section{Thesis outline}

In Chapter~\ref{chap:introzoom}, we introduce a purely feed-forward architecture for semantic segmentation. We map pixels/superpixels to rich feature representations extracted from a sequence of nested regions of increasing extent. These regions are obtained by "zooming out" from the pixel/superpixel all the way to scene-level resolution. This approach exploits statistical structure in the image and label space without requiring explicit structured prediction mechanisms, thus avoiding complex and expensive inference. Instead image elements are classified by a feedforward multilayer network. We show that our rich image representation with a simple linear classifier on top, significantly outperforms models based on state-of-the-art hand designed features processed by complex CRF models.

Most recent progress in semantic segmentation can be attributed to advances in deep learning and to availability of large, manually labeled datasets. However, the cost and complexity of annotating segmentation labels are significantly higher than that for
classification. There is mounting evidence that this task, while difficult, is not hopeless. Units sensitive to object localization have been shown to emerge as part of the representations learned by convolutional neural networks (CNNs) trained for image classification. In Chapter~\ref{chap:chapter2} we propose an approach for learning category-level semantic segmentation purely from image-level classification tags indicating presence of categories. In the first chapter we proposed zoom-out architecture to extract rich pixel-level features for the task of semantic segmentation. We can tailor zoom-out architecture for the task of image classification by performing global pooling on top of zoom-out features. This allows us to exploit localization cues that emerge from training classification-tasked
convolutional networks, to drive a weakly supervised process that automatically labels a sparse, diverse training set of points likely to belong to classes of interest. Our approach has almost no hyperparameters, is modular and obtains competitive results on the VOC 2012 segmentation benchmark~\cite{Mostajabidivsample}.

In computer vision, annotation is a sufficiently precious resource that it is commonplace to pretrain systems on millions of labeled ImageNet~\cite{ImageNet} examples.
These systems absorb a useful generic visual representation ability during pretraining, before being fine-tuned to perform more specific tasks using fewer labeled examples. Current state-of-the-art semantic segmentation methods~\cite{DeepParsing, DeepLab,PSPNet} follow such a strategy.  Its necessity is driven by the high relative cost of annotating ground truth for spatially detailed segmentations~\cite{PASCAL,COCO}, and the accuracy gains achievable by combining different data sources and label modalities during training.  A collection of many images, coarsely annotated with a single label per image (ImageNet~\cite{ImageNet}), is still quite informative in comparison to a smaller collection with detailed per-pixel label maps for each image (PASCAL~\cite{PASCAL} or COCO~\cite{COCO}). In Chapter~\ref{chap:labelemb} we show that detailed ground truth annotation of this latter form contains additional information that existing schemes for training deep convolutional neural networks (CNNs) fail to exploit. By designing a new training procedure, we are able to capture some of this information, and as a result increase
accuracy at test time~\cite{Mostajabi_2018_CVPR}.

\chapter{Preliminaries}  \label{chap:perm}

We present definitions and methods used in later chapters of this dissertation.   

\section{Segmentation}

Segmentation refers to task of partitioning all or some part of the pixels in an image into coherent regions, where pixels in each region share similar attributes. A coherent region can be considered as a region in which pixels belong to it share similar feature statistics such as similar colors or texture patterns. Here we categorize segmentation methods into two main categories of semantic and class agnostic. There is also instance segmentation where the goal is to identify all instances of the segmented objects in an image. However, we are not going to address instance segmentation in this thesis.

\subsection{Class agnostic segmentation}

\begin{figure}
  \centering
\begin{tabular}{cccc}
\includegraphics[width=.22\textwidth]{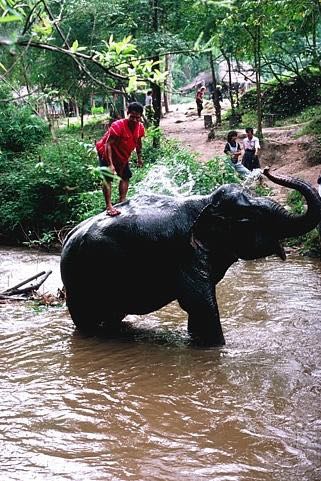} &
 \includegraphics[width=.22\textwidth]{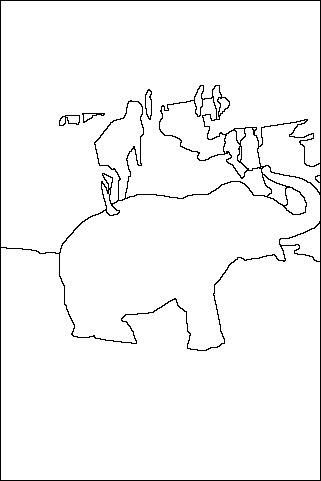}&
 \includegraphics[width=.22\textwidth]{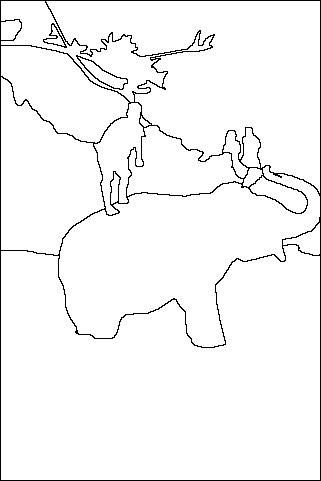} &
  \includegraphics[width=.22\textwidth]{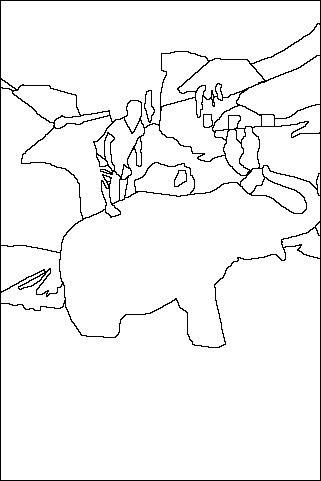} 
\end{tabular}
  \caption{An example image from BSDS dataset with its segmentation annotations provided by 3 different annotators.}\label{fig:sbsdsexamples}  
\end{figure}

In class agnostic segmentation, all or some of the pixels in the image will be partitioned into coherent regions, but unlike semantic segmentation we won't assign a class label to these regions. Class agnostic segmentation methods can be supervised or unsupervised. An example of rich dataset that can be used for supervised class agnostic segmentation is BSDS~\cite{BSDS500,BSDS300}. It contains segmentation annotations by 3 different annotators for natural images. Figure~\ref{fig:sbsdsexamples} shows an example image with its segmentation annotations. 

Large part of class agnostic segmentation literature addresses unsupervised learning of segmentation. Unsupervised segmentation methods can be applied to wide variety of images without requiring manually-labeled datasets and yet partition pixels to coherent regions. Since no ground truth annotations are provided, coherency is usually defined by distance between features of pixels withing the same segment.

Regardless of being supervised or unsupervised, part of class agnostic segmentation litrature is focused on oversegmentation. Oversegmentation of an image produces regions which is called superpixels. Superpixel methods can be based on low-level features such as color, intensity and texture statistics\cite{achanta2012slic, meanshift_pami02, carreira_cvpr10, ncut_pami00} or complex learned features of deep CNNs~\cite{deepsupix18,deepsupixwnet}. One of the benefit of using superpixel representation is reducing computational cost in many interesting segmentation methods. Semantic segmentation is viewed as structured prediction task and hence it is common to use a random field or structured support vector machine. Using superpixels will reduce the number of variables from order of millions (i.e. number of pixels) to order of hundreds which in turn significantly reduces computational cost of semantic segmentation methods. 

One of the benefits of using a superpixel representation of an image over a pixel representation is that superpixels provide a rich spatial neighborhood around each  pixel which respects contours. This neighborhood can be used to compute local features, rather than extracting local features from individual pixels or a random neighborhood around each pixel which may not respect image contours or objects boundaries. In Chapter~\ref{chap:introzoom} we show the effectiveness of implicitly encoding long and short range dependencies between image elements by extracting features from different spatial levels area around each superpixel.  Hence, class agnostic segmentation can be viewed as a useful preprocessing step for downstream tasks such as semantic segmentation. In Chapter~\ref{chap:introzoom} we show that the segments that are produced by unsupervised segmentation methods can be subsumed by our semantic segmentation method to assign class labels to superpixels. In the following we explain how SLIC~\cite{achanta2012slic}, an unsupervised segmentation method for producing superpixels works. We will use SLIC in Chapter~\ref{chap:introzoom}.

\subsubsection{SLIC}

Simple linear iterative clustering (SLIC) is a class agnostic segmentation method that produces regular and compact superpixels. With a tuned set of parameters SLIC superpixels respect contours and objects' boundaries. If we consider $N$ to be number of pixels in an image then the running time of SLIC is $\mathcal{O}(N)$. Having linear running time and producing high quality compact superpixels, makes SLIC one of the fastest superpixel algorithms to be used as a preprocessing step in variety of downstream tasks, such as semantic segmentation and depth estimation. 

SLIC clusters pixels based on color features and spatial proximity. The color spaced that is used in SLIC is CIELAB. In CIELAB color space, colors are defined by $lab$ values which has a nice property that the amount of visually perceived changes in color and brightness corresponds to similar changes in $lab$ values. SLIC performs clustering on five-dimensional [$labxy$] space in which $xy$ is the pixel coordinate. SLIC takes 2 parameters as input. The main parameter is $k$, the desired number of superpixels. Once $k$ is defined approximate size of each superpixel in an image with $N$ pixels will be $\frac{N}{K}$ pixels. This leads to a superpixel roughly centered at every grid interval $S = \sqrt{\frac{N}{K}}$.

\begin{figure}
 \centering

\subfloat[ $k = 75, m = 15$]{\includegraphics[width=.48\textwidth]{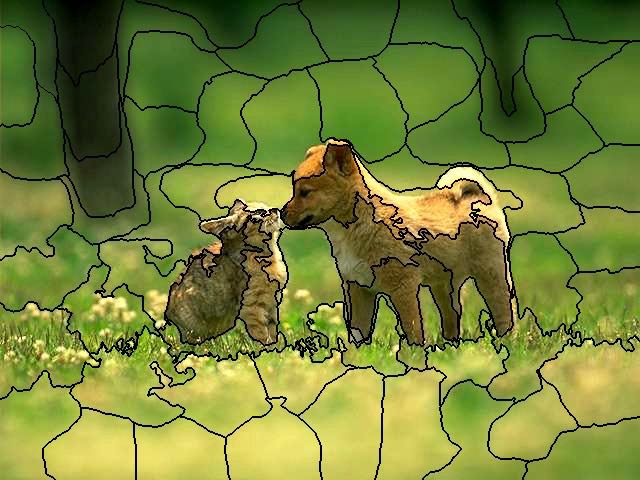}} 
\hspace{2mm}
 \subfloat[$k = 75, m = 75$]{\includegraphics[width=.48\textwidth]{figs/7575.png}} \\
\subfloat[ $k = 500, m = 15$]{ \includegraphics[width=.48\textwidth]{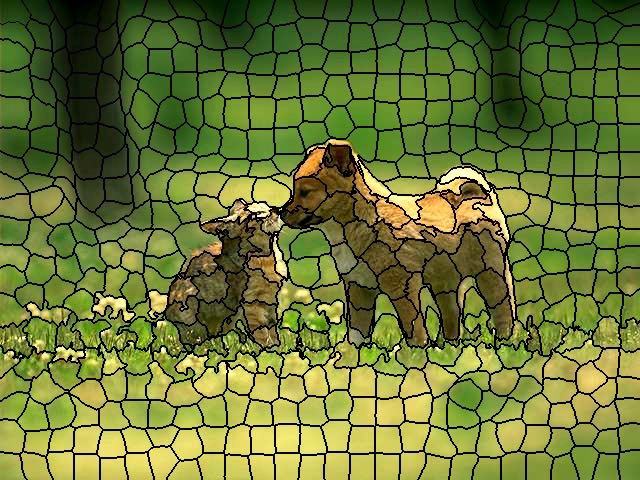}} 
\hspace{2mm}
\subfloat[ $k = 500, m = 75$]{ \includegraphics[width=.48\textwidth]{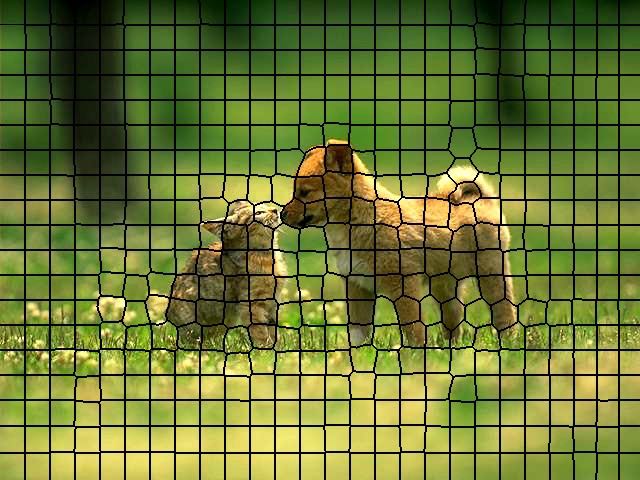}}

  \caption{SLIC segmentation results with different values for number of superpixels $k$ and compactness factor $m$. Higher values of $m$ produce superpixels that align less with image contours and more with spatial grid.}\label{fig:slicexamples}  
\end{figure}

%

The distance between each pixel to the center of a superpixel is based on two features: CIELAB color values and spatial coordinates. Spatial distance is normalized in SLIC, since directly using $xy$ coordinates makes SLIC sensitive to the size of the image. In Equation~\ref{eq:SLICD} $m$ is an input parameter to SLIC which controls relative weight of spatial distance with respect to color distance. Larger values of $m$ leads to more compact superpixels.

\begin{align*}
    d_{lab} &= \sqrt{(l_k - l_i)^{2} + (a_k - a_i)^{2} + (b_k - b_i)^{2}} \\
     d_{xy}  &= \sqrt{(x_k - x_i)^{2} + (y_k - y_i)^{2}} \\
   D_s &= d_{lab} + \frac{m}{S}d_xy \numberthis \label{eq:SLICD} 
\end{align*}
Initially, superpixel centers are placed along a regular spatial grid at interval $S$. The superpixel centers are associated with 5-dimensional vectors of color and coordinate features. These superpixel centers are moved to a position that has the lowest image gradient within a $3\times3$ neighborhood. This step prevents initial seeds from being placed at an edge or a noisy pixel. Next, each pixel in the image will be assigned to a closest superpixel center whose search area overlaps this pixel. The search area for each superpixel is  $2S\times2S$  area around the superpixel center on the xy plane. Limiting the search space to $2S\times2S$  area around each superpixel center makes SLIC one of the fastest superpixel algorithms. Once all pixels are associated with their nearest superpixel centers, then new superpixel centers will be computed as the average of 5-dimensional vectors of pixels belonging to them. This process will be iteratively repeated till convergence. Algorithm~\ref{SLICalg} summarizes SLIC. Figure~\ref{fig:slicexamples} shows superpixels produced by SLIC with different values for number of superpixels $k$ and compactness factor $m$.

\begin{algorithm}
\caption{SLIC }\label{SLICalg}
\begin{algorithmic}[1]
\State Initialize $k$ cluster centers along a regular spatial grid with grid intervals $S$.
\State Move cluster center's to a position that has the lowest gradient within a $3\times3$ neighborhood.
\State Residual error $E \gets \infty$.
\State Pixel distance $d(p) \gets \infty $ for each pixel $p$. 
\While {$E \geqslant threshold$}
	\For {each cluster center $c_i$}
		\For {each pixel $p$ within $2S\times2S$ square neighborhood of $c_i$}
			\State compute the distance $D$ (Eq.~\ref{eq:SLICD}) between cluster center $c_i$  and $p$. 
			\If {distance $D < d(p)$}
				\State Assign $p$ to $c_i$.
				\State $d(p) \gets D$.
			\EndIf
		\EndFor
	\EndFor
	\State Compute new cluster centers based on pixel assignments
	\State Compute residual error $E$ between recomputed cluster centers and previous ones
\EndWhile 
\end{algorithmic}
\end{algorithm}

\subsection{Semantic segmentation} \label{sec:semseg}
Semantic segmentation is a supervised segmentation in which the goal is to assign a label, from a predefined set of class labels, to each pixel of a given image. In semantic segmentation, ground truth annotations are provided. These annotations serve as a reference of what the segmentation method should produce. In semantic segmentation, the set of object classes that we want to segment is well defined. Semantic segmentation does not differentiate between multiple instances of the same object class. In instance segmentation the goal is to identify all instances of the segmented objects in an image. However, instance segmentation does not address non-object or stuff classes such as sky. We can be more ambitious and consider panoptic segmentation~\cite{Panoptic} which unifies semantic segmentation and instance segmentation. In Panoptic segmentation the goal is to asssign category level label to every pixel in the given image and also identify object instances as well. This makes it a more difficult task than semantic segmentation or instance segmentation. In this thesis, we focus on semantic segmentation. Conditional random fields (CRFs) have been widely used in semantic segmentation~\cite{kohli_cvpr08,shotton_ijcv09,ladicky_iccv09,lempitsky2011pylon,boix_ijcv12,koltun2011efficient}. In the following we give an overview of CRFs and discuss about pairwise CRFs for semantic segmentation.


%
%

\subsubsection{Conditional random fields}\label{subsec:CRFs}

Consider each pixel as a random discrete variable $X_i$. Then $\textbf{X} = \lbrace X_1, X_2, X_3, \cdots, X_N \rbrace$ is a set of random variables corresponding to image pixels in an image $I$ with $N$ pixels. Each variable $X_i$ can take a label $\textbf{x} \in \lbrace l_1, l_2, l_3, \cdots, l_C \rbrace$. $P(X=\textbf{x})$ is the probability of random variables $X$ taking any labeling $\textbf{x}$. Then the posterior distribution $P(X=\textbf{x}|I)$ over the labellings of the CRF is a Gibbs distribution:

\begin{equation}
  \label{eq:Gibs}
  P(X=\textbf{x}|I) = \frac{1}{Z}exp \left( -\sum_{c \in \mathcal{C}_\mathcal{G}} \phi_c(\textbf{x}_c|I) \right),
\end{equation}
where $Z$ is a normalization constant called partition function and $\mathcal{G} = (\mathcal{V},\mathcal{E})$ is a graph over set of variables $X$. $\phi_c(\textbf{x}_c|I)$ is the potential function over the variables in the clique $c$. The corresponding Gibbs energy will be

\begin{equation}
  \label{eq:Gibsenergy}
    E(\textbf{x}) = -log P(X=\textbf{x}|I) - log Z = \sum_{c \in \mathcal{C}_\mathcal{G}} \phi_c(\textbf{x}_c|I) .
\end{equation}

\subsubsection{Pairwise CRFs for semantic segmentation}

Semantic segmentation is widely treated as a structured prediction task in which labels themselves exhibit internal structure. A common approach is to utilize conditional random fields and frame semantic segmentation as an energy minimization problem over pixel or superpixel labelings. Energy minimization will be done over unary, pairwise and possibly higher-order terms.

In our notations we use $\psi_c(\textbf{x}_c)$ to denote $ \phi_c(\textbf{x}_c|I)$. The corresponding energy for CRF model with unary and pairwise potentials is
\begin{equation}
  \label{eq:pairwiseCRF}
    E(\textbf{x}) = \sum_{i \in \mathcal{V}} \psi_i(x_i) + \sum_{(i, j) \in \mathcal{E}} \psi_{ij}(x_i, x_j).
\end{equation}

The unary term is usually based on estimated probability of a pixel or superpixel to take on a particular label produced by a classifier. Pairwise terms capture labeling dependencies between pair of variables corresponding to
pixels or superpixels. Pairwise terms are often called smoothness terms, because they advocates neighboring pixels receive same labels. Depending on the CRF model, pairwise connections can be limited to neighboring pixels, which results in capturing only short-range dependecies. In dense and fully connected CRFs, pairwise term can capture both short-range and long-range dependecies between pairs of variables. Local evidence is often incorporated in unary terms, while interactions between label assignments are captured by pairwise terms. Kr\"{a}henb\"{u}hl et al.~\cite{koltun2011efficient} introduce one of the most successful and widely-used inference methods for fully connected CRFs. When the graph $\mathcal{G} = (\mathcal{V},\mathcal{E})$ in Equation~\ref{eq:pairwiseCRF} is a complete graph on $\textbf{X}$, then the corresponding CRF model is a fully connected pairwise CRF. The pairwise potential in~\cite{koltun2011efficient} has the form

 \begin{equation}
  \label{eq:fullyconnectedpairwiseCRF}
     \psi_{ij}(x_i, x_j) = \mu(x_i, x_j) \sum_{m = 1}^{K}  w^{(m)}k^{(m)}(\textbf{f}_i, \textbf{f}_j).
\end{equation}
In Equation~\ref{eq:fullyconnectedpairwiseCRF}, $\mu$ is a smoothness term which encourages nearby similar pixels to have the same label. $w^{(m)}$ are linear combination weights and each $k^{(m)}$ is a Gaussian kernel

 \begin{equation}
  \label{eq:kkernels}
     k^{(m)}(\textbf{f}_i, \textbf{f}_j) = exp \left( -\frac{1}{2}(\textbf{f}_i - \textbf{f}_j)^T \Lambda^{(m)} (\textbf{f}_i - \textbf{f}_j) \right),
\end{equation}
where $\textbf{f}_i$, $\textbf{f}_j$ are feature vectors of pixels $i$ and $j$. In Chapter~\ref{chap:chapter2} we apply fully connected pairwise CRFs to the output of our weakly supervised semantic segmentation model.   

\pagebreak

\chapter{Zoom-out Features}  \label{chap:introzoom}

We introduce the zoom-out architecture, a purely feed-forward architecture for semantic segmentation. We map small image elements to rich feature representations extracted from a sequence of nested regions of increasing extent. These regions are obtained by "zooming out" from the pixel/superpixel
all the way to scene-level resolution. This approach exploits
statistical structure in the image and in the label space without setting up explicit structured prediction mechanisms,
and thus avoids complex and expensive inference. In Chapter~\ref{chap:chapter2} we will use zoom-out representations to learn category-level semantic segmentation purely from image-level classification tags. 

\section{Introduction}
Learning rich representations of data is an essential block of recognition in modern
computer vision. Consider one of the central vision tasks, \emph{semantic
  segmentation}: assigning to each
pixel in an image a category-level label. Despite the attention it has
received, it remains challenging, largely due to complex
interactions between neighboring as well as distant image elements,
the importance of global context, and the interplay between semantic
labeling and instance-level detection.
A widely accepted conventional wisdom, followed in much of
the segmentation literature, is that 
segmentation should be treated as a structured prediction task, which most
often means using a random field or structured support vector machine
model of considerable complexity. Refer to Section~\ref{sec:semseg} for further details.

This in turn brings up severe challenges, among them the intractable
nature of inference and learning in many ``interesting'' models. To alleviate this,
many previously proposed methods rely on a pre-processing
stage, or a few stages, to produce a manageable number of hypothesized
regions, or even complete segmentations, for an image. These are then
scored, ranked or combined in a variety of ways.

\begin{figure}[!th]
\centering
  \begin{tikzpicture}
    \node[rectangle,draw=black,thick,fill overzoom
    image=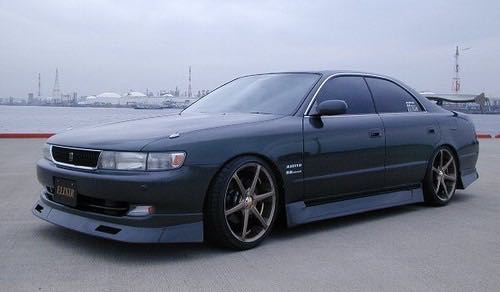,minimum width=7cm,minimum height=4.08cm,opacity=.5] (image) {};
    \node[fit=(image),draw=green!60!black,line width=2pt,inner
    sep=0pt,opacity=.8] (global) {};
    
    \draw[red,line width=1pt] plot[smooth] file {figs/2007_003143_suppix252.txt};
    \draw[name path=zo2,cyan,line width=1pt] plot[smooth] file {figs/2007_003143_sp252_zo1.txt};

    \draw[name path=zo6,olive,line width=1pt] plot[smooth] file {figs/2007_003143_sp252_zo6.txt};

    \draw[name path=zo10,blue!50!red,line width=1pt] plot[smooth] file {figs/2007_003143_sp252_zo10.txt};

    \draw[name path=zo13,blue!80!black,line width=1pt] plot[smooth] file {figs/2007_003143_sp252_zo13.txt};

    \node[rectangle,draw=orange,line width=2pt,minimum
    width=2.492cm,minimum height=1.904cm,opacity=.8] (n3) at (-1.295,-0.189) {};
    
    \node[above=of global.north west,above=15pt] (flocal) {\textcolor{cyan}{$\bfgreek{phi}_1$}};
    \node[right=of flocal,right=0pt] (dots1) {\ldots};
    \node[right=of dots1,right=0pt] (f6) {\textcolor{olive}{$\bfgreek{phi}_6$}};
    \node[right=of f6,right=0pt] (dots2) {\ldots};
    \node[right=of dots2,right=0pt] (f10) {\textcolor{blue!50!red}{$\bfgreek{phi}_{10}$}};
    \node[right=of f10,right=0pt] (dots3) {\ldots};
    \node[right=of dots3,right=0pt] (f13) {\textcolor{blue!80!black}{$\bfgreek{phi}_{13}$}};
    \node[right=of f13,right=0pt] (fn3) {\textcolor{orange}{$\bfgreek{phi}_{window}$}}; 
    \node[right=of fn3,right=0pt] (fglobal) {\textcolor{green!60!black}{$\bfgreek{phi}_{scene}$}}; 
    \node[fit=(flocal)(fglobal),inner sep=0pt,draw=black,thick]
    (feat) {};

    \coordinate (pt1) at (-1.456,0);
    \draw[thick,cyan,-latex] (pt1) -- (flocal);

    \coordinate (pt6) at (-1.218,0.210);
    \draw[thick,olive,-latex] (pt6) -- (f6);

    \coordinate (pt10) at (0.224,0.700);
    \draw[thick,blue!50!red,-latex] (pt10) -- (f10);

    \coordinate (pt13) at (0.644,1.498);
    \draw[thick,blue,-latex] (pt13) -- (f13);

    \coordinate (ptw) at (0.042,-0.140);
    \draw[thick,orange,-latex] (ptw) -- (fn3);

    \path (image.north east) -- ++(-.5,0) coordinate (ptglobal);
    \draw[thick,green!60!black,-latex] (ptglobal) -- (fglobal);

    \node[above=of feat,draw=black,thick,above=10pt] (mlp)
    {multi-layer perceptron};
    \draw[thick,-latex] (feat)--(mlp);
    \draw[thick,-latex] (mlp.north) -- ++(0,10pt) node[above] {``car''};
  \end{tikzpicture}
  \caption{Schematic description of our approach. Features are
    extracted from a nested sequence of ``zoom-out'' regions around
    the superpixel at hand (red). Here we show four out of
    thirteen levels: 1
    (cyan), 6 (olive), 10 (purple) and 13 (blue), as well as the
    window level (orange) and scene level (green). The features
    computed at all levels are comcantenated and fed to a multi-layer
    perceptron that classifies the superpixel.}\label{fig:main}
\end{figure}
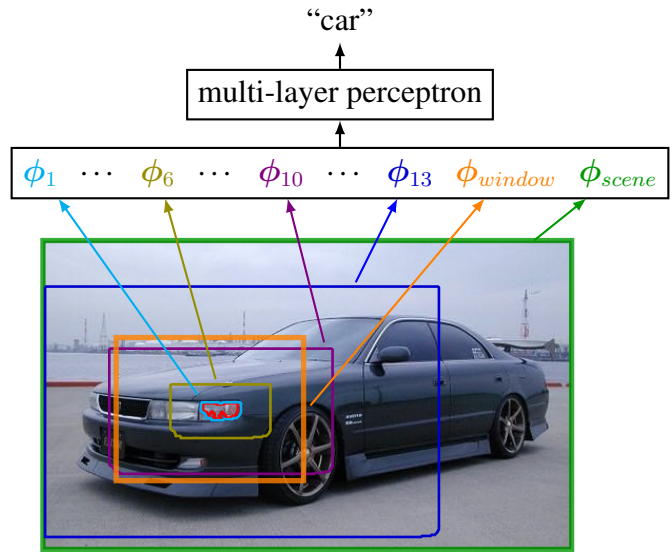

Here we consider a departure from these conventions, and approach semantic
segmentation as a single-stage classification task, in which each
image element (pixel or superpixel) is labeled by a feedforward model, based on the evidence
computed from the image. Surprisingly, in experiments on PASCAL VOC 2012
segmentation benchmark, we show that this simple sounding approach
leads to results significantly surpassing previously published
ones.

The ``secret'' behind our method is that the evidence used in the
feedforward classification is not computed from a small local region
in isolation, but collected from a sequence of levels, obtained by
``zooming out'' from the close-up view of the superpixel. Starting from the superpixel itself, to a small region surrounding it,
to a larger region around it and all the way to the entire image, we
compute a rich feature representation at each level and combine all
the features before feeding them to a classifier. This allows us to
exploit statistical structure in the label space and dependencies
between image elements at different resolutions without explicitly
encoding these in a complex model.

We do not mean to dismiss structured
prediction or inference, and as we discuss in
Section~\ref{sec:zoomconc}, these tools are complementary to our
architecture. In this chapter we explore how far we can go
without resorting to explicitly structured models.

We use convolutional neural networks (CNNs) to extract features from
larger zoom-out regions. CNNs, (re)introduced
to vision in 2012, have facilitated a dramatic advance in classification,
detection, fine-grained recognition and other vision tasks. Segmentation has remained conspicuously left out from this wave of
progress; while image classification and detection accuracies on
VOC has improved by nearly 50\% (relative), segmentation numbers has
improved only modestly. A big reason for this was that neural networks
were inherently geared for ``non-structured'' classification and
regression. In this work we propose a way to
leverage the power of representations learned by CNNs, by framing
segmentation as classification and making the structured aspect of it
implicit. 
Finally, we show that use of
a multi-layer neural network trained with asymmetric loss to
classify pixels/superpixels represented by zoom-out features, leads
to significant improvement in segmentation accuracy over simpler
models and conventional (symmetric) loss.

Below we give a high-level description of our method, then discuss
related work and position our work in its context. Most of the
technical details are deferred to Section~\ref{sec:expzoom}, in which we
describe implementation and report on results, before concluding in Section~\ref{sec:zoomconc}.

\section{Related work}\label{sec:related}

The literature on segmentation is vast, and here we only mention work
that is either significant as having achieved state-of-the-art
performance in recent times, or is closely related to ours in some
way. In Sections~\ref{sec:expzoom} and ~\ref{sec:zoomfurther} we compare our performance to
that of most of the methods mentioned here.

Many prominent segmentation methods rely on conditional random fields (CRFs) over nodes
corresponding to pixels or superpixels.
Such models incorporate local evidence in unary potentials, while
interactions between label assignments are captured by pairwise and
possibly higher-order potentials. This includes various hierarchical CRFs~\cite{shotton_ijcv09,ladicky_iccv09,lempitsky2011pylon,boix_ijcv12}. In
contrast, we let the zoom-out features (in CRF terminology, the unary potentials) capture higher-order structure.

Another recent trend has been to follow a multi-stage
approach: First a set of proposal regions is generated, by a
category-independent~\cite{carreira2012cpmc,UijlingsIJCV2013} or
category-aware~\cite{arbelaez_cvpr12} mechanism.
Then the regions are scored or ranked based on their compatibility
with the target classes. Work in
this vein
includes~\cite{carreira2012semantic,ion2011probabilistic,arbelaez_cvpr12,carreira_ijcv12,Li2013codemaps}.
A similar approach is taken
in~\cite{yadollahpour2013discriminative}, where multiple segmentations
obtained from~\cite{carreira2012semantic}
are re-ranked using a discriminatively trained model.
Recent advances along these lines include~\cite{hariharan2014simultaneous}, which
uses CNNs, 
and~\cite{dhruv-new}, which improves upon the re-ranking
in~\cite{yadollahpour2013discriminative}, also using
CNN-based features.
In contrast to most of the work in this group, we do not rely on
region generators, and directly operate on pixels or limit preprocessing to
over-segmentation of the image into a large number of superpixels.

The idea of using non-local evidence in segmentation, and specifically of
computing features over a neighborhood of superpixels, was
introduced in~\cite{fulkerson2009class} and~\cite{lim2009context};
other early work on using forms of context for segmentation
includes~\cite{shotton_ijcv09}. 
A study in~\cite{lucchi2011spatial} concluded that non-unary terms may be unnecessary when neighborhood and global
information is captured by unary terms, but the results were
significantly inferior to state of the art at the time.

Recent work closest to ours
includes~\cite{farabet2013learning,pinheiro2014recurrent,socher2011parsing,mostajabi2014robust,huang2013deep}. 
In~\cite{farabet2013learning}, the same CNN is applied on
different resolutions of the image and
combined with a tree-structured graph over superpixels to impose
smoothness. In~\cite{mostajabi2014robust} the features applied to
multiple levels are
also homogeneous, and hand-crafted rather than learned. In~\cite{pinheiro2014recurrent} there is also a single
CNN, but it is applied in a recurrent fashion, i.e., input to the
network includes, in addition to the scaled image, the feature maps
computed by the network at a previous level. A similar idea is pursued
in~\cite{huang2013deep}, where it is applied to boundary detection in
3D biological data. In contrast with all of these, we use different
feature extractors across levels, some of them with a much smaller
receptive field than any of the networks in the
literature. We show in
Section~\ref{sec:exp} that our approach obtains better performance (on
Stanford Background Dataset) than that reported
for~\cite{farabet2013learning,pinheiro2014recurrent,socher2011parsing,mostajabi2014robust};
no comparison to~\cite{huang2013deep} is available.

Finally, our work shares some ideas with other concurrent
efforts. In Fully Convolutional Networks (FCN)~\cite{FCN} the fully connected layers are converted to convolutional layers. Segmentation predictions are produced at different layers in the network and then upsampled and summed to produce final segmentation output. The main difference with Fully Convolutional Networks (FCN)~\cite{FCN} is that we concatenate features, rather than summing predictions up across
levels. This allows a multi-layer network on top of zoom-out features to combine them with a 
learned non-linear function. 
Hpercolumns~\cite{Hypercolumns} is closest to our method where they concatenate features across different layers of network. However we incorporate a much wider range of zoom-out
levels, including scene level features for each superpixel. As a result we achieve a significantly
better performance on VOC 2012 test set than either of these methods.

\section{Zoom-out feature fusion}\label{sec:zoommethod}
\begin{figure*}[!th]
  \centering
  \begin{tabular}{ccc}
  \includegraphics[height=1.6in]{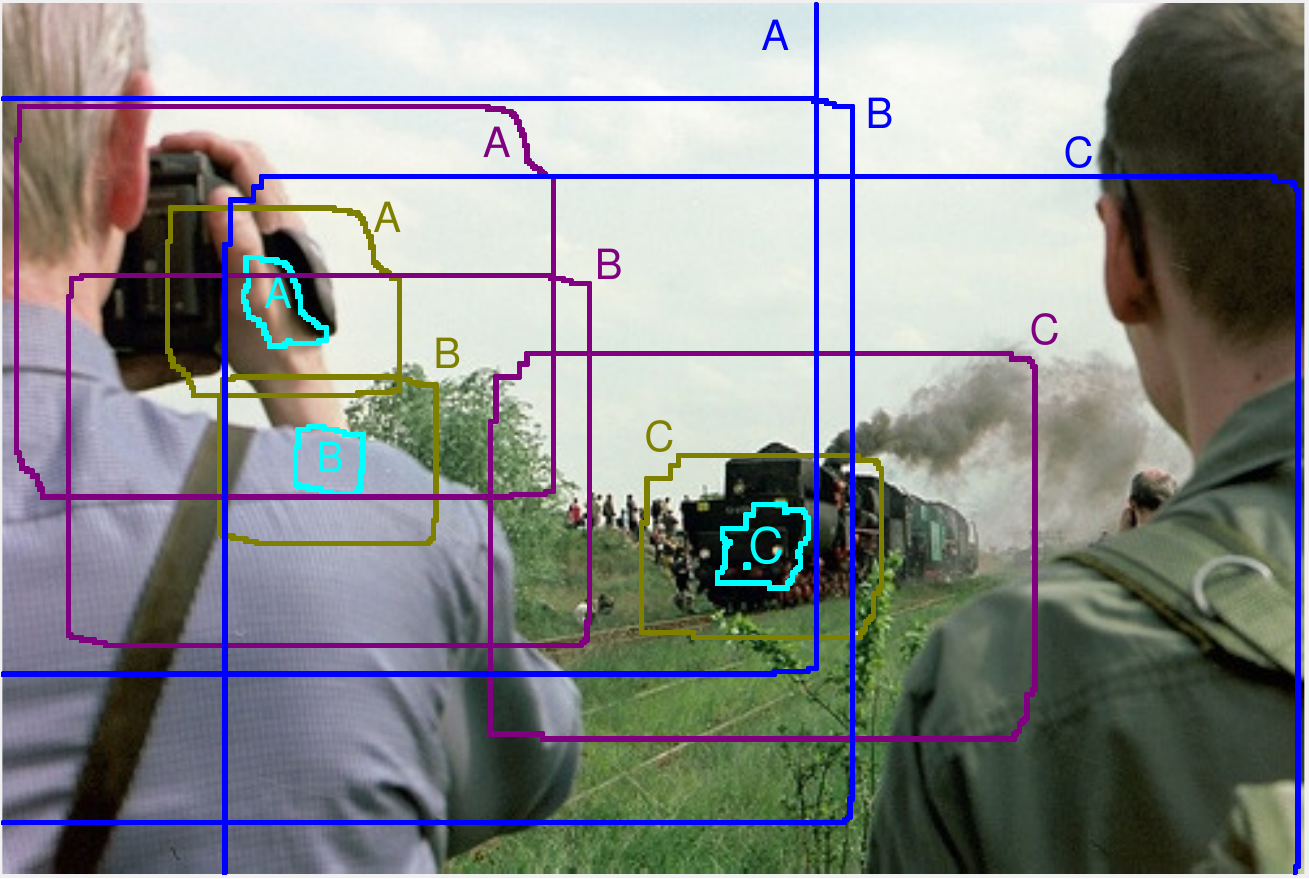} &
  \includegraphics[height=1.6in]{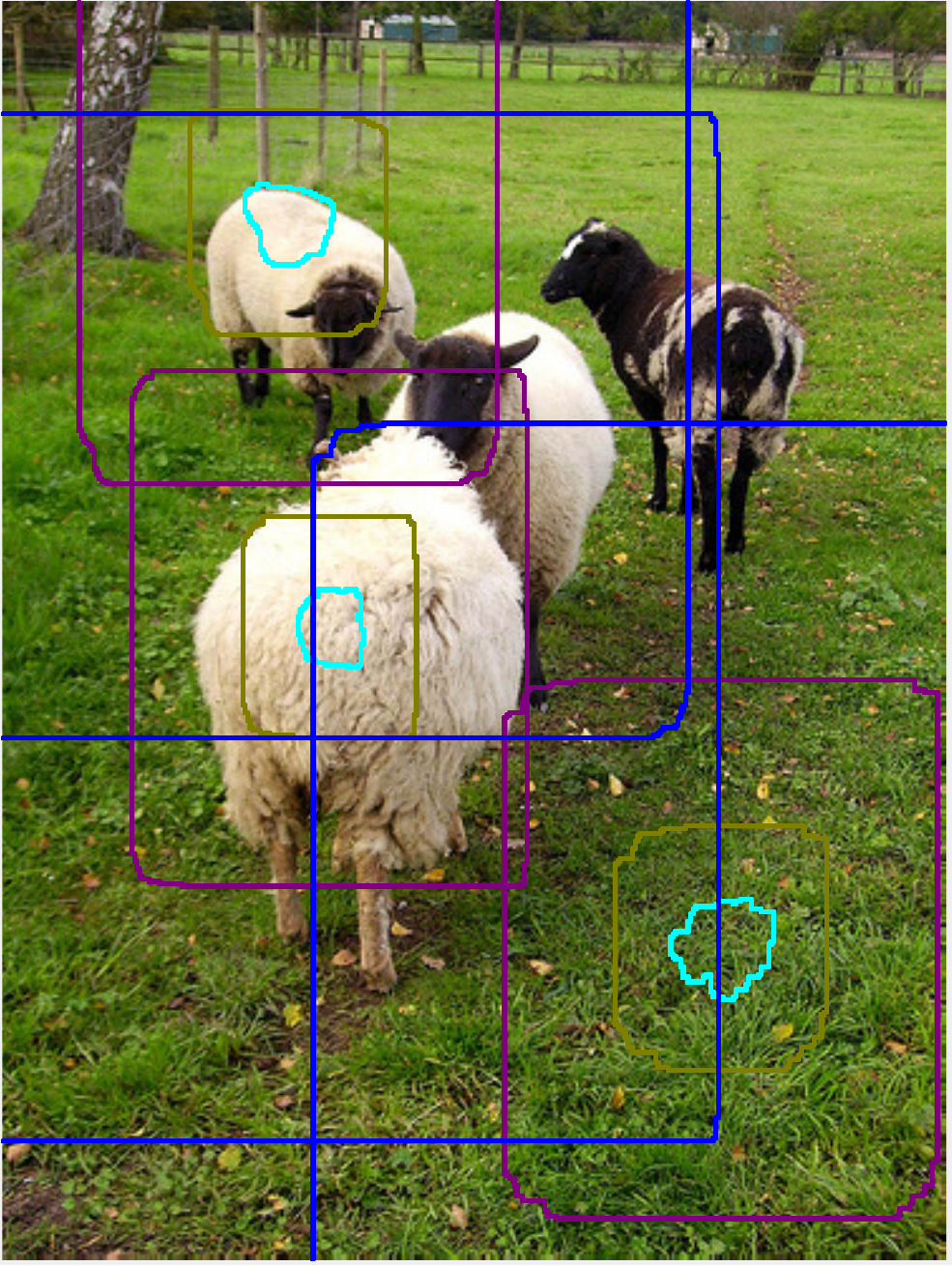} &
  \includegraphics[height=1.6in]{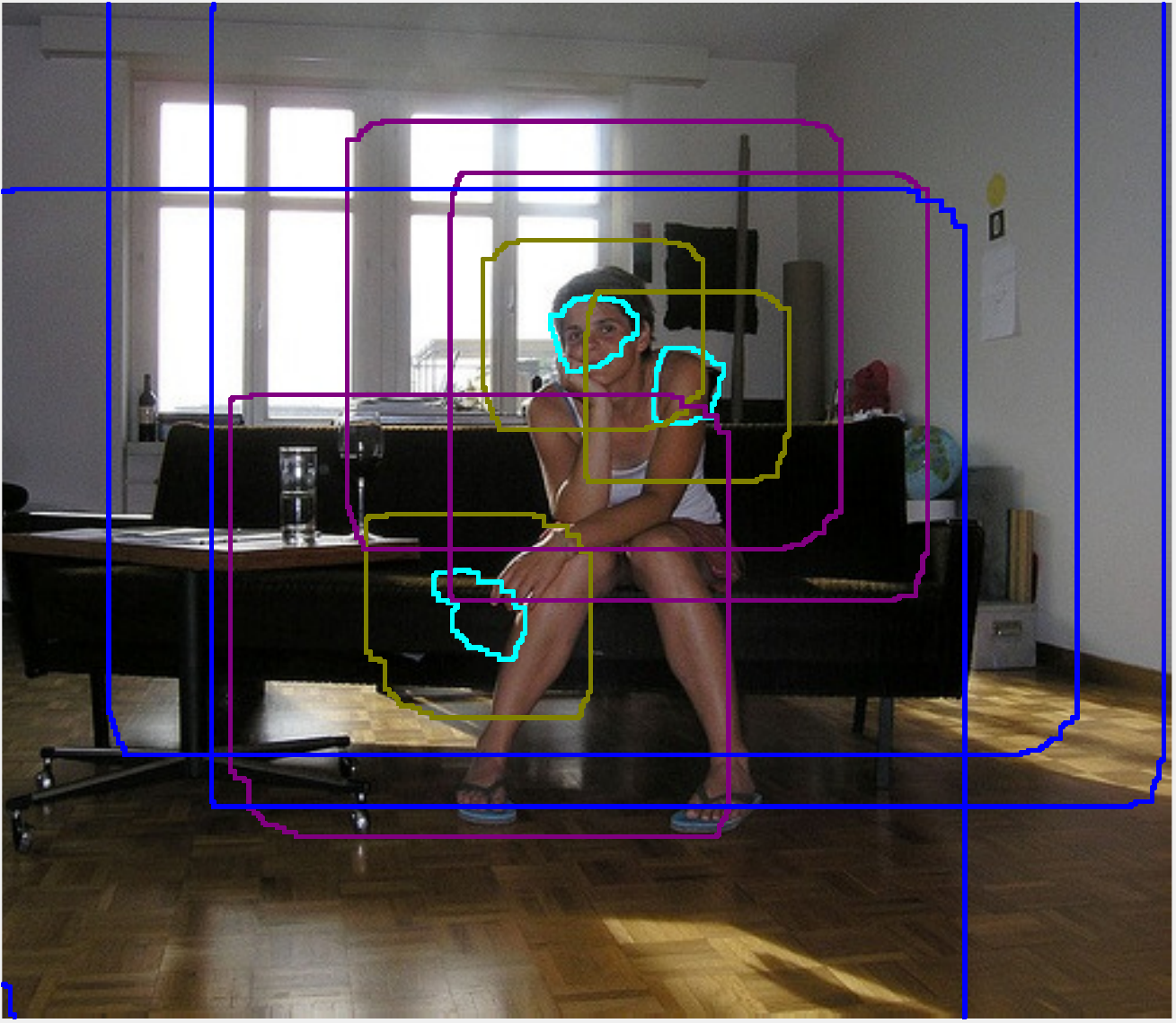}
  \end{tabular}

  \caption{Examples of zoom-out regions. We show four out of fifteen
    levels: \textcolor{cyan}{1}(\textcolor{cyan}{cyan}, nearly
    matching the superpixel boundaries), \textcolor{olive}{6 (olive)}, \textcolor{blue!50!red}{10 (purple)} and \textcolor{blue}{13 (blue)}.}
  \label{fig:regions}
\end{figure*}

\begin{figure}[H]
\begin{minipage}{.35\textwidth}
\begin{tabular}{c                c}
    \begin{tabular}{c}
        \includegraphics[width=\textwidth]{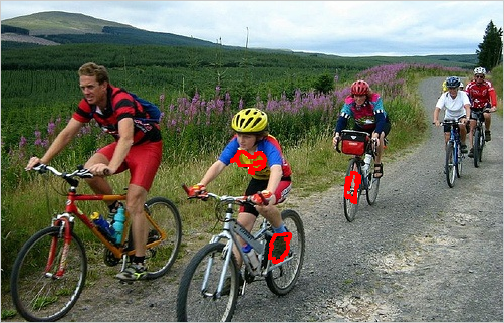}\\
    \begin{tabular}{ccc}
        \includegraphics[width=.3\textwidth]{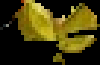} &
        \includegraphics[width=.3\textwidth]{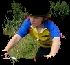} &
        \includegraphics[width=.3\textwidth]{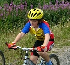} \\
        \includegraphics[width=.24\textwidth]{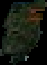} &
        \includegraphics[width=.3\textwidth]{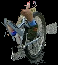} &
        \includegraphics[width=.3\textwidth]{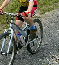} \\
        \includegraphics[width=.18\textwidth]{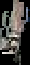} &
        \includegraphics[width=.3\textwidth]{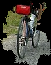} &
        \includegraphics[width=.3\textwidth]{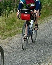}
    \end{tabular}
    \end{tabular} &
    \begin{tabular}{c}
    \includegraphics[width=.78\textwidth]{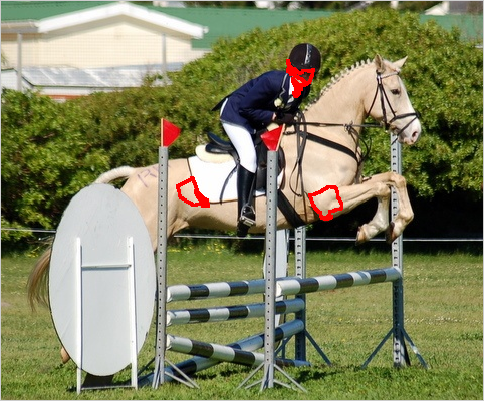}\\
    \begin{tabular}{ccc}
        \includegraphics[width=.3\textwidth]{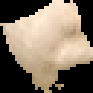} &
        \includegraphics[width=.3\textwidth]{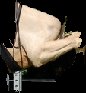} &
        \includegraphics[width=.3\textwidth]{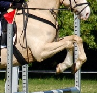} \\
        \includegraphics[width=.24\textwidth]{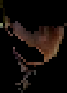} &
        \includegraphics[width=.3\textwidth]{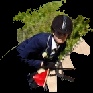} &
        \includegraphics[width=.3\textwidth]{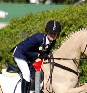} \\
        \includegraphics[width=.3\textwidth]{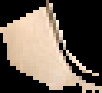} &
        \includegraphics[width=.3\textwidth]{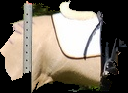} &
        \includegraphics[width=.3\textwidth]{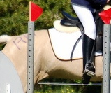}
    \end{tabular}
    \end{tabular} \\
    
    \begin{tabular}{c}
    \includegraphics[width=.78\textwidth]{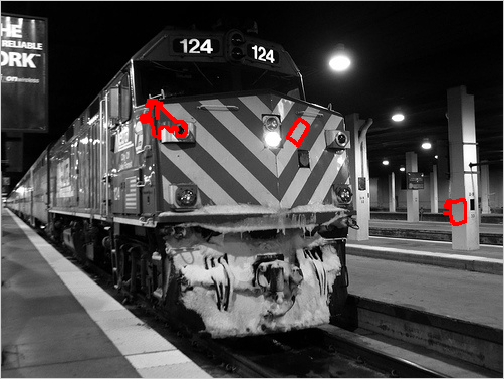}\\
    \begin{tabular}{ccc}
        \includegraphics[width=.3\textwidth]{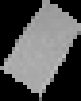} &
        \includegraphics[width=.3\textwidth]{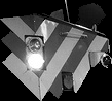} &
        \includegraphics[width=.3\textwidth]{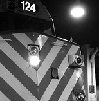} \\
        \includegraphics[width=.24\textwidth]{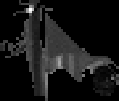} &
        \includegraphics[width=.3\textwidth]{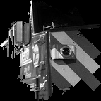} &
        \includegraphics[width=.3\textwidth]{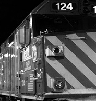} \\
        \includegraphics[width=.3\textwidth]{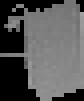} &
        \includegraphics[width=.3\textwidth]{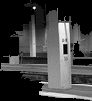} &
        \includegraphics[width=.3\textwidth]{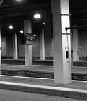}
    \end{tabular}
    \end{tabular} &
\begin{tabular}{c}
    \includegraphics[width=.78\textwidth]{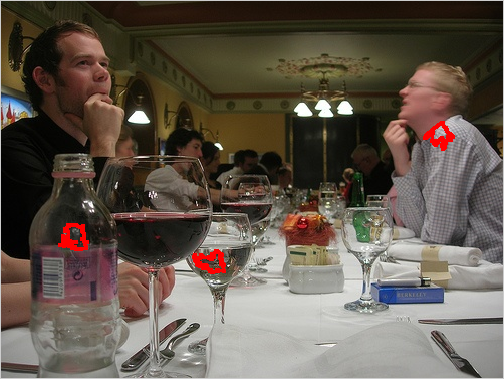}\\
    \begin{tabular}{ccc}
        \includegraphics[width=.3\textwidth]{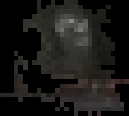} &
        \includegraphics[width=.3\textwidth]{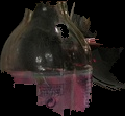} &
        \includegraphics[width=.3\textwidth]{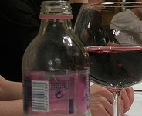} \\
        \includegraphics[width=.24\textwidth]{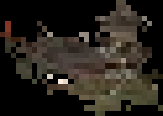} &
        \includegraphics[width=.3\textwidth]{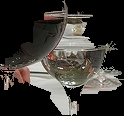} &
        \includegraphics[width=.3\textwidth]{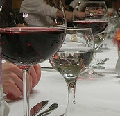} \\
        \includegraphics[width=.3\textwidth]{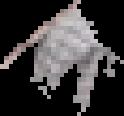} &
        \includegraphics[width=.3\textwidth]{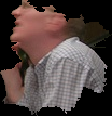} &
        \includegraphics[width=.3\textwidth]{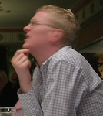}
    \end{tabular}
    \end{tabular} 
\end{tabular}
\end{minipage}
    \caption{Showing three superpixels in each image (top), followed
      by corresponding zoom-out regions that are seen by the
      segmenation process, (left) superpixel, (center) proximal
      region, (right) distant. As we zoom in from image to
      the superpixel level, it becomes increasingly hard to tell what
      we are looking at; however, the higher zoom-out levels provide
      rich contextual information.}
 \label{fig:region-examples}
\end{figure}

We cast category-level segmentation of an image as classifying a set
of pixels/superpixels. In the case of superpixels, since we expect to apply the same classification
machine to every superpixel, we would like the nature of the superpixels
to be similar, in particular their size. In our experiments we use
SLIC~\cite{achanta_pami12}, but other methods that produce
nearly-uniform grid of superpixels might work similarly
well. Figures~\ref{fig:regions} and~\ref{fig:region-examples}  provides a few illustrative examples
for this discussion. As it is shown in~\ref{fig:region-examples} local patches are ambiguous in a sense that it is difficult to tell which objects they belong to. As we zoom out and look at larger regions we can access rich contextual information. In the following we describe different levels of zoom-out features. For the following sections we focus on superpixels. In Section~\ref{sec:zoomfurther} we discuss zoom-out descriptors for pixels.

\subsection{Scoping the zoom-out features}

The main idea of our zoom-out architecture is to allow features
extracted from different levels of spatial context around the
superpixel to contribute to the labeling decision at that
superpixel. Before going into specifics of how we define the zoom-out
levels, we discuss the role we expect different levels to play.

\paragraph{Local} The narrowest scope is the superpixel itself. We
expect the features extracted here to capture local evidence: color,
texture, small
intensity/gradient patterns, and other properties computed over a
relatively small contiguous set of pixels. The
local features may be quite different even for neighboring
superpixels, especially if these straddle category or object
boundaries.

\paragraph{Proximal} As we zoom out and include larger spatial area
around the superpixel,
we can capture visual cues from surrounding superpixels. Features
computed from these levels may capture
information not available in the local scope; e.g., for locations at the
boundaries of objects they will represent the appearance of both
categories. For classes with non-uniform appearance they may better
capture characteristic distributions for that class. We can expect
somewhat more complex features to be useful at this level, but it is
usually 
still too myopic for confident reasoning about presence of objects.

Two neighboring
superpixels could still have quite different features at this level,
however some degree of smoothness is likely to arise from the
significant overlap between neighbors' proximal regions, e.g., A and B
in Fig.~\ref{fig:regions}.
As another example,
consider color features over the body of a leopard; superpixels for individual dark brown spots might
appear quite different from their neighbors (yellow fur) but their
proximal regions will have pretty similar distributions (mix of yellow
and brown).
Superpixels that are sufficiently far from each other could still, of
course, have drastically different proximal features, e.g., A and C in Fig.~\ref{fig:regions}.

\paragraph{Distant} Zooming out further, we move to the distant levels
:
regions large enough to include sizeable fractions of objects, and
sometimes entire objects. At this level our scope is wide enough to
allow reasoning about shape, presence of more complex patterns in color and
gradient, and the spatial layout of such patterns. Therefore we can expect
more complex features that represent these properties to be useful here.
Distant regions are more
likely to straddle true boundaries in the image, and so this
higher-level feature extraction may include a significant area in both
the category of the superpixel at hand and nearby categories. For
example, consider a person sitting on a chair; bottle on a dining
table; pasture animals on the background of grass, etc. Naturally we
expect this to provide useful information on both the appearance of a
class and its context.

For nearby superpixels and far enough zoom-out level, distant regions will have a very large
overlap, which will gradually diminish with distance between superpixels.  This
is likely to lead to somewhat gradual changes
in features, and to impose a system of implicit smoothness
``terms'', which depend both on the distance in the image and on the
similarity in appearance in and around superpixels. Imposing
such smoothness in a CRF usually leads to
a very complex, intractable model.

\paragraph{Scene} The final zoom-out scope is the entire scene. Features computed at this level capture ``what
kind of an image'' we are looking at. One aspect of this global
context is image-level classification: since state of the art in image
classification seems to be dramatically higher than that of detection
or segmentation~\cite{everingham2014pascal,russakovsky2014imagenet}. We can expect
image-level features to help determine presence of categories in the
scene and thus guide the segmentation.

More subtly, features that are useful for classification can be
directly useful for global support of local labeling decisions; e.g.,
lots of green in an image supports labeling a (non-green) superpixel
as cow or sheep more than it supports labeling that superpixel as
chair or bottle, other things being equal. Or, lots of
straight vertical lines in an image would perhaps suggest man-made
environment, thus supporting categories relevant to indoors or urban
scenes more than, say, wildlife.

At this global level, all superpixels in an image will of course have the same
features, imposing (implicit, soft) global constraints. This is yet another form
of high-order interaction that is hard to capture in
a CRF framework, despite numerous attempts~\cite{boix_ijcv12}.

\subsubsection{CNN-based zoom-out}\label{sec:zoommethod}
\begin{figure}[!th]
  \centering
  \input{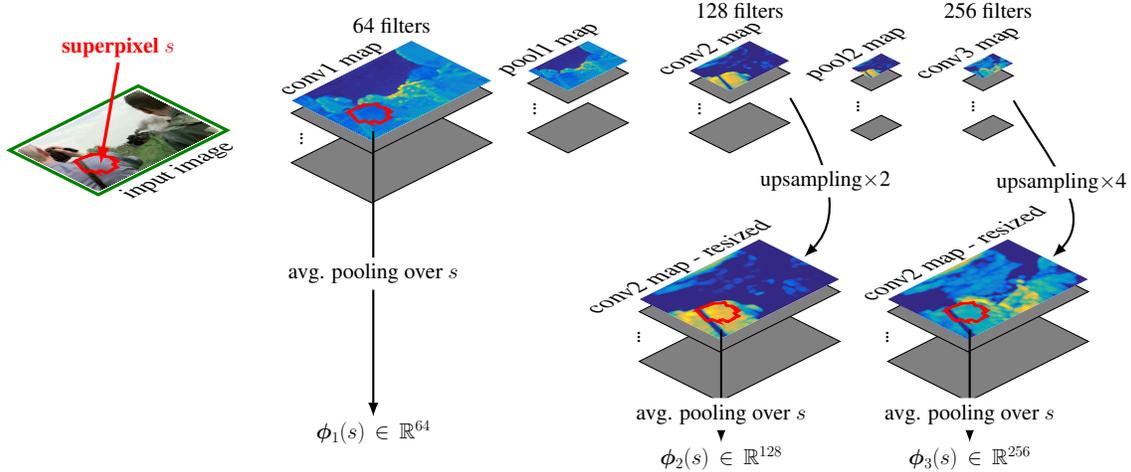}
  \caption{Illustration of zoom-out feature extraction procedure for a simple network with three convolutional and two pooling layers.}
  \label{fig:zo}
\end{figure}

 A feature map computed by a convolutional layer with $k$ filters assigns a
$k$-dimensional feature vector to each receptive field of that
layer. For most layers, this feature map is of lower resolution than
the original image, due to sub-sampling induced by pooling and
possibly stride in filtering at previous layers. We 
upsample the feature map to the original image resolution, if necessary. This
produces a $k$-dimensional feature vector for every \emph{pixel} in
the image. Pooling these vectors over a superpixel gives us a
$k$-dimensional feature vector describing that superpixel.
Figure~\ref{fig:zo} illustrated this feature computation for a
superpixel with a toy network with three convolutional layers,
interleaved with two pooling layers (2$\times$2 non-overlapping
pooling receptive fields). We can also directly classify pixels rather than superpixels, once we have computed $k$-dimensional 
feature vector for every \emph{pixel} in the image. In Section~\ref{sec:zoomfurther} we will discuss the results of directly 
classifying pixels with zoom-out features.

\subsection{Learning to label with asymmetric loss}
Once we have computed the zoom-out features we simply concatenate them
into a feature vector representing a superpixel. For superpixel $s$ in
image $\mathcal{I}$, we will denote this feature vector as
\[\bfgreek{phi}_{\text{zoom-out}}(s,\mathcal{I})\,=\,\left[
\bfgreek{phi}_1(s,\mathcal{I})\,\ldots,\bfgreek{phi}_L(s,\mathcal{I})\right]
\]
where $L$ is the number of levels in the zoom-out architecture.
For the training data, we will associate a single category label $y_s$
with
each superpixel $s$. This decision carries some risk, since in any non-trivial
over-segmentation some of the superpixels will not be perfectly
aligned with ground truth boundaries. In Section~\ref{sec:expzoom} we
evaluate this risk empirically for our choice of superpixel settings
and confirm that it is indeed minimal.

Now we are ready to train a classifier that maps $s$ in image
$\mathcal{I}$ to $y_s$ based on $\bfgreek{phi}_{\text{zoom-out}}$;
this requires choosing the empirical loss function to be minimized,
subject to regularization.
In semantic
segmentation settings, a factor that must impact this choice is the highly imbalanced nature of the labels. Some categories are
much more common than others, but our goal (encouraged by the
way benchmarks like VOC evaluate segmentations) is to predict them
equally well. It is well known that training on imbalanced data
without taking precautions can lead to
poor
results~\cite{farabet2013learning,pinheiro2014recurrent,lempitsky2011pylon}. A
common way to deal with this is to stratify the
training data; in practice this means that we throw away a large
fraction of the data corresponding to the more common
classes. We follow an alternative which we find less wasteful, and
which in our experience often produces dramatically better results:
use all the data, but change the loss. There has been some work on
loss design for learning segmentation~\cite{tarlow2012structured}, but
the simple weighted loss we describe below has to our knowledge been missed in
segmentation literature, with the exception
of~\cite{kuettel2012segmentation} and~\cite{lempitsky2011pylon}, where
it was used for binary segmentation.

Let the frequency of class $c$ in the
training data be
$f_c$, with $\sum_cf_c=1$. Our choice of loss is log-loss. We
scale it by the inverse frequency of
each class, effectively giving each pixel of less frequent classes
more importance:
\begin{equation}
  \label{eq:asymloss}
  -\frac{1}{N}\sum_{i=1}^N\frac{1}{f_{y_i}}\log \widehat{p}\left(y_i|\bfgreek{phi}(s_i,\mathcal{I}_i)\right),
\end{equation}
where $\widehat{p}\left(y_i|\bfgreek{phi}(s_i,\mathcal{I}_i)\right)$
is the estimated probability of the correct label for segment $s_i$ in
image $\mathcal{I}_i$, according to our model. 
 The loss in~\eqref{eq:asymloss} is still convex, and
only requires minor changes in implementation.

\section{Experiments}\label{sec:expzoom}

Our main set of experiments focuses on the PASCAL VOC category-level segmentation
benchmark with 21 categories, including the catch-all background
category. VOC is widely considered to be one of the main semantic
segmentation benchmarks. The original data set labeled with segmentation ground truth
consists of {\tt train}
and {\tt val} portions
(about 1,500 images in each). Ground
truth labels for additional 9,118 images have been provided by authors
of~\cite{BharathICCV2011}, and are commonly used in training segmentation models.
In all experiments below, we used the combination of these additional images with the original
{\tt train}
set for training, and {\tt val} was used only as held out validation set, to tune
parameters and to perform ablation studies.

The
main measure of success is accuracy on the test, which for VOC 2012
consists of 1,456 images. No ground truth is available for test, and
accuracy on it can only be obtained by uploading predicted
segmentations to the evaluation server.
The standard evaluation measure for category-level segmentation in VOC
benchmarks is per-pixel accuracy, defined as Jaccard index or intersection of the
predicted and true sets of pixels for a given class $c$, divided by their
union (IoU in short).

\begin{equation}
  \label{eq:asymloss}
    IoU_c(\hat{y},y) = \frac{|\lbrace \hat{y}=c \rbrace \cap  \lbrace y=c \rbrace| }{| \lbrace \hat{y}=c \rbrace \cup  \lbrace y=c \rbrace|}
\end{equation}
\\
in which $y$ is provided ground truth labels and $\hat{y}$ is predicted labels. This is averaged across the 21 classes to provide a single accuracy
number, mean IoU, usually used to measure overall performance of a method.

\subsection{Zoom-out feature extraction}
We obtained roughly 500 SLIC superpixels~\cite{achanta_pami12} per
image (the exact number varies per image), with
the parameter $m$ that controls the tradeoff between spatial and color
proximity set to 15, producing superpixels which tend to be of uniform
size and regular shape, but adhere to local
boundaries when color evidence compels it. This results
in average superpixel region of 450 pixels.

One way to compute zoom-out features is to set the number and scope of
zoom-out levels manually, and use hand-crafted features to
describe these levels. However, as we described in Section~\ref{sec:zoommethod}, we can utilize CNNs to
provide both the feature values and the zoom-out levels. As it is shown in Table~\ref{tab:manvsconv} the latter leads to a simpler implementation and superior performance comparing to
zoom-out manual.

\subsubsection{Manual zoom-out}

To represent a superpixel we use a number of well known features as well
as a small set of learned features.

\begin{description}
\item[Color]
We compute histograms separately for each of the three L*a*b color
channels, using 32 as well as 8 bins, using equally spaced binning;
this yields 120 feature dimensions.  We also compute the entropy of
each 32-bin histogram as an additional scalar feature (+3 dimensions). Finally, we
also re-compute histograms using adaptive binning, based on observed quantiles
in each image (+120 dimensions).

\item[Texture]
Texture is represented by histogram of texton assignments, with
64-texton~\cite{shotton_ijcv09,textonboost_eccv06} dictionary computed over a sampling of images from the
training set. This histogram is augmented with the value of its
entropy. In total there are 65 texture-related channels.

\item[SIFT] A richer set of features capturing appearance
is based on ``bag of words'' representations computed over SIFT~\cite{sift} descriptors. The descriptors are computed over a regular grid (every 8
pixels), on 8- and 18-pixel patches, separately for each L*a*b
channel. All the descriptors are assigned to a dictionary of 500
visual words. Resulting assignment histograms are averaged for two
patch sizes in each channel, yielding a total of 1500 values, plus 6
values for the entropies of the histograms.

\item[Location] Finally, we encode superpixel's location, by
computing its image-normalized coordinates relative to the center as
well as its shift from the center (the absolute value of the
coordinates); this produces four feature values.

\item[Local CNN] Instead of using hand-crafted features we
could learn a representation for the superpixel using a CNN. We
trained a network with 3 convolutional (conv) + pooling + RELU layers,
with 32, 32 and 64 filters respectively, followed by two
fully connected layers (1152 units each) and finally a softmax layer. The input to this
network is the bounding box of the superpixel, resized to 25$\times$25
pixels and padded to $35\times35$, in L*a*b color space. The filter
sizes in all layers are 5$\times$5; the pooling layers all have 3$\times$3
receptive fields, with stride of 2. We trained the network using
back-propagation, with the objective of minimizing log-loss for
superpixel classification. The output of the softmax layer of this
network is used as a 21-dimensional feature vector. Another
network with the same architecture was trained on binary
classification of foreground vs. background classes; that gives us two
more features.
\end{description}


\subsubsection{Proximal features}
We use the same set of hand-crafted features as for local regions, but average them over superpixels in proximal region around each superpixel. This results in 1818 feature dimensions.

\subsubsection{Distant and scene features}
For distant and scene features we use deep CNNs originally
trained to classify images. In our initial experiments we used the CNN-S network
in~\cite{simonyan2014return}. It has 5 convolution layers
(three of them followed by pooling) and two fully connected layers. To extract the relevant features, we
resize either the distant region or the entire image to
224$\times$224 pixels, feed it to the network, and record the
activation values of the last fully-connected layer with 4096 units.
In a subsequent set of experiments we switched from CNN-S to a 16
layer network introduced in~\cite{simonyan2014very}. This network,
which we refer to as VGG-16, contains more layers that apply
non-linear transformations, but with smaller filters, and thus may learn richer representation
with fewer parameters. It has
produced excellent results on image classification and recently has
been reported to lead to a much better performance when used in
detection~\cite{girshick2014rich}. As reported below, we also observe
a significant improvement when using VGG-16 to extract distant
and global features, compared to performance with CNN-S.
    Note that both networks we used were originally trained for 1000-category ImageNet
classification task, and we did not fine-tune it in any way on VOC data.

\subsubsection{CNN-based zoom-out}

Given a CNN, we associate a zoom-out level with every
convolutional layer in the network. For the following experiments, we use
the 16-layer network from~\cite{simonyan2014very}, with the final
classification layer removed. This network, which we refer to as VGG-16, has 13
convolutional layers, yielding 13 zoom-out levels (see
Fig.~\ref{fig:zo}). In Section~\ref{sec:zoomfurther} we apply zoom-out to modern backbones such as ResNet~\cite{ResNet}, DenseNet~\cite{DenseNet} and NASNet~\cite{Nasnet} .

\begin{table}[!t]
  \centering
  \begin{tabular}[c]{|l|l|l|l|l|}
    \hline
    Group &Level & Dim & Unit RF size & Region size\\
    \hline
    G1 & 1 & 64 & 3 & 32\\
    G1 & 2 & 64 & 5 & 36\\
    G2 &3 & 128 & 10 & 45\\
    G2 &4 & 128 & 14 & 52\\
    G3 & 5 & 256 & 24 & 70\\
    G3 &6 & 256 & 32 & 84\\
    G3& 7 & 256 & 40 & 98\\
    G4 & 8 & 512 & 60 & 133\\
    G4 & 9 & 512 & 76 & 161\\
    G4& 10 & 512 & 92 & 190\\
    G5 & 11 & 512 & 132 & 250\\
    G5& 12 & 512 & 164 & 314 \\
    G5&13 & 512 & 196 & 365\\
    \hline
    S1&subscene & 4096 & -- & 130\\
    S2&scene & 4096 & -- & varies\\
    \hline
  \end{tabular}
  \caption{Statistics of the zoom-out features induced by the 16 layer
    CNN. Dim: dimension of feature vector. Unit RF size is size of
    receptive field of a CNN unit in pixels of
    $256\times256$ input image. Region size: average size of
    receptive field of the zoom-out feature in pixels of original VOC
    images. The group designation is referred to by design of
    experiments in Section~\ref{sec:ablation}.}
  \label{tab:feat}
\end{table}

To compute sub-scene level features, we take the bounding box of
superpixels within radius three from the superpixel at hand (i.e.,
neighbors, their neighbors and their neighbors' neighbors). This
bounding box is warped to canonical resolution of $256\times256$
pixels, and fed to the CNN; the activations of the last fully
connected layer are the window level features.
Finally, the scene level features are computed by feeding the entire image,
again resized to canonical resolution of $256\times256$ pixels, to the
CNN, and extracting the activations of the last fully connected
layer. 

Feature parameters are summarized in
Table~\ref{tab:feat}. Concatenating all the features yields a
12,416-dimensional representation for a superpixel. Following common practice, we also extract the features at all levels
from the mirror image (left-right reflection, with the superpixels
mirrored as well), and take element-wise max over the resulting two
feature vectors.

\begin{table}[!th]
  \centering
{\small
  \begin{tabular}{|l|l|}
    \hline
Method & mean IoU\\
\hline
manual zoom-out CNN-S &  57.1\\
manual zoom-out  VGG-16 & 63.5\\
\hline
CNN-based zoom-out VGG-16 & 69.9\\
\hline
  \end{tabular}}
  \caption{comparison of setting the number and scope of
zoom-out levels manually versus CNN-based zoom-out.}
  \label{tab:manvsconv}
\end{table}



\subsection{Learning setup}

To rule out significant loss of accuracy due to reduction of image
labeling to superpixel labeling, we evaluated the achievable accuracy
under ``oracle'' labeling. Assigning each superpixel a category label based on the majority vote by
pixels in it produces mean IoU of 94.4\% on VOC 2012 {\tt val}. Thus, we can assume that loss of accuracy due to our
commitment to superpixel boundaries is going to play a minimal role
in our results.

With more than 10,000 images and roughly 500 superpixels per image, we
have more than 5 million training examples. 
We trained various classifiers on this data, with asymmetric
log-loss~\eqref{eq:asymloss}, using Caffe~\cite{jia2014caffe} on a single machine equipped with a Tesla K40
GPU. During training we used fixed learning rate of 0.0001, and
weight decay factor of 0.001.

\begin{table}[!th]
  \centering
{\small
  \begin{tabular}{|l|l|l|}
    \hline
Region type & oracle mIoU & zoom-out mIoU\\
\hline
SLIC-500 &  94.4& 69.9\\
\hline
rectangles-500 & 87.2 & 64.3\\
\hline
  \end{tabular}}
  \caption{Effect of replacing SLIC superpixels with an equal number of rectangular regions. Results on VOC 2012  {\tt val}, mean class IoU.}
  \label{tab:SLIC}
\end{table}

\subsection{Analysis of contribution of zoom-out levels}\label{sec:ablation}
To assess the importance of features extracted
at different zoom-out levels, we experimented with various feature
subsets, as shown in Table~\ref{tab:ablation}. For each subset, we
train a linear (softmax) classifier on VOC 2012 {\tt train} and
evaluate performance on VOC 2012 {\tt val}. The feature set designations refer to the groups
listed in Table~\ref{tab:feat}.

 It is evident that each of the zoom-out
levels contributes to the eventual accuracy. Qualitatively, we observe
that complex features computed at sub-scene and scene levels play a
role in establishing the right set of labels for an image, while
features derived from convolutional layers of the CNN are important in localization of object boundaries; a few
examples in Figure~\ref{fig:levels} illustrate this.
We also confirmed empirically that learning with asymmetric loss leads to
dramatically better performance compared to standard, symmetric loss
(and no data balancing). 

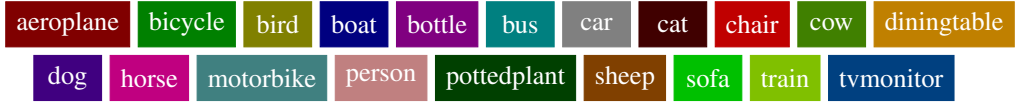
\begin{figure*}[!h]
  \centering
\begin{tikzpicture}[every node/.style={font=\footnotesize}]
  \definecolor{aeroplane-co}{rgb}{0.502,0,0}
    \node[text=white,fill=aeroplane-co,draw=aeroplane-co,minimum
    height=0.6cm,minimum width=0.9cm] (aeroplane) {aeroplane};

    \definecolor{bicycle-co}{rgb}{0,0.502,0}    
    \node[text=white,fill=bicycle-co,draw=bicycle-co,minimum
    height=0.6cm,minimum width=0.9cm,right=of aeroplane,right=.1cm] (bicycle) {bicycle};

    \definecolor{bird-co}{rgb}{0.502,0.502,0}    
    \node[text=white,fill=bird-co,draw=bird-co,minimum
    height=0.6cm,minimum width=0.9cm,right=of bicycle,right=.1cm] (bird) {bird};

    \definecolor{boat-co}{rgb}{0,0,0.502}    
        \node[text=white,fill=boat-co,draw=boat-co,minimum
    height=0.6cm,minimum width=0.9cm,right=of bird,right=.1cm] (boat) {boat};

    \definecolor{bottle-co}{rgb}{0.502,0,0.502}    
    \node[text=white,fill=bottle-co,draw=bottle-co,minimum
    height=0.6cm,minimum width=0.9cm,right=of boat,right=.1cm] (bottle) {bottle};
    
    \definecolor{bus-co}{rgb}{0,0.502,0.502}    
    \node[text=white,fill=bus-co,draw=bus-co,minimum
    height=0.6cm,minimum width=0.9cm,right=of bottle,right=.1cm] (bus) {bus};

    \definecolor{car-co}{rgb}{0.502,0.502,0.502}    
    \node[text=white,fill=car-co,draw=car-co,minimum
    height=0.6cm,minimum width=0.9cm,right=of bus,right=.1cm] (car)
    {car};

    \definecolor{cat-co}{rgb}{0.251,0,0}    
    \node[text=white,fill=cat-co,draw=cat-co,minimum
    height=0.6cm,minimum width=0.9cm,right=of car,right=.1cm] (cat)
    {cat};

    \definecolor{chair-co}{rgb}{0.753,0,0}    
    \node[text=white,fill=chair-co,draw=chair-co,minimum
    height=0.6cm,minimum width=0.9cm,right=of cat,right=.1cm] (chair)
    {chair};

    \definecolor{cow-co}{rgb}{0.251,0.502,0}    
    \node[text=white,fill=cow-co,draw=cow-co,minimum
    height=0.6cm,minimum width=0.9cm,right=of chair,right=.1cm] (cow)
    {cow};

	\definecolor{diningtable-co}{rgb}{0.753,0.502,0}    
    \node[text=white,fill=diningtable-co,draw=diningtable-co,minimum
    height=0.6cm,minimum width=0.9cm,right=of cow,right=.1cm] (diningtable)
    {diningtable};


    \definecolor{dog-co}{rgb}{0.251,0,0.502}    
    \node[text=white,fill=dog-co,draw=dog-co,minimum
    height=0.6cm,minimum width=0.9cm,below=of aeroplane,below=.1cm] (dog)
    {dog};

    \definecolor{horse-co}{rgb}{0.753,0,0.502}    
    \node[text=white,fill=horse-co,draw=horse-co,minimum
    height=0.6cm,minimum width=0.9cm,right=of dog,right=.1cm] (horse)
    {horse};

    \definecolor{motorbike-co}{rgb}{0.251,0.502,0.502}    
    \node[text=white,fill=motorbike-co,draw=motorbike-co,minimum
    height=0.6cm,minimum width=0.9cm,right=of horse,right=.1cm] (motorbike)
    {motorbike};

    \definecolor{person-co}{rgb}{0.753,0.502,0.502}    
    \node[text=white,fill=person-co,draw=person-co,minimum
    height=0.6cm,minimum width=0.9cm,right=of motorbike,right=.1cm] (person)
    {person};

    \definecolor{pottedplant-co}{rgb}{0,0.251,0}    
    \node[text=white,fill=pottedplant-co,draw=pottedplant-co,minimum
    height=0.6cm,minimum width=0.9cm,right=of person,right=.1cm] (pottedplant)
    {pottedplant};

    \definecolor{sheep-co}{rgb}{0.502,0.251,0}    
    \node[text=white,fill=sheep-co,draw=sheep-co,minimum
    height=0.6cm,minimum width=0.9cm,right=of pottedplant,right=.1cm] (sheep)
    {sheep};

    \definecolor{sofa-co}{rgb}{0,0.753,0}    
    \node[text=white,fill=sofa-co,draw=sofa-co,minimum
    height=0.6cm,minimum width=0.9cm,right=of sheep,right=.1cm] (sofa)
    {sofa};

    \definecolor{train-co}{rgb}{0.502,0.753,0}    
    \node[text=white,fill=train-co,draw=train-co,minimum
    height=0.6cm,minimum width=0.9cm,right=of sofa,right=.1cm] (train)
    {train};

   \definecolor{tvmonitor-co}{rgb}{0,0.251,0.502}    
    \node[text=white,fill=tvmonitor-co,draw=tvmonitor-co,minimum
    height=0.6cm,minimum width=0.9cm,right=of train,right=.1cm] (tvmonitor)
    {tvmonitor};

\end{tikzpicture}
  \caption{Color code for VOC categories. Background is black.}
  \label{fig:colorcode}
\end{figure*}
\begin{figure*}[!h]
  \centering
  \begin{tabular}{cccccc}
    \includegraphics[width=.13\textwidth]{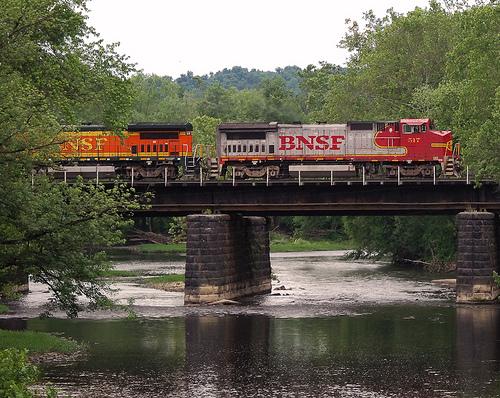} &
    \includegraphics[width=.14\textwidth]{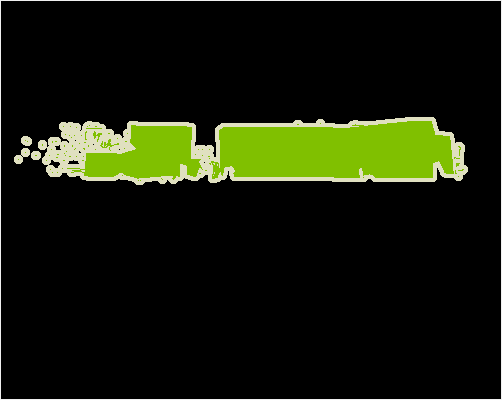} &
    \includegraphics[width=.14\textwidth]{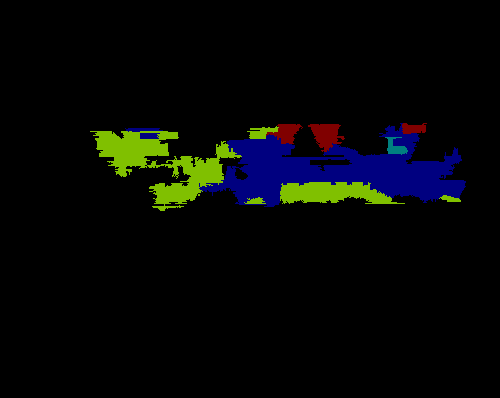} &
    \includegraphics[width=.14\textwidth]{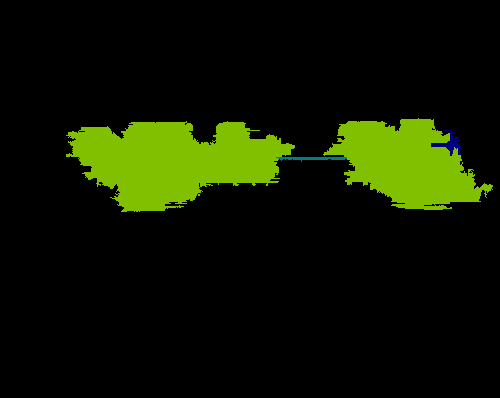} &
    \includegraphics[width=.14\textwidth]{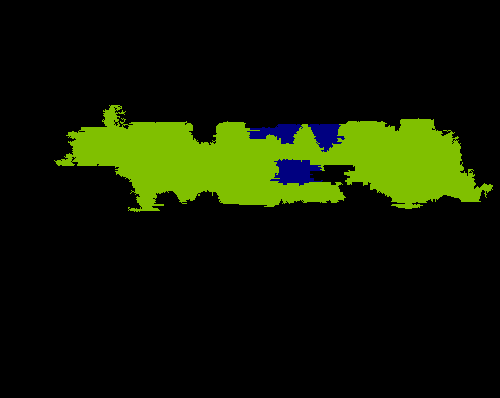} &
    \includegraphics[width=.14\textwidth]{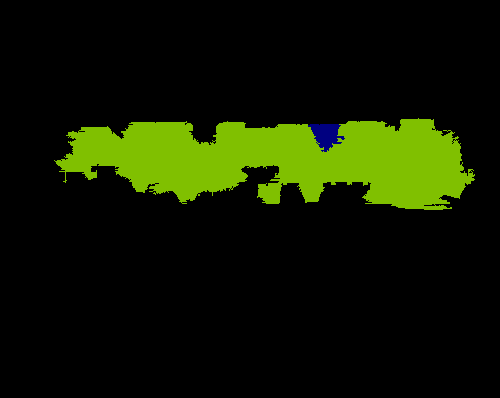}\\

    \includegraphics[width=.14\textwidth]{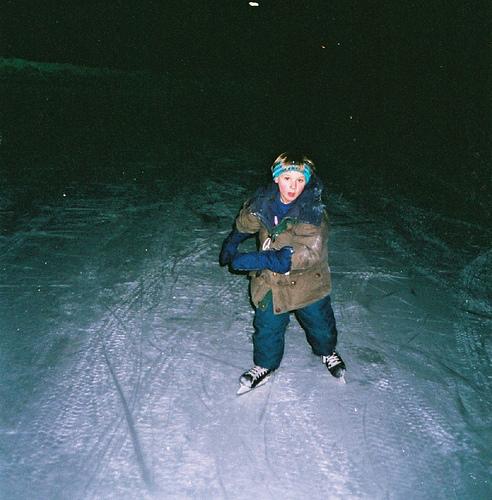} &
    \includegraphics[width=.14\textwidth]{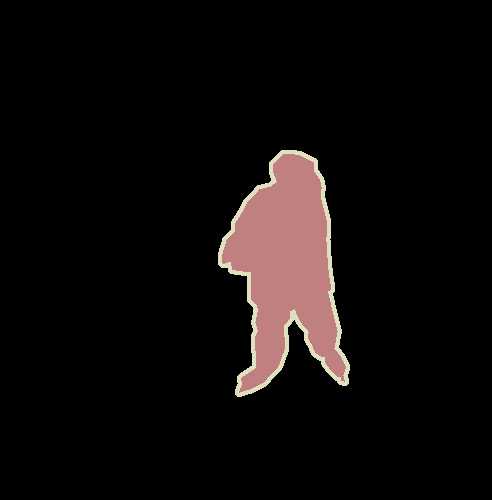} &
    \includegraphics[width=.14\textwidth]{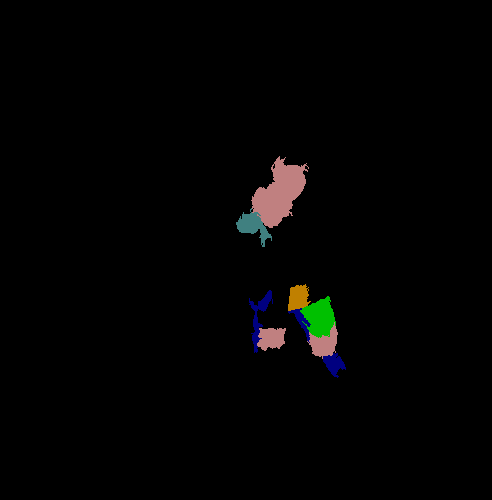} &
    \includegraphics[width=.14\textwidth]{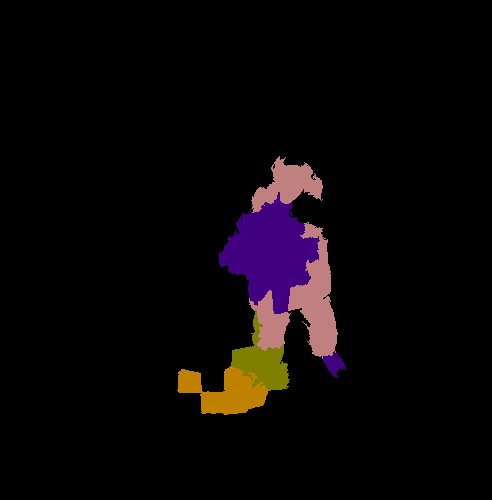} &
    \includegraphics[width=.14\textwidth]{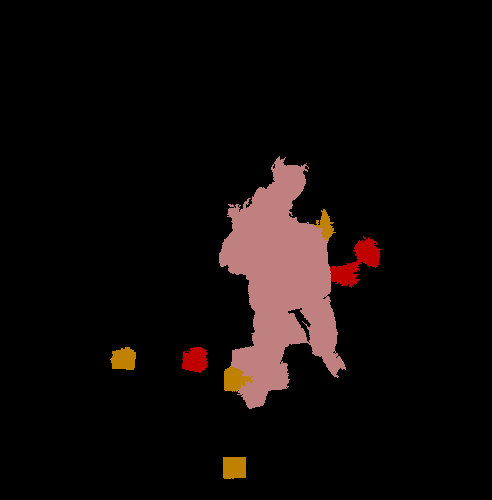} &
    \includegraphics[width=.14\textwidth]{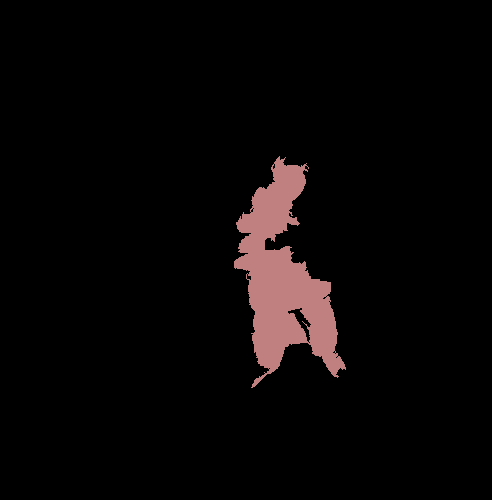}\\
    \includegraphics[width=.14\textwidth]{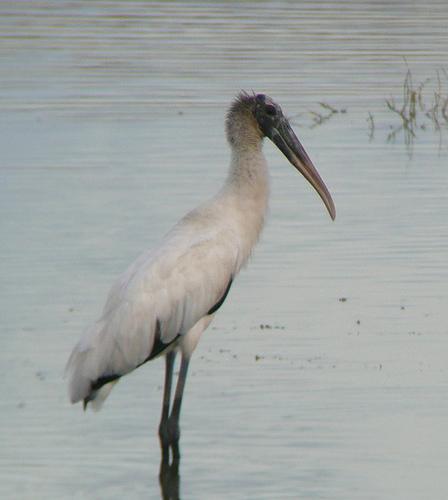} &
    \includegraphics[width=.14\textwidth]{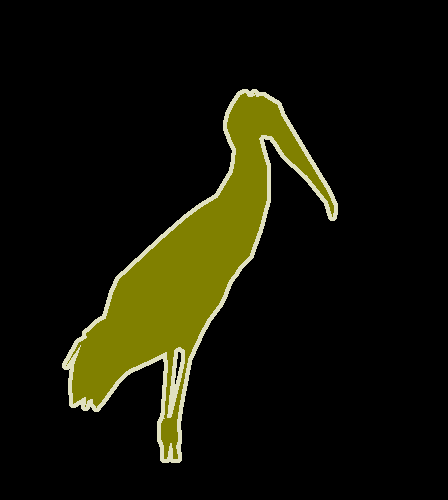} &
    \includegraphics[width=.14\textwidth]{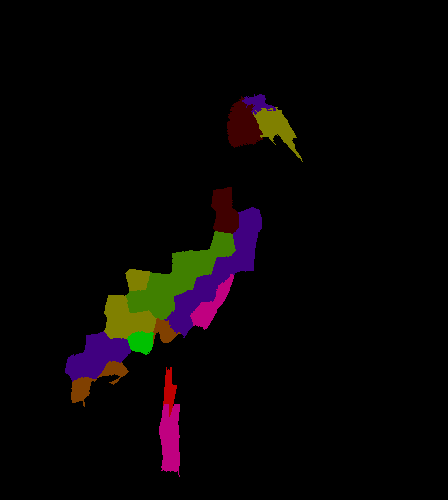} &
    \includegraphics[width=.14\textwidth]{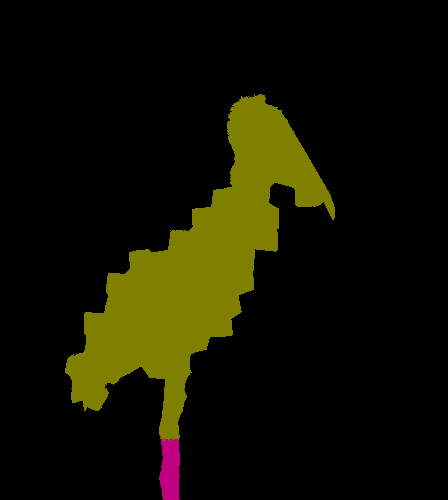} &
    \includegraphics[width=.14\textwidth]{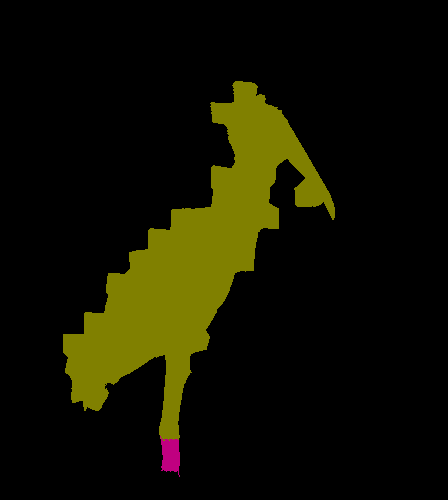} &
    \includegraphics[width=.14\textwidth]{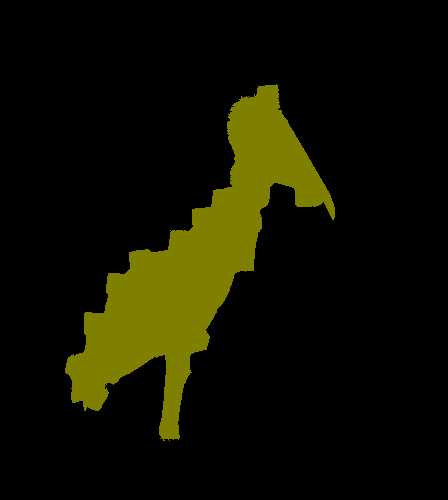}
  \end{tabular}
  \caption{Examples illustrating the effect of zoom-out levels. From
    left: original image, ground truth, levels G1:3,
    G1:5, G1:5+S1, and the full
    set of zoom-out features G1:5+S1+S2. In all cases a linear model is used to
    label superpixels. See Figure~\ref{fig:colorcode} for category
    color code, and Table~\ref{tab:feat} for level notation.}
  \label{fig:levels}
\end{figure*}

\begin{table}[!th]
  \centering
{\small
  \begin{tabular}{|l|l|}
    \hline
Feature set & mean IoU\\
\hline
G1 &   \hspace{0.2em} 6.0\\
G1-2 & 10.1\\
G1-3 & 16.3\\
G1-4 & 26.3\\
G1-5& 41.8\\
\hline
G1-5+S1& 51.2\\
G1-5+S2 & 57.3\\
G4-5+S1+S2 & 58.0\\
\hline
full zoom-out: G1-5+S1+S2 & \textbf{58.6}\\
\hline
  \end{tabular}}
  \caption{Ablation study: importance of features from different
    levels under linear superpixel classification. Results on VOC 2012
    {\tt val}, mean class IoU.}
  \label{tab:ablation}
\end{table}

Next we explored the impact of switching from linear softmax models to
multilayer neural networks, evaluating a sequence of models on VOC
2012 {\tt val}. Introducing a single layer of 1024
hidden units, with RELU nonlinearity, increased IoU from 58.6 to 68.4;
additional hidden units (1500 or 2048) didn't increase it
further. Adding another layer with 1024 units, and introducing
dropout~\cite{srivastava2014dropout} improved IoU to 69.9, and this is
the model we adopt for final evaluation on the test set. The test results are shown in Table~\ref{tab:voc-old}.

\begin{table}[!th]
  \centering
  \begin{tabular}{|>{\small}l||l|l||l|}
    \hline
Method & {\small VOC2010} & {\small VOC2011} & {\small VOC2012}\\
\hline
zoom-out (ours), CVPR15 & {\bf 69.9} & {\bf 69.4} & {\bf 69.6}\\
\hline
FCN-8s, CVPR15~\cite{long2014fully} & -- & 62.7 & 62.2\\ 
Hypercolumns, CVPR15~\cite{hariharan2014hypercolumns} & -- & -- & 59.2\\
DivMbest+CNN, arXiv14~\cite{dhruv-new} & -- & -- & 52.2\\
SDS, ECCV14~\cite{hariharan2014simultaneous} & -- & 52.6 & 51.6\\
DivMbest+rerank, CVPR13~\cite{yadollahpour2013discriminative} & --&--& 48.1\\
Codemaps, ICCV13~\cite{Li2013codemaps} & --&-- &48.3\\
$O2P$, ECCV12~\cite{carreira2012semantic} & -- & 47.6 & 47.8\\
Regions \& parts, CVPR12~\cite{arbelaez_cvpr12} & -- & 40.8 & --\\
Harmony potentials, IJCV12~\cite{boix_ijcv12} & 40.1 & -- & --\\
D-sampling, ICCV11~\cite{lucchi2011spatial} & 33.5 & -- & --\\
\hline
  \end{tabular}
  \caption{Results on VOC 2010, 2011 and 2012 test. Mean IoU is shown, see Table~\ref{tab:details-2012} for per-class
    accuracies of the zoom-out method.}
  \label{tab:voc-old}
\end{table}


To evaluate importance of our reliance on superpixels we also
evaluated an architecture in which SLIC superpixels are replaced by an
equal number of rectangular regions. The achievable accuracy on VOC
2012 {\tt val} with
this over-segmentation is 87.2, compared to 94.4 with superpixels. The
difference is due to failure of the rectangular grid to adhere to
boundaries around thin structures or fine shape elements. Similar gap
persists when we apply the full zoom-our architecture to the
rectangular regions instead of superpixels: we get mean IoU of 64.3,
more than 5 points below the result with superpixels. These results are summarized in Table~\ref{tab:SLIC}. In Section~\ref{sec:zoomfurther} we run set of experiments in which individual pixels are
being classified rather than superpixels.

Finally, we investigated the effect of replacing VGG-16 from~\cite{simonyan2014very}
with a previously widely used 7-layer CNN referred to as CNN-S
in~\cite{simonyan2014return}. This net almost identical to the
celebrated AlexNet~\cite{krizhevsky2012imagenet}. Our experience is
consistent with previously reported results where the two networks
were compared as feature extractors on a variety of
tasks~\cite{girshick2014rich,simonyan2014very,long2014fully}: there is
a significant drop in performance when we use the full zoom-out
architecture using CNN-S compared to the zoom-out with VGG-16 (mean
IoU 45.4 vs. 58.6). 

\renewcommand{\arraystretch}{1}
\setlength{\tabcolsep}{0.5pt}
\begin{table*}[!th]
  \centering
{\small
  \begin{tabu}{|l|c||c|c|c|c|c|c|c|c|c|c|c|c|c|c|c|c|c|c|c|c|c|}
    \hline
    class &
    \rot{90}{\bf mean} &
    \rot{90}{bg} &
    \includegraphics[width=5mm]{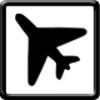}&
    \includegraphics[width=5mm]{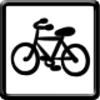}&
    \includegraphics[width=5mm]{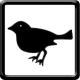}&
    \includegraphics[width=5mm]{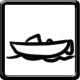}&
    \includegraphics[width=5mm]{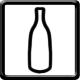}&
    \includegraphics[width=5mm]{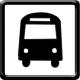}&
    \includegraphics[width=5mm]{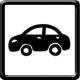}&
    \includegraphics[width=5mm]{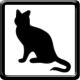}&
    \includegraphics[width=5mm]{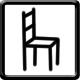}&
    \includegraphics[width=5mm]{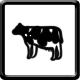}&
    \includegraphics[width=5mm]{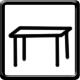}&
    \includegraphics[width=5mm]{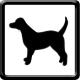}&
    \includegraphics[width=5mm]{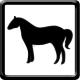}&
    \includegraphics[width=5mm]{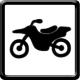}&
    \includegraphics[width=5mm]{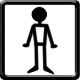}&
    \includegraphics[width=5mm]{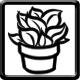}&
    \includegraphics[width=5mm]{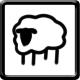}&
    \includegraphics[width=5mm]{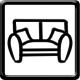}&
    \includegraphics[width=5mm]{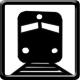}&
    \includegraphics[width=5mm]{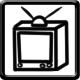}
    \\
    \hline
    \rowfont{\footnotesize}
    {\normalsize acc} &
    69.6 &
    91.9 &
    85.6 & 37.3 & 83.2 & 62.5 & 66 &
    85.1 & 80.7 & 84.9 & 27.2 & 73.3 &
    57.5 & 78.1 & 79.2 & 81.1 & 77.1 &
    53.6 & 74 & 49.2 & 71.7 &63.3\\
    \hline
  \end{tabu}
}
\vspace{1em}\caption{Detailed results of our method on VOC 2012 test.}
  \label{tab:details-2012}
\end{table*}

\begin{figure*}[!th]
  \centering
  \begin{tabular}{cccccc}
\includegraphics[width=.16\textwidth]{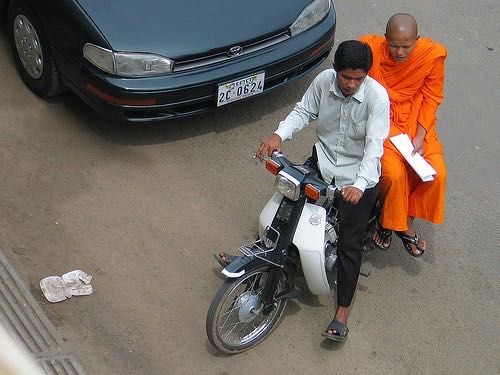} &
 \includegraphics[width=.16\textwidth]{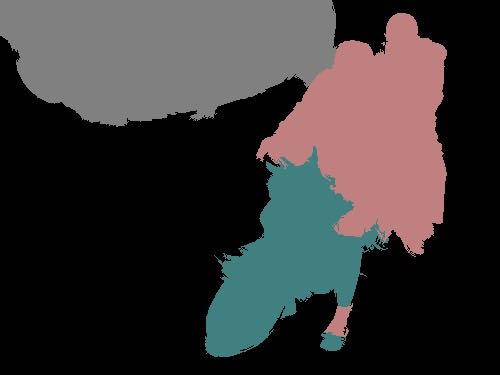}&
\includegraphics[width=.15\textwidth]{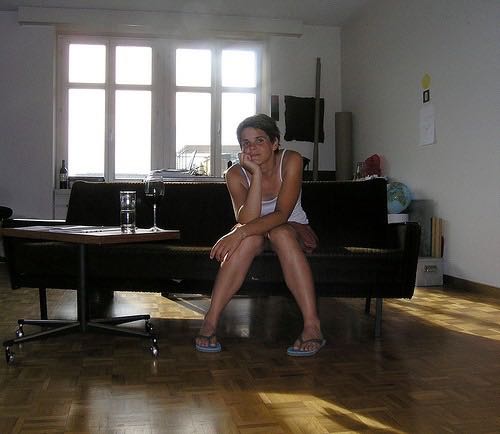} &
 \includegraphics[width=.15\textwidth]{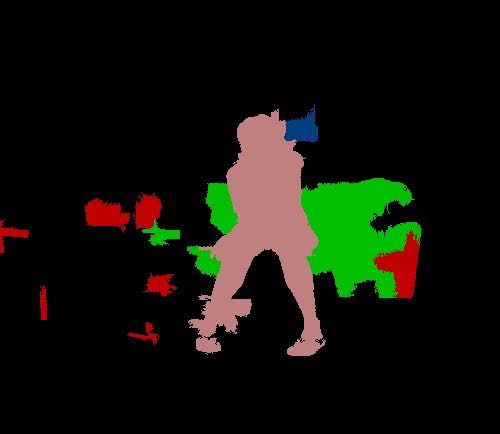}&
\includegraphics[width=.16\textwidth]{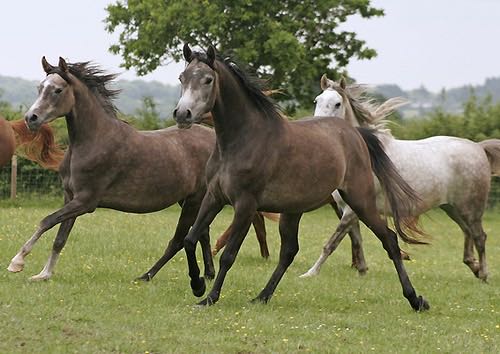} &
 \includegraphics[width=.16\textwidth]{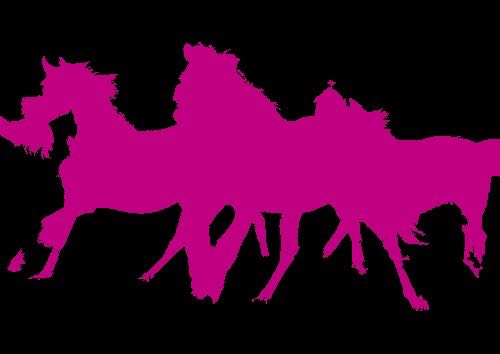}\\
\includegraphics[width=.16\textwidth]{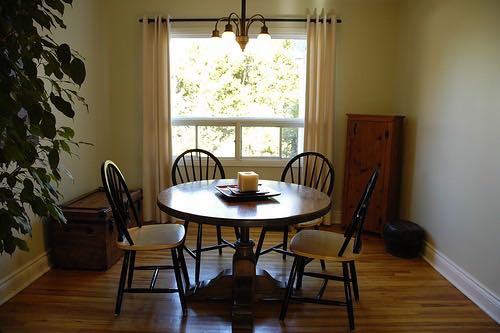} &
 \includegraphics[width=.16\textwidth]{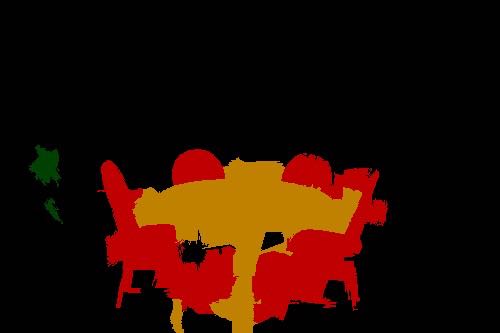}&
\includegraphics[width=.14\textwidth]{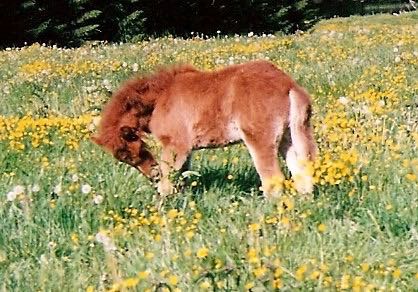} &
 \includegraphics[width=.14\textwidth]{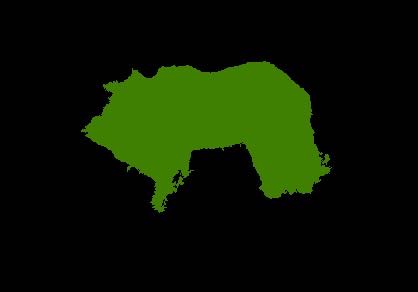}&
\includegraphics[width=.16\textwidth]{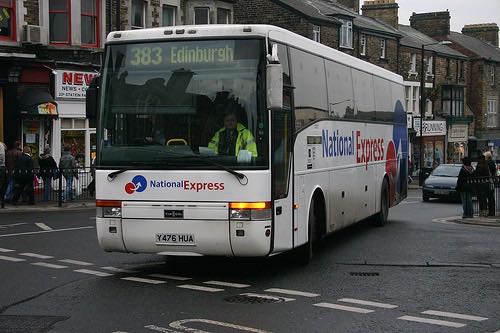} &
 \includegraphics[width=.16\textwidth]{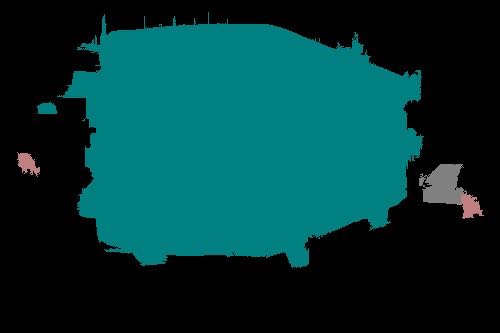}\\
\includegraphics[width=.10\textwidth]{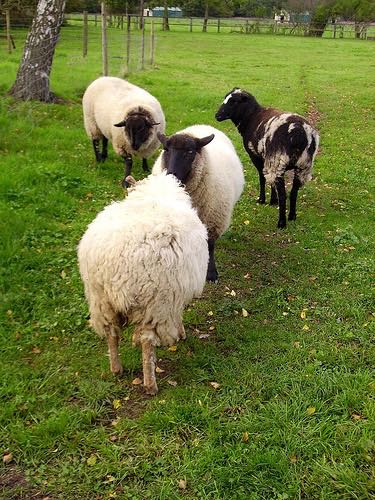} &
 \includegraphics[width=.10\textwidth]{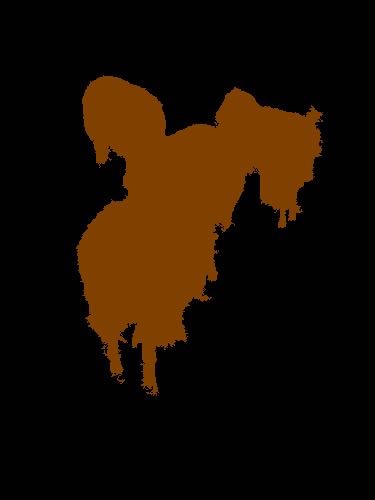}&
\includegraphics[width=.16\textwidth]{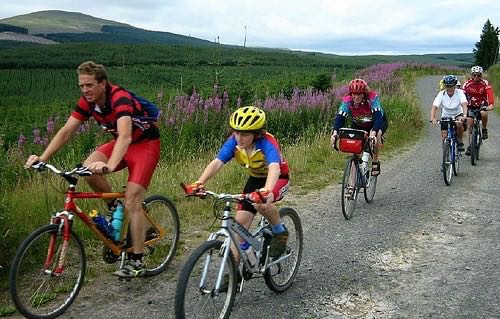} &
 \includegraphics[width=.16\textwidth]{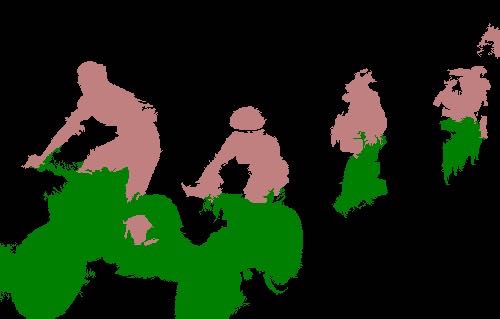}&
\includegraphics[width=.16\textwidth]{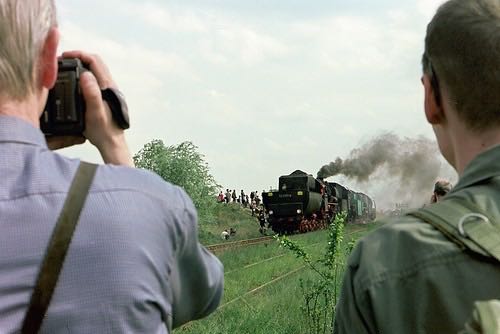} &
 \includegraphics[width=.16\textwidth]{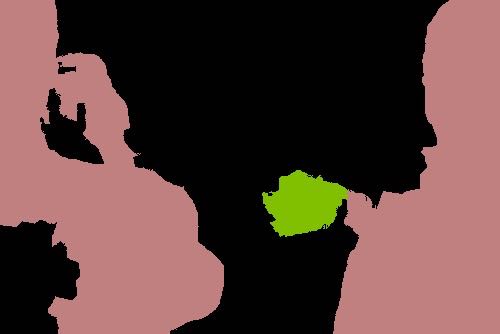}\\
\includegraphics[width=.16\textwidth]{figs/2007_003143.jpg} &
 \includegraphics[width=.16\textwidth]{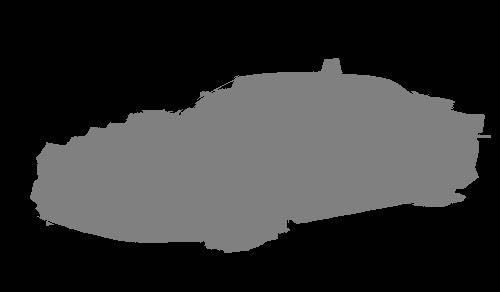}&
\includegraphics[width=.16\textwidth]{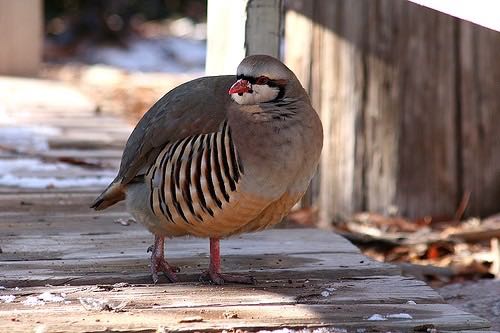} &
 \includegraphics[width=.16\textwidth]{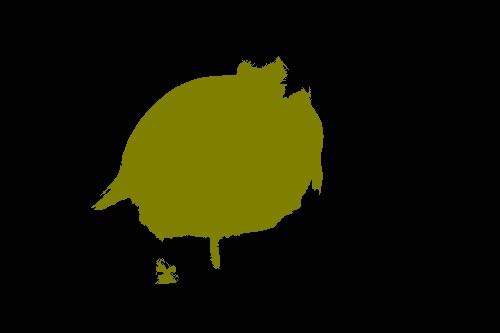}&
\includegraphics[width=.16\textwidth]{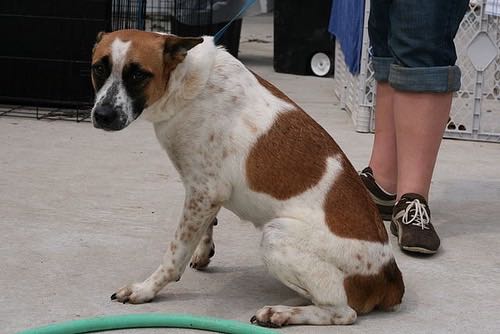} &
 \includegraphics[width=.16\textwidth]{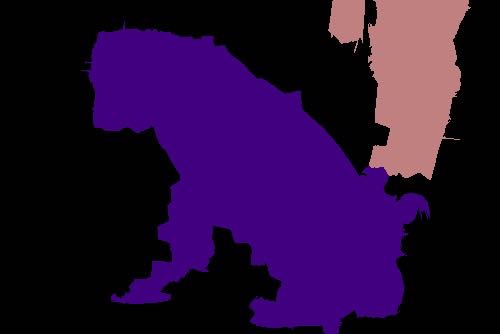}\\
\includegraphics[width=.16\textwidth]{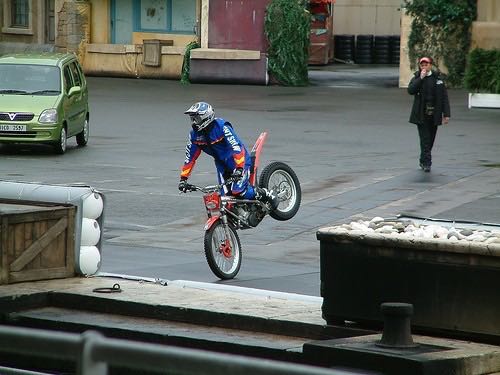} &
 \includegraphics[width=.16\textwidth]{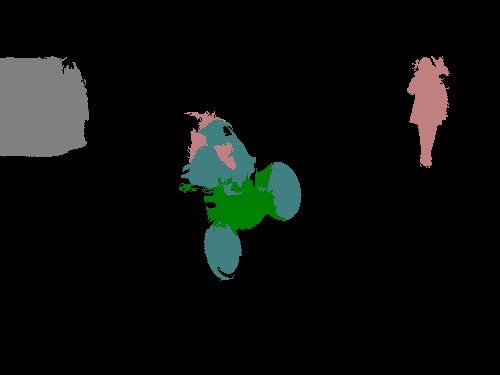}&
\includegraphics[width=.16\textwidth]{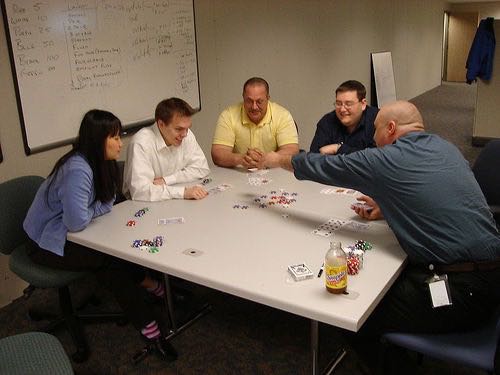} &
 \includegraphics[width=.16\textwidth]{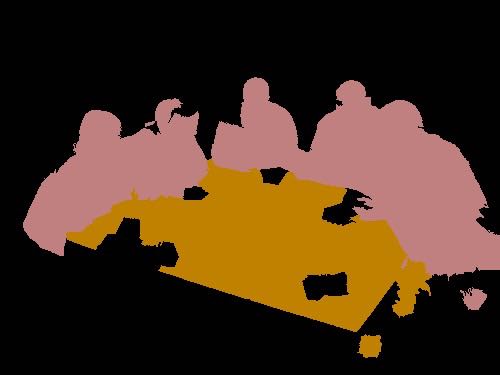}&
\includegraphics[width=.16\textwidth]{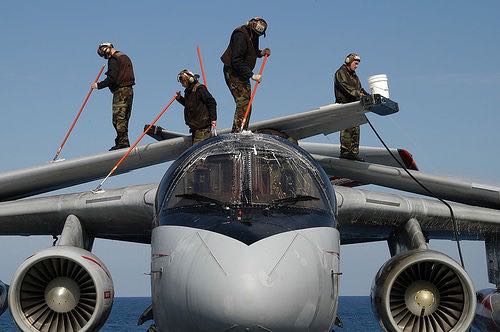} &
 \includegraphics[width=.16\textwidth]{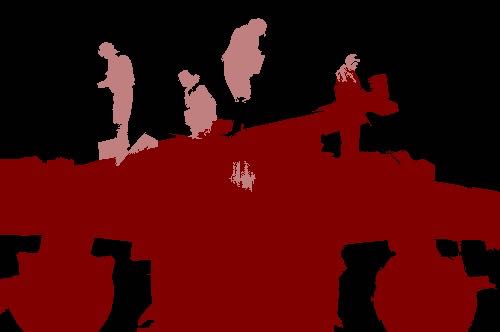}\\
 
\includegraphics[width=.16\textwidth]{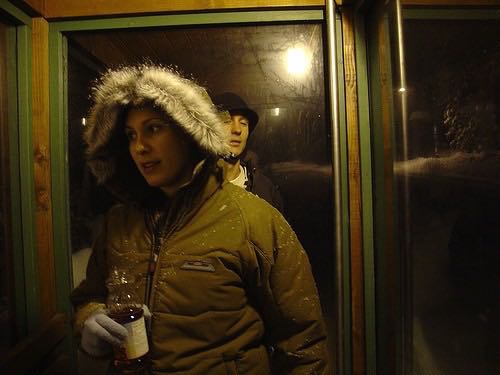} &
 \includegraphics[width=.16\textwidth]{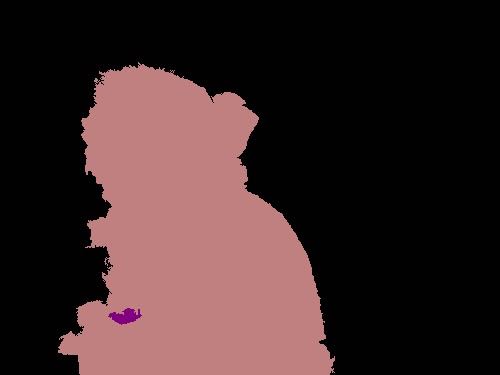}&
 \includegraphics[width=.18\textwidth]{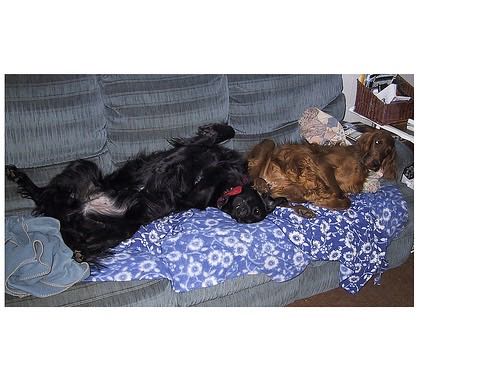} &
 \includegraphics[width=.16\textwidth]{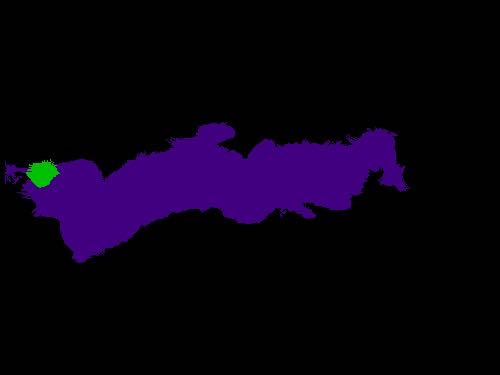}&
\includegraphics[width=.16\textwidth]{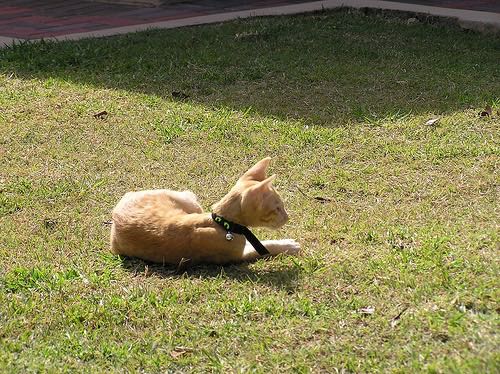}&
\includegraphics[width=.16\textwidth]{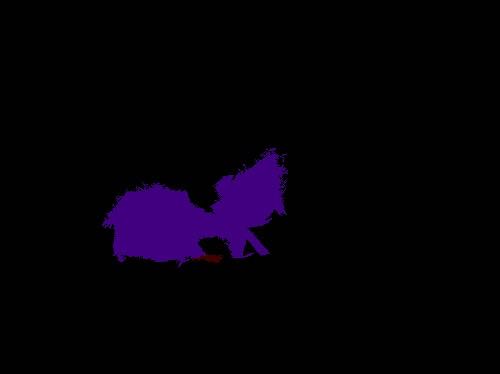}\\ 
 
\includegraphics[width=.15\textwidth]{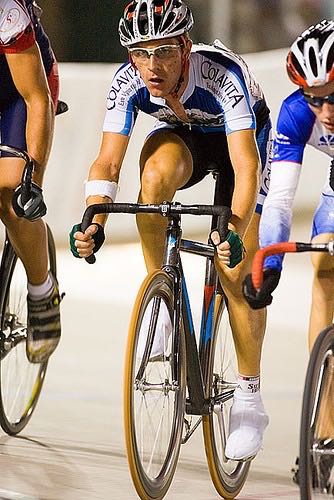} &
 \includegraphics[width=.15\textwidth]{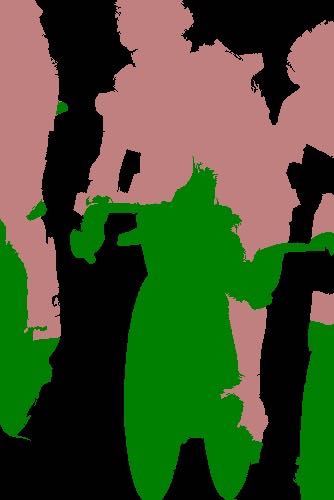}&
\includegraphics[width=.15\textwidth]{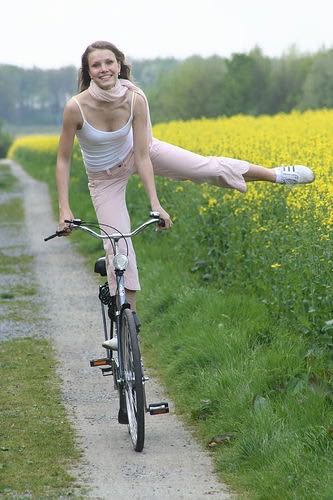} &
 \includegraphics[width=.15\textwidth]{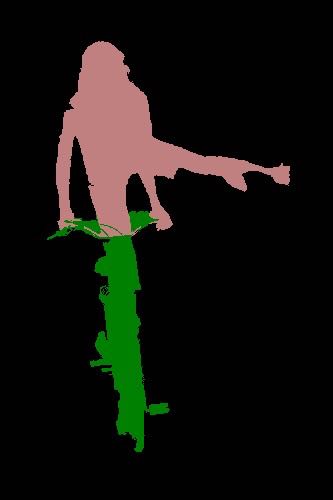}&
\includegraphics[width=.15\textwidth]{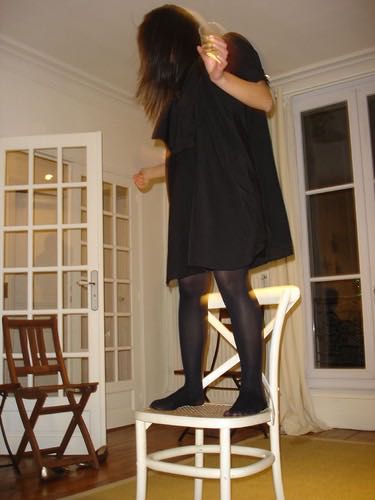} &
 \includegraphics[width=.15\textwidth]{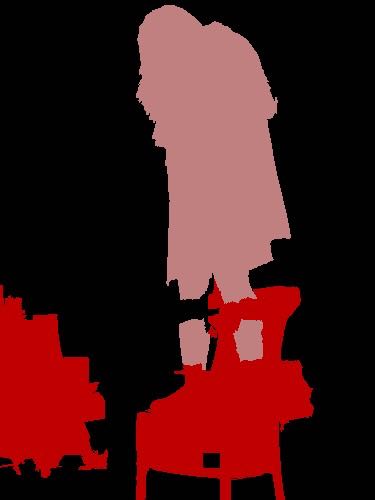}\\
 \includegraphics[width=.16\textwidth]{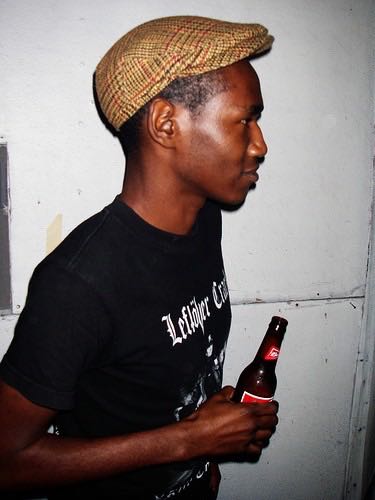} &
 \includegraphics[width=.16\textwidth]{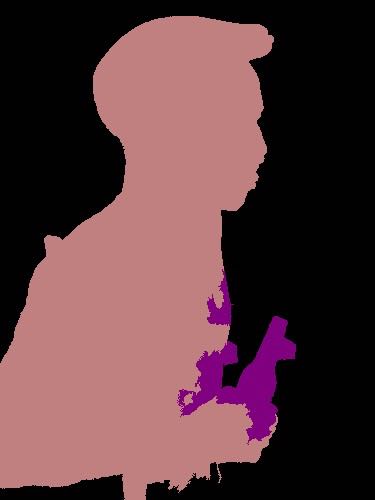}&
\includegraphics[width=.14\textwidth]{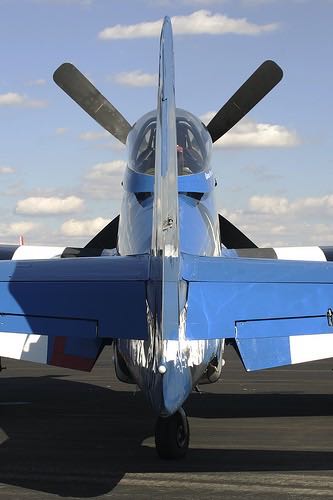} &
 \includegraphics[width=.14\textwidth]{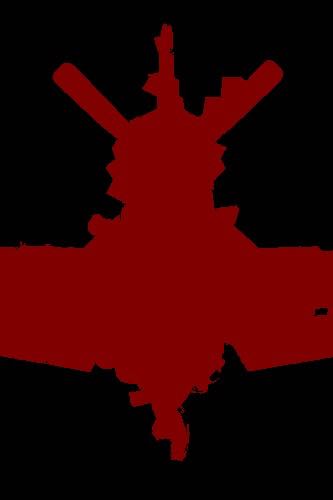}&
\includegraphics[width=.16\textwidth]{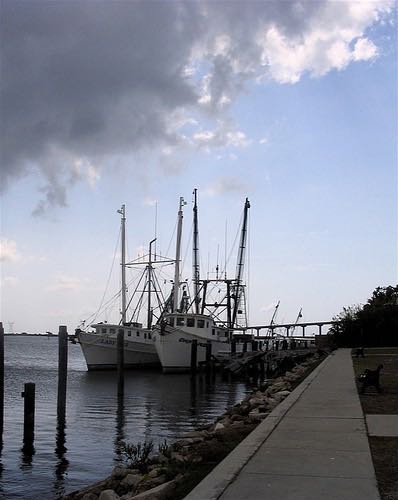}&
 \includegraphics[width=.16\textwidth]{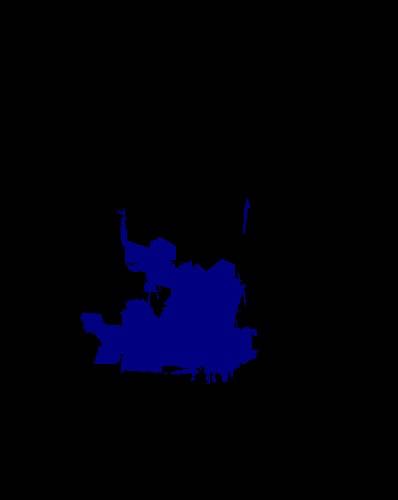}

  \end{tabular}
  \caption{Example segmentations on VOC 2012 val with 3-layer neural
    network used to classify full zoom-out representation of
    superpixels (15 zoom-out levels). See Figure~\ref{fig:colorcode}
    for category color code.}
  \label{fig:exampleszoomold}
\end{figure*}

Figure~\ref{fig:exampleszoomold} displays example
segmentations.
Many of the segmentations have moderate to high accuracy, capturing
correct classes, in correct layout, and sometimes including level of detail
that is usually missing from over-smoothed segmentations obtained by
CRFs or generated by region proposals. On the other hand, despite
the smoothness imposed by higher zoom-out levels, the segmentations we
get do tend to be \emph{under}-smoothed, and in particular include
little ``islands'' of often irrelevant categories. To some extent this
might be alleviated by post-processing; we found that we could learn a
classifier for isolated regions that with reasonable accuracy decides
when these must be ``flipped'' to the surrounding label, and this
improves results on {\tt val} by about 0.5\%, while making the segmentations
more visually pleasing. However, We do not pursue this ad-hoc approach, and
instead discuss in Chapter~\ref{chap:labelemb} more principled
remedies. 

\subsection{Results on Stanford Background Dataset}
For some of the closely related recent work results on VOC are not
available, so to allow for empirical comparison, we also ran an
experiment on Stanford Background Dataset (SBD). It has 715 images of
outdoor scenes, with
dense labels for eight categories. We applied the same zoom-out
architecture to this dataset as to VOC, except that the
classifier had only 128 hidden units (due to much smaller size of the
data set).

There is no standard train/test partition of SBD; the established
protocol calls for reporting 5-fold cross validation results. There is
also no single performance measure; two commonly reported measures are
per-pixel accuracy and average class accuracy (the latter is
different from the VOC measure in that it does not directly penalize false positives). The results in Table~\ref{tab:sbd} show that the zoom-out architecture obtains results better than those in~\cite{pinheiro2014recurrent}
and~\cite{farabet2013learning}, both in class accuracy and in pixel
accuracy.

\begin{table}[!th]
  \centering
{\small
  \begin{tabular}{|l||l|l|}
    \hline
Method & pixel accuracy & class accuracy\\
\hline
zoom-out (ours) & {\bf 86.1} & {\bf 80.9}\\
\hline
Multiscale CNN~\cite{farabet2013learning} & 81.4 & 76.0\\
Recurrent CNN~\cite{pinheiro2014recurrent} & 80.2 & 69.9\\
Pylon~\cite{lempitsky2011pylon} & 81.9 & 72.4 \\
Recursive NN~\cite{socher2011parsing}& 78.1 & -- \\
Multilevel~\cite{mostajabi2014robust} & 78.4 & --\\
\hline
  \end{tabular}
  \caption{Results on Stanford Background Dataset}
  \label{tab:sbd}}
\end{table}

\section{Zoom-out with modern deep CNNs}
\label{sec:zoomfurther}

In this section we explore the performance of zoom-out architecture by end-to-end training of zoom-out descriptors. We utilize modern deep CNNs such as ResNet~\cite{ResNet}, DenseNet~\cite{DenseNet} and NASNet~\cite{Nasnet} as a backbone of zoom-out architecture. 

\begin{figure*}[!t]
  \centering
    \includegraphics[width=0.9\textwidth]{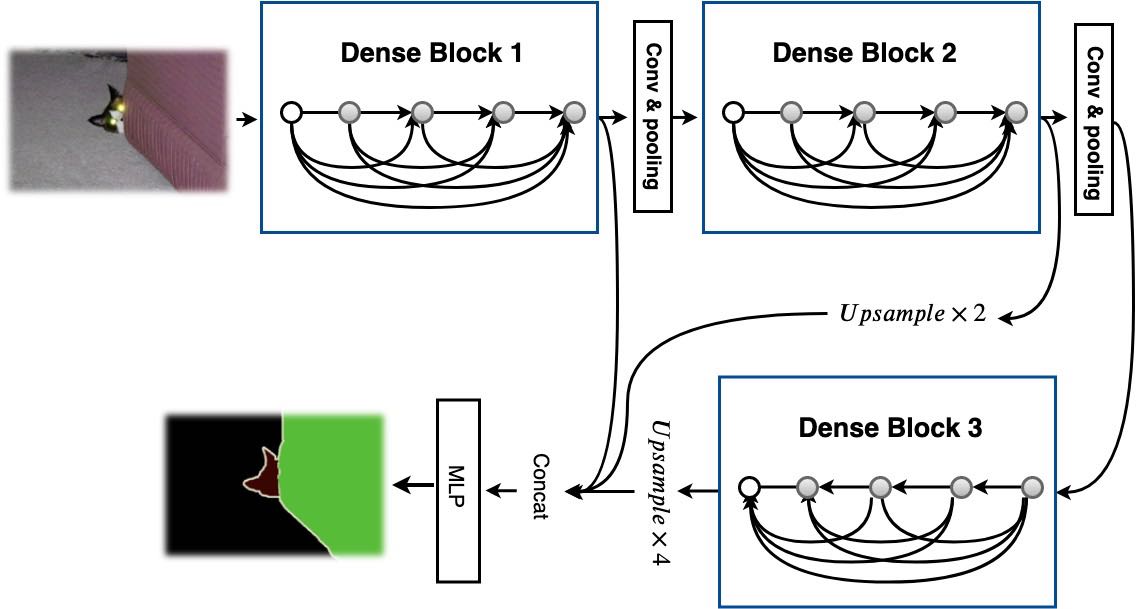}
  \caption{Illustration of zoom-out DenseNet with three dense blocks.}
  \label{fig:zoomoutdensnet}
\end{figure*}

 Huang \textit{et al}.~\cite{DenseNet} introduced Dense Convolutional Network "DenseNet".  DenseNet represents the latest high-performance evolution of ResNet-like
designs. A DenseNet contains multiple dense blocks. A dense block consists of multiple convolutional layers where each layer is connected to every other layer. All layers within the same dense block operate on the feature maps of the same spatial size. Each convolutional layer within a dense block produces $k$ feature maps at its output. The hyperparameter $k$ is called the growth rate of DenseNet. After each dense block a down sampling operation is applied which increases the receptive field of the subsequent dense blocks. By using DenseNet as a backbone of the 	zoom-out architecture, we create direct connections from dense blocks at different scales to the prediction layer. The pixel classifier on top of zoom-out features consists of 3-layer CNN with $1\times1$ convolutions with 1000 hidden units in each layer. Figure~\ref{fig:zoomoutdensnet} illustrates zoom-out DenseNet with three dense blocks. We use ImageNet pretrained 121-layer and 201-layer DenseNets. The growth rate for both DenseNets is 32.

\begin{figure*}[!t]
  \centering
    \includegraphics[width=1.0\textwidth]{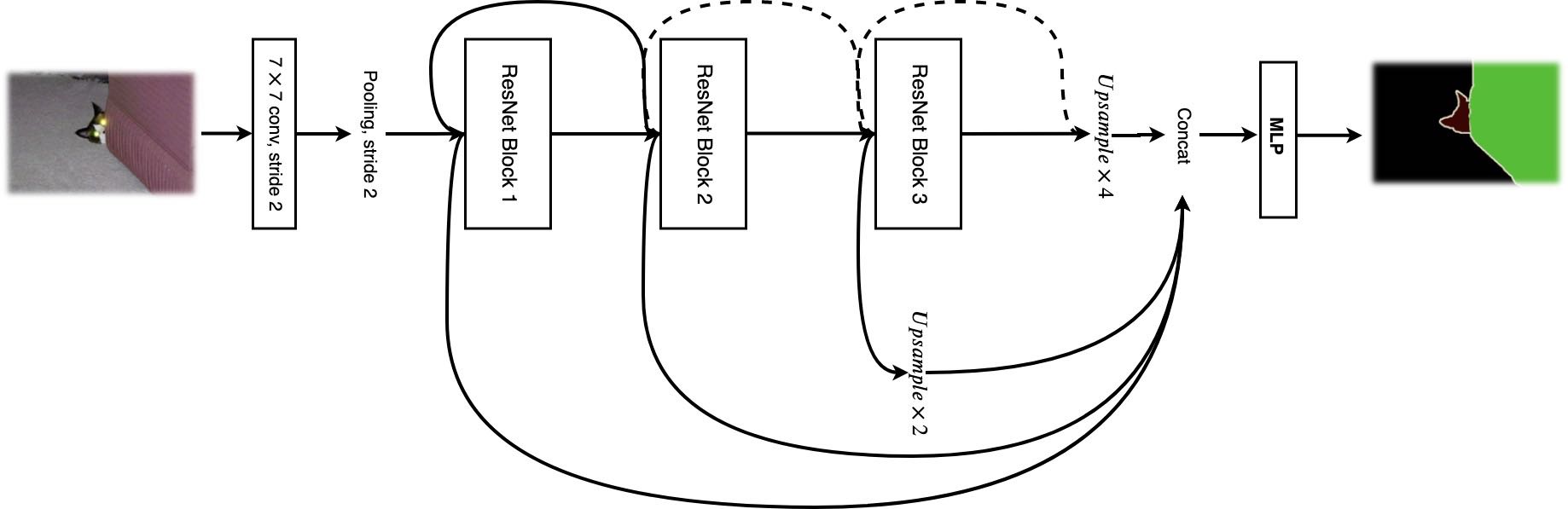}
  \caption{Illustration of zoom-out ResNet with three ResNet blocks. Dashed lines denote spatial size reduction of feature maps by the factor of two.}
  \label{fig:zoomoutresnetvertical}
\end{figure*}

We have also used ResNet as backbone of zoom-out architecture. For the case of ResNet, we extract zoom-out features at the end of each ResNet block before downsampling. Figure~\ref{fig:zoomoutresnetvertical} shows zoom-out ResNet with three ResNet blocks. We also apply pyramid pooling module~\cite{PSPNet} to zoom-out features.

\subsection{Experiments} 
Similar to previous settings, zoom-out spatial feature size is set to 1/4 of input image size. For the following experiments, we directly classify pixels rather than superpixels. Experiments are done in PyTorch~\cite{PyTorch}, using the Adam~\cite{Adam} update
rule when training networks. We keep the batch norm parameters of ImageNet
pretrained backbones frozen. We train the classifier and fine-tune the backbone for $60$ epochs with learning rates of $\text{10}^{-4}$ and $\text{10}^{-5}$  respectively.
Then we decrease the learning rate to $\text{10}^{-5}$ for the classifier and to $\text{10}^{-6}$ for the backbone and  train end-to-end for
additional $40$ epochs. We use a batch size of
$12$ when the models are trained on input of size $256 \times 256$ and
batch size of $8$ when the input size is $384 \times 384$.

Data augmentation includes: a random horizontal flip and a crop of random size in the 0.5 to 2.0 range of the original size and a random aspect ratio of 3/4 to 4/3 of
the original aspect ratio, which is finally resized to create a $256 \times 256$ or $384 \times 384$
image. Pretrained models are based
on the PyTorch torchvision library~\cite{torchvision}.

When COCO~\cite{COCO} pretraining is used, we only train on images that contain one or more object classes that are in PASCAL. We consider COCO object classes that are not in PASCAL as background. This leaves us with around 80000 training images. Once the model is converged, we fine tune it on PASCAL with initial learning
rate of  $\text{10}^{-5}$ for the pixel classifier and  and $\text{10}^{-6}$ for the COCO
pretrained model.

Multi-scale inference refers to running inference on the validation images at multiple scales. We resize validation images such that the smaller side has lengths of 256, 384 and original length. The larger side of the image is resized such that resized image keeps the original aspect ratio. We also perform inference on the horizontally flipped images as well. For each of these inputs, we compute segmentation predictions and resize it to the original input size. At the end we average these predictions.

Table~\ref{tab:Pascalmodern} summarizes zoom-out performance with DenseNet, ResNet and NASNet backbones on PASCAL VOC 2012 validation set. Using DenseNet, ResNet, or NASNet significantly improves results over zoom-out with VGG-16. This is consistent with the observation that we had earlier in section~\ref{sec:expzoom} when we replaced CNN-S with VGG-16. These results also demonstrate that zoom-out can easily be integrated with modern architectures and achieve comparable accuracy with recently proposed methods for semantic segmentation~\cite{YK:ICLR:2016,CPKMY:arXiv:2016}. Applying fully connected CRFs (\ref{subsec:CRFs}) to our best performing model, zoom-out NASNet-A adds 0.1\% to mIoU. This minor improvement in mIoU suggests our zoom-out architecture has captured short-range and long-range dependencies between image elements.

\begin{table}[!th]
  \centering
{\small
\setlength{\tabcolsep}{12pt}
  \begin{tabular}{l|	c    | 	c    | 	   c|l}

Method		& Crop size      & MS inference	     	 &  COCO 		& Mean IoU\\
\hline
zoom-out DenseNet-121 & $256 \times 256$  &  - & - & 73.7\\
zoom-out DenseNet-201  &  $256 \times 256$  &  - & - & 77.4\\
zoom-out DenseNet-201  &   $256 \times 256$  & - & - & 78.6\\
zoom-out DenseNet-201  &   $384 \times 384$  & \checkmark & - & 79.6\\
zoom-out NASNet-A  &   $384 \times 384$  & \checkmark & - & 81.4\\
zoom-out NASNet-A  &   $384 \times 384$  & \checkmark & \checkmark & 84.7\\
zoom-out NASNet-A CRF  &   $384 \times 384$  & \checkmark & \checkmark & 84.8\\
\hline
  \end{tabular}}
  \caption{VOC12 validation mIoU of zoom-out with DenseNet and NASNet backbones.}
  \label{tab:Pascalmodern}
\end{table}

\subparagraph{Comparison with recent methods based on dilated/atrous convolution} \label{dilatedconv}
 We compare zoom-out with recent approaches for sematic segmentation. The key idea behind atrous/dilated convolution was originally developed by Holschneider \textit{et al}.~\cite{Holschneider} for wavelet decomposition. Dilated convolutions for semantic segmentation were introduced after zoom-out by Yu \textit{et al}.~\cite{YK:ICLR:2016}. Dilated convolution is designed to exponentially increase the receptive field of convolutional layers without sacrificing spatial resolution of feature maps by applying convolution to input with defined gaps. For the dilated ResNet experments we use the model released by Zhou \textit{et al}.~\cite{zhou2018semantic}. A similar approach, called atrous convolution, has been also deployed in DeepLab architecture~\cite{CPKMY:arXiv:2016}. As it is shown in table~\ref{tab:dialzoomout}, zoom-out achieves similar accuracy to the methods that use dilated/atrous convolution approach. However, it achieves similar accuracy with lower processing and memory consumption costs. In dilated convolution based architectures, mid to late stage DenseNet/ResNet blocks operate on feature responses with spatial resolution of size 1/8 of input, while in a normal ResNet/DenseNet these layers operate on gradually decreasing resolutions of 1/8, 1/16 and 1/32. Mid/late stage ResNet/DenseNet blocks contain convolutional layers with an increased number of channels, compared to early stage convolutional layers. Operating on large spatial resolution significantly reduces network speed. Table~\ref{tab:Pascalmodern} shows further improvement in accuracy can be achieved by using zoom-out in conjunction with dilated ResNet.

\begin{table}[!th]
  \centering
{\small
\setlength{\tabcolsep}{12pt}
  \begin{tabular}{l|	c|l}

Method		& Crop size 	& Mean IoU\\
\hline
dilated ResNet-101 &  $256 \times 256$  & 77.4\\
\hline
zoom-out ResNet-101 &  $256 \times 256$  &  77.6\\
zoom-out dilated ResNet-101 &  $256 \times 256$  &78.4\\
zoom-out dilated ResNet-101 &  $384 \times 384$  &79.8\\
\hline
  \end{tabular}}
  \caption{VOC12 validation mIoU of zoom-out with dilated ResNet backbones.}
  \label{tab:dialzoomout}
\end{table}

\begin{figure}
  \centering
  \setlength{\tabcolsep}{4pt}
\begin{tabular}{cc     cc}
\includegraphics[width=.22\textwidth]{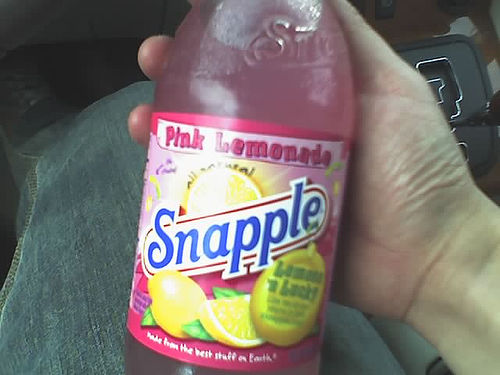} &
 \includegraphics[width=.22\textwidth]{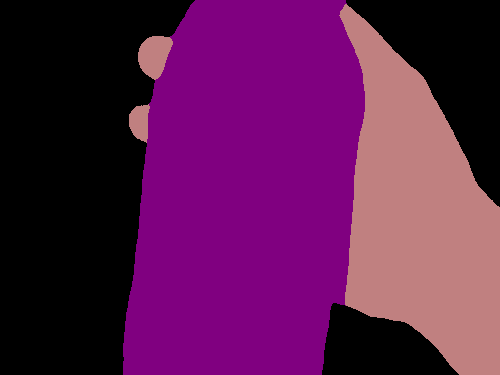}&
         \includegraphics[width=.22\textwidth]{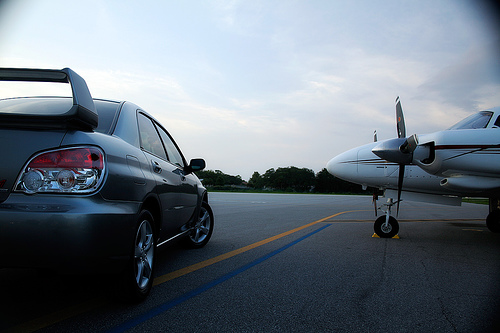} &
         \includegraphics[width=.22\textwidth]{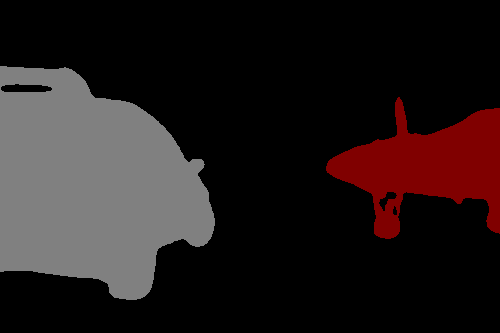}\\
  
  \includegraphics[width=.22\textwidth]{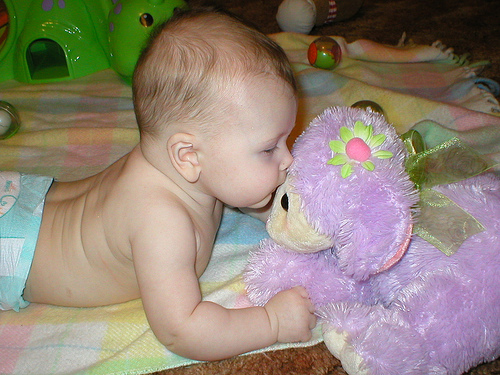} &
 \includegraphics[width=.22\textwidth]{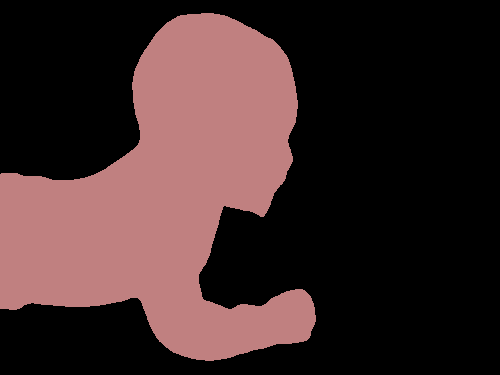}&
         \includegraphics[width=.22\textwidth]{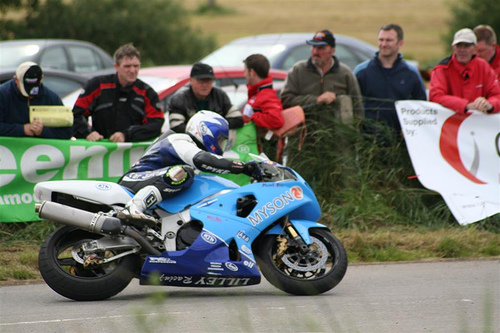} &
         \includegraphics[width=.22\textwidth]{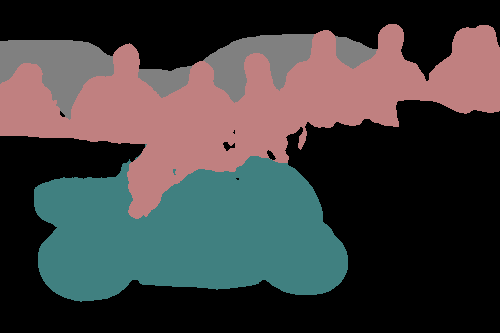}\\
         
 \includegraphics[width=.22\textwidth]{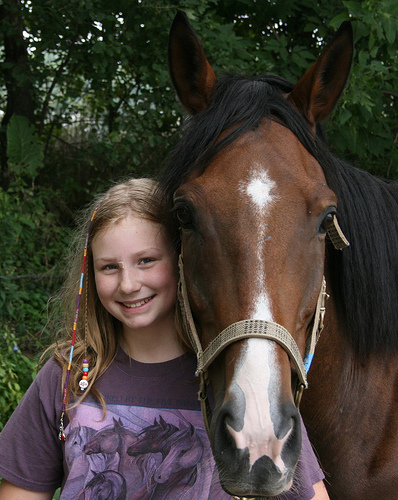} &
 \includegraphics[width=.22\textwidth]{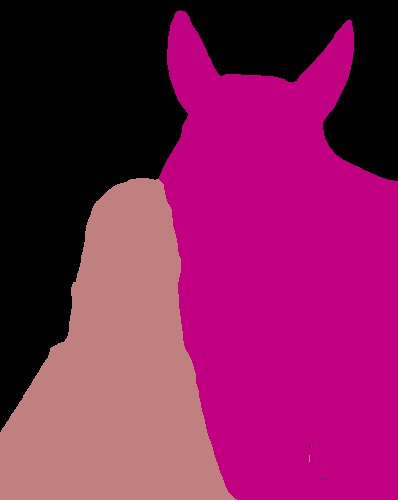}&
         \includegraphics[width=.22\textwidth]{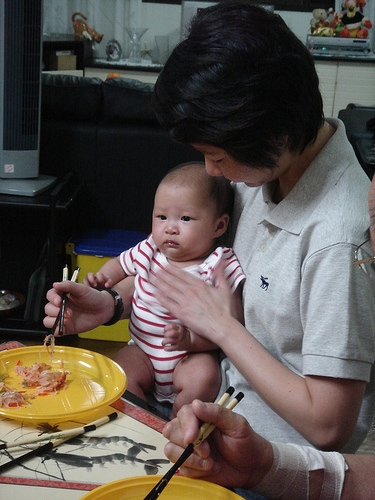} &
         \includegraphics[width=.22\textwidth]{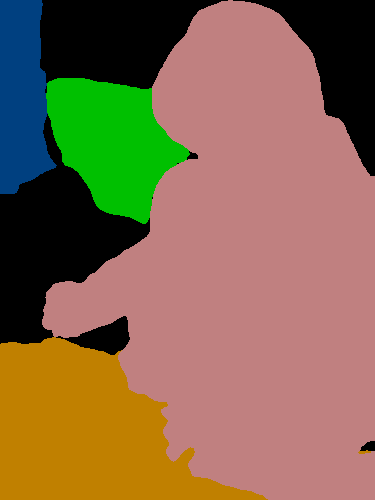}\\

 \includegraphics[width=.22\textwidth]{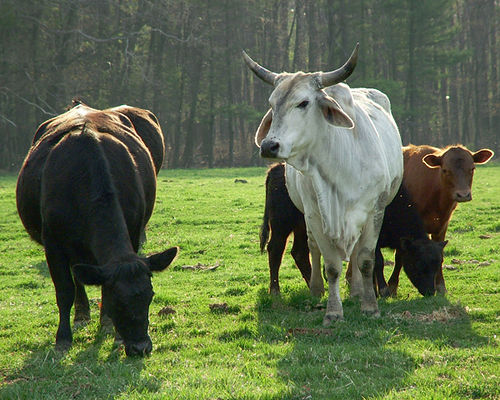} &
 \includegraphics[width=.22\textwidth]{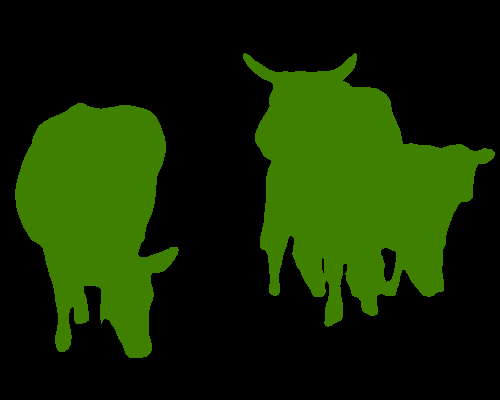}&
         \includegraphics[width=.22\textwidth]{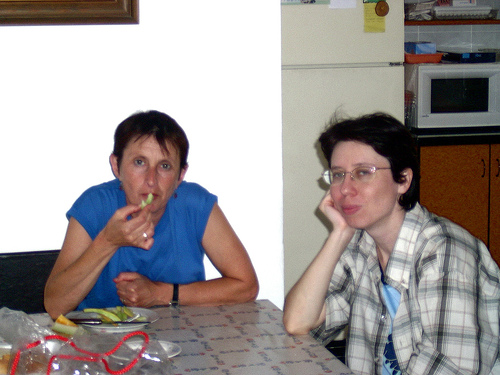} &
         \includegraphics[width=.22\textwidth]{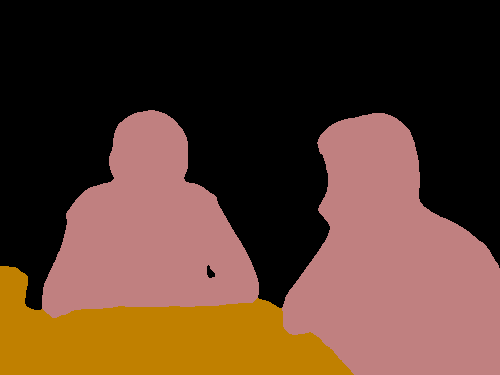}\\

 \includegraphics[width=.22\textwidth]{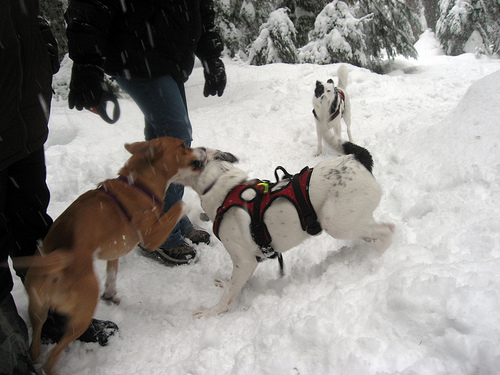} &
 \includegraphics[width=.22\textwidth]{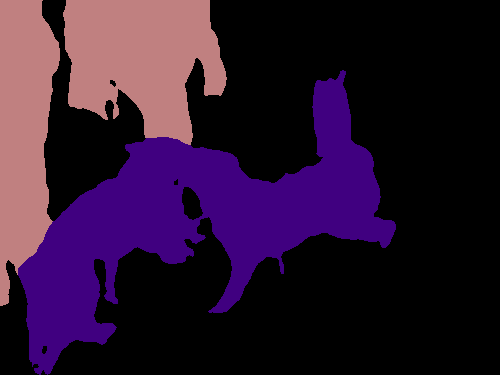}&
         \includegraphics[width=.22\textwidth]{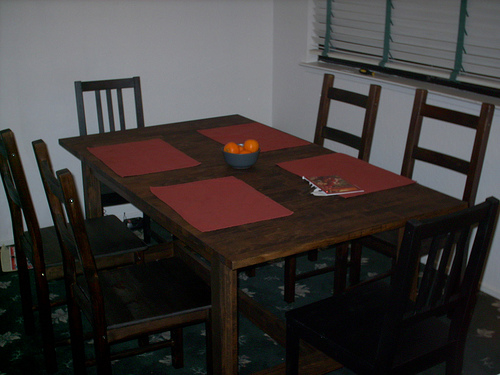} &
         \includegraphics[width=.22\textwidth]{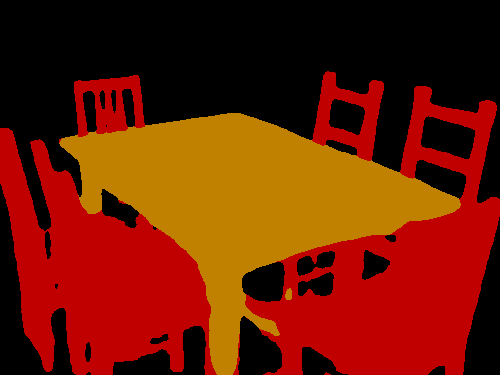}\\

 \includegraphics[width=.22\textwidth]{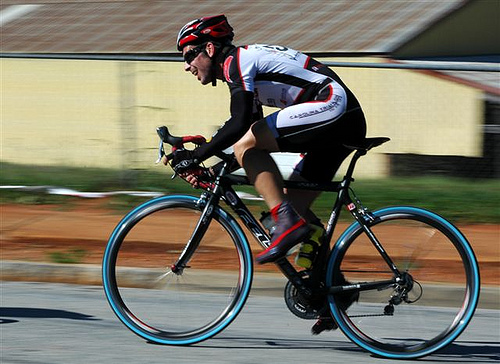} &
 \includegraphics[width=.22\textwidth]{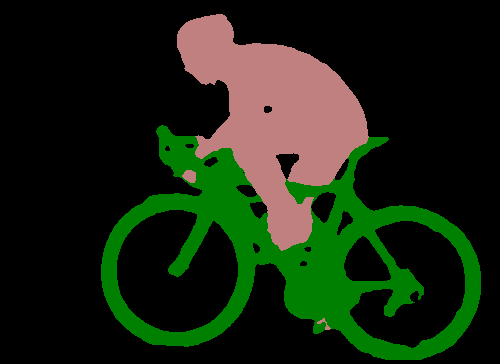}&
         \includegraphics[width=.22\textwidth]{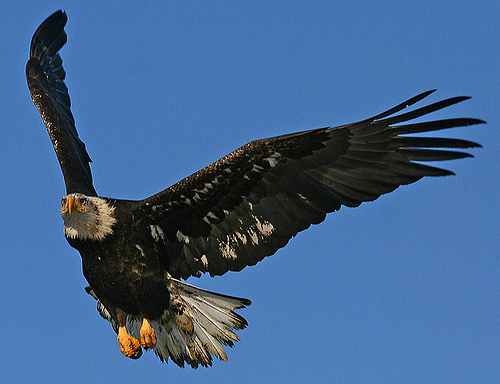} &
         \includegraphics[width=.22\textwidth]{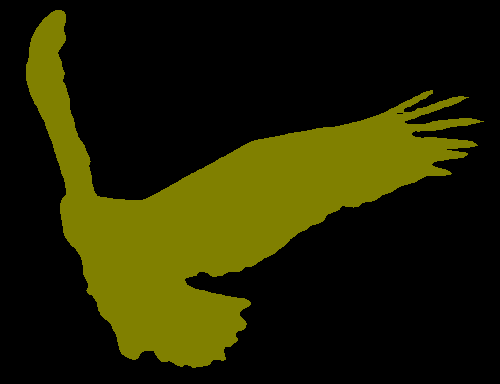}
  
\end{tabular}
  \caption{Example segmentations on VOC 2012 test set with our zoom-out NASNet-A.}\label{fig:pascalnewvis}  
\end{figure}

We pick our best perfoming model, zoom-out NASNet-A and evaluate it on the test set of PASCAL dataset. Table~\ref{tab:newtest} shows zoom-out results on VOC 2012 test set compared to state-of-the-art architectures. Figure~\ref{fig:pascalnewvis} visualizes our results on the test set of VOC 2012.

\begin{table}[!th]
  \centering
{\small
\setlength{\tabcolsep}{12pt}
  \begin{tabular}{l | l}

Method &  mIoU\\
\hline
PSPNet~\cite{PSPNet} &   85.4\\
DeepLabv3~\cite{deeplabv3}  & 85.7\\
DeepLabv3+ (Xception)~\cite{deeplabv3plus} & 87.8\\
MSCI~\cite{MSCI} &   88\\
\hline
zoom-out NASNet-A  &  86.6\\
\hline
  \end{tabular}}
  \caption{Semantic segmentation results on VOC 2012 test. See Table~\ref{tab:newdetails-2012} for per-class
    accuracies of the zoom-out method.}
  \label{tab:newtest}
\end{table}

\renewcommand{\arraystretch}{1}
\setlength{\tabcolsep}{0.5pt}
\begin{table*}[!th]
  \centering
{\small
  \begin{tabu}{|l|c||c|c|c|c|c|c|c|c|c|c|c|c|c|c|c|c|c|c|c|c|c|}
    \hline
    class &
    \rot{90}{\bf mean} &
    \rot{90}{bg} &
    \includegraphics[width=5mm]{img/aeroplane-icon}&
    \includegraphics[width=5mm]{img/bicycle-icon}&
    \includegraphics[width=5mm]{img/bird-icon}&
    \includegraphics[width=5mm]{img/boat-icon}&
    \includegraphics[width=5mm]{img/bottle-icon}&
    \includegraphics[width=5mm]{img/bus-icon}&
    \includegraphics[width=5mm]{img/car-icon}&
    \includegraphics[width=5mm]{img/cat-icon}&
    \includegraphics[width=5mm]{img/chair-icon}&
    \includegraphics[width=5mm]{img/cow-icon}&
    \includegraphics[width=5mm]{img/diningtable-icon}&
    \includegraphics[width=5mm]{img/dog-icon}&
    \includegraphics[width=5mm]{img/horse-icon}&
    \includegraphics[width=5mm]{img/motorbike-icon}&
    \includegraphics[width=5mm]{img/person-icon}&
    \includegraphics[width=5mm]{img/pottedplant-icon}&
    \includegraphics[width=5mm]{img/sheep-icon}&
    \includegraphics[width=5mm]{img/sofa-icon}&
    \includegraphics[width=5mm]{img/train-icon}&
    \includegraphics[width=5mm]{img/tv-icon}
    \\
    \hline
    \rowfont{\footnotesize}
    {\normalsize acc} &
    86.6 &
    96.7 &
    96.4 & 78.3 & 96.3 & 79.6 & 85.4 &
    97.6 & 85.7 & 95.4 & 52.9 & 91.8 &
    81.4 & 92.5& 96.4 & 93.3 &90.8&
    76.9&90 & 69.3& 89.9 &81.9\\
    \hline
  \end{tabu}
}
\vspace{1em}\caption{Detailed results of our method on VOC 2012 test.}
  \label{tab:newdetails-2012}
\end{table*}

\begin{figure}[!th]
  \centering
\begin{tabular}{c       c}
\includegraphics[width=.42\textwidth]{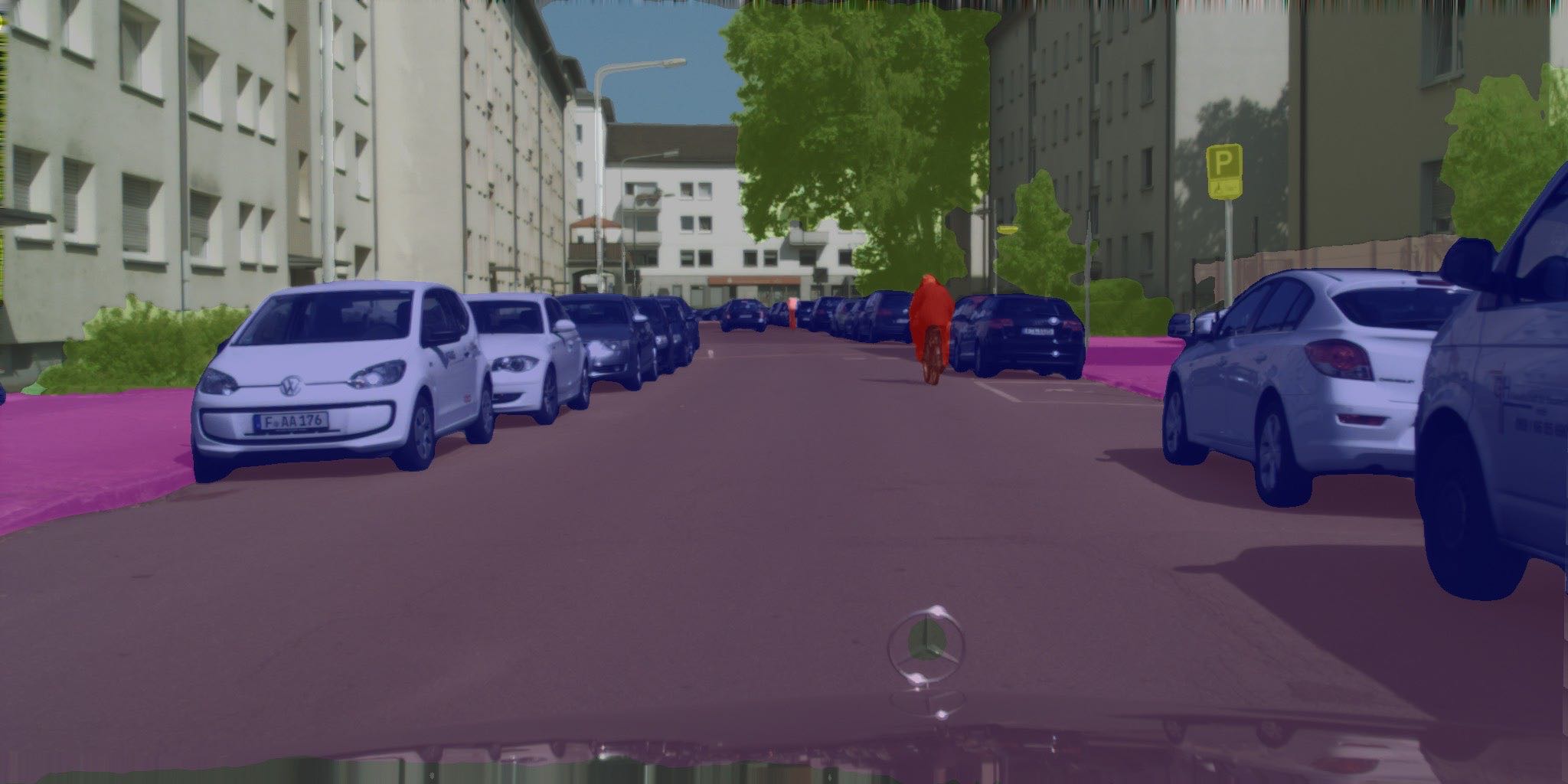} &
     \includegraphics[width=.42\textwidth]{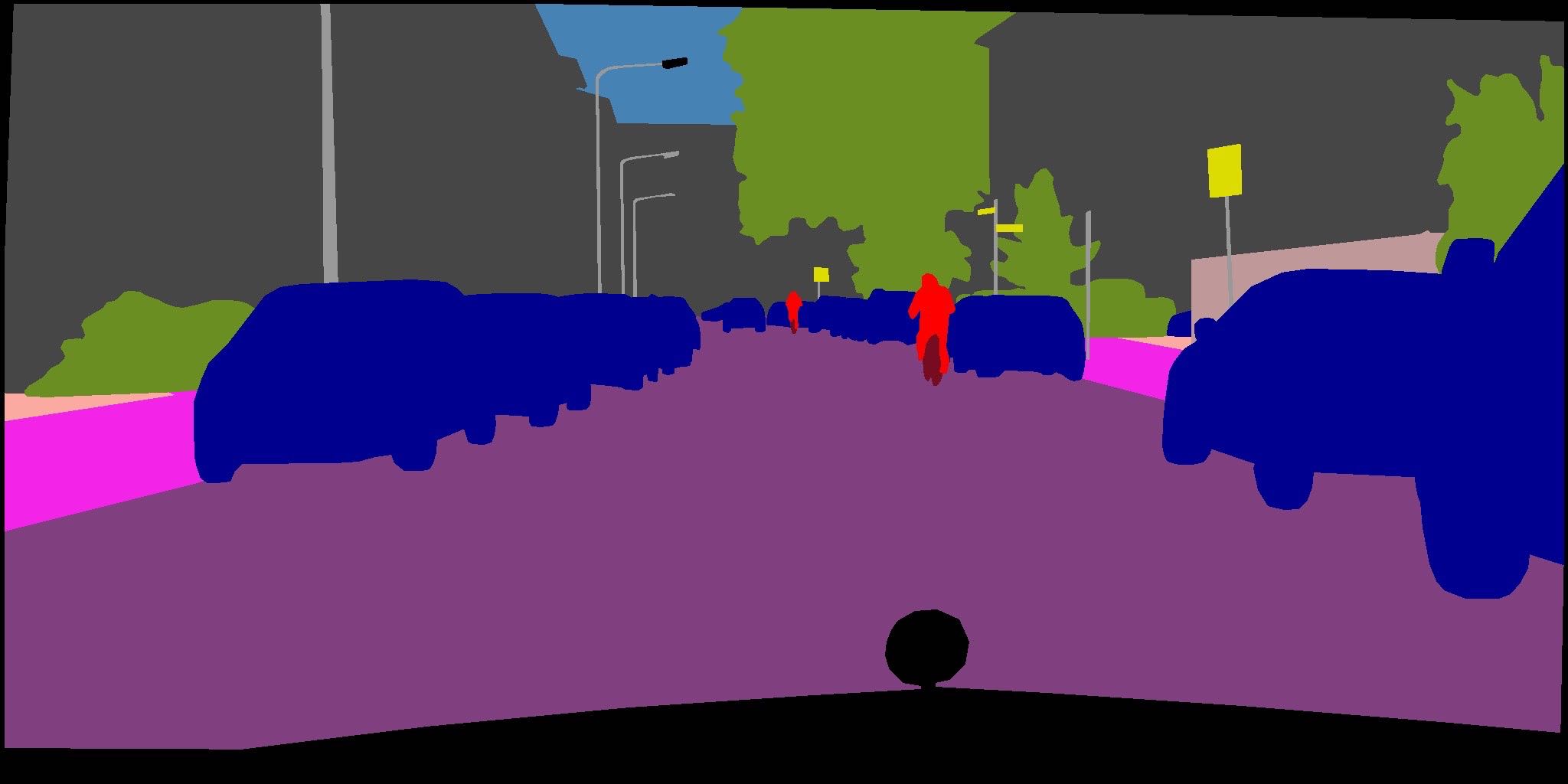}\\
 \includegraphics[width=.42\textwidth]{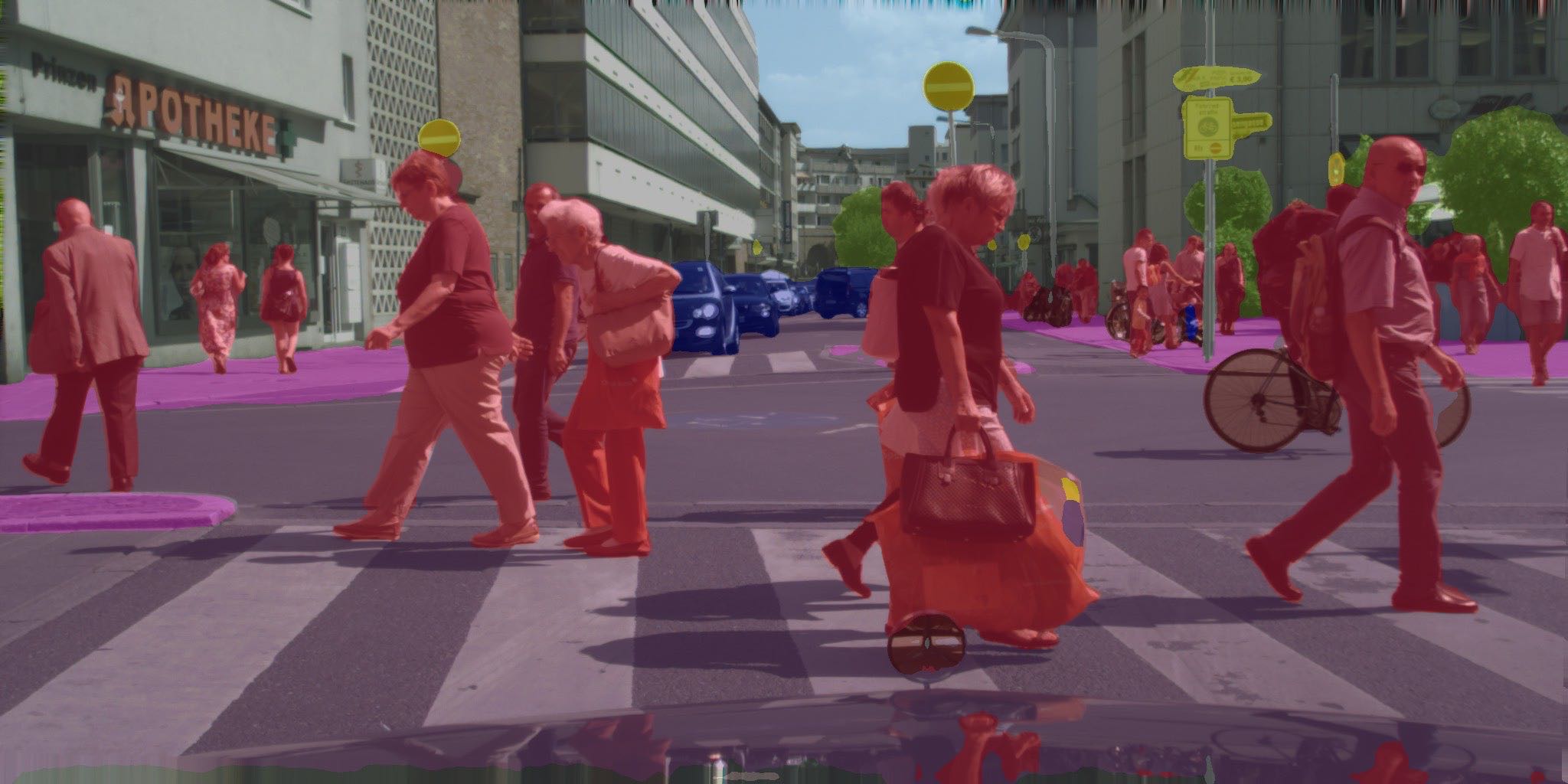} &
      \includegraphics[width=.42\textwidth]{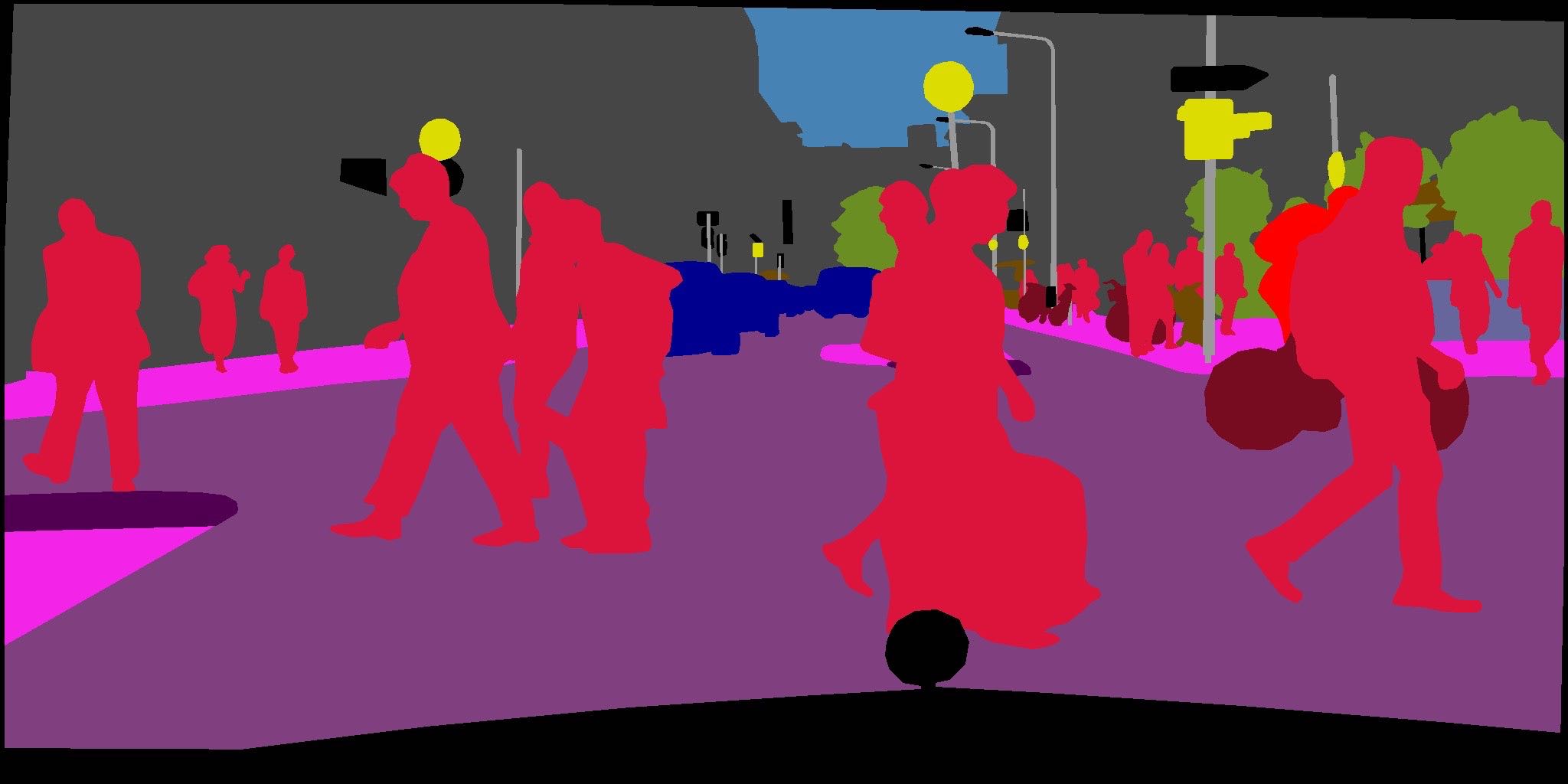} \\
\includegraphics[width=.42\textwidth]{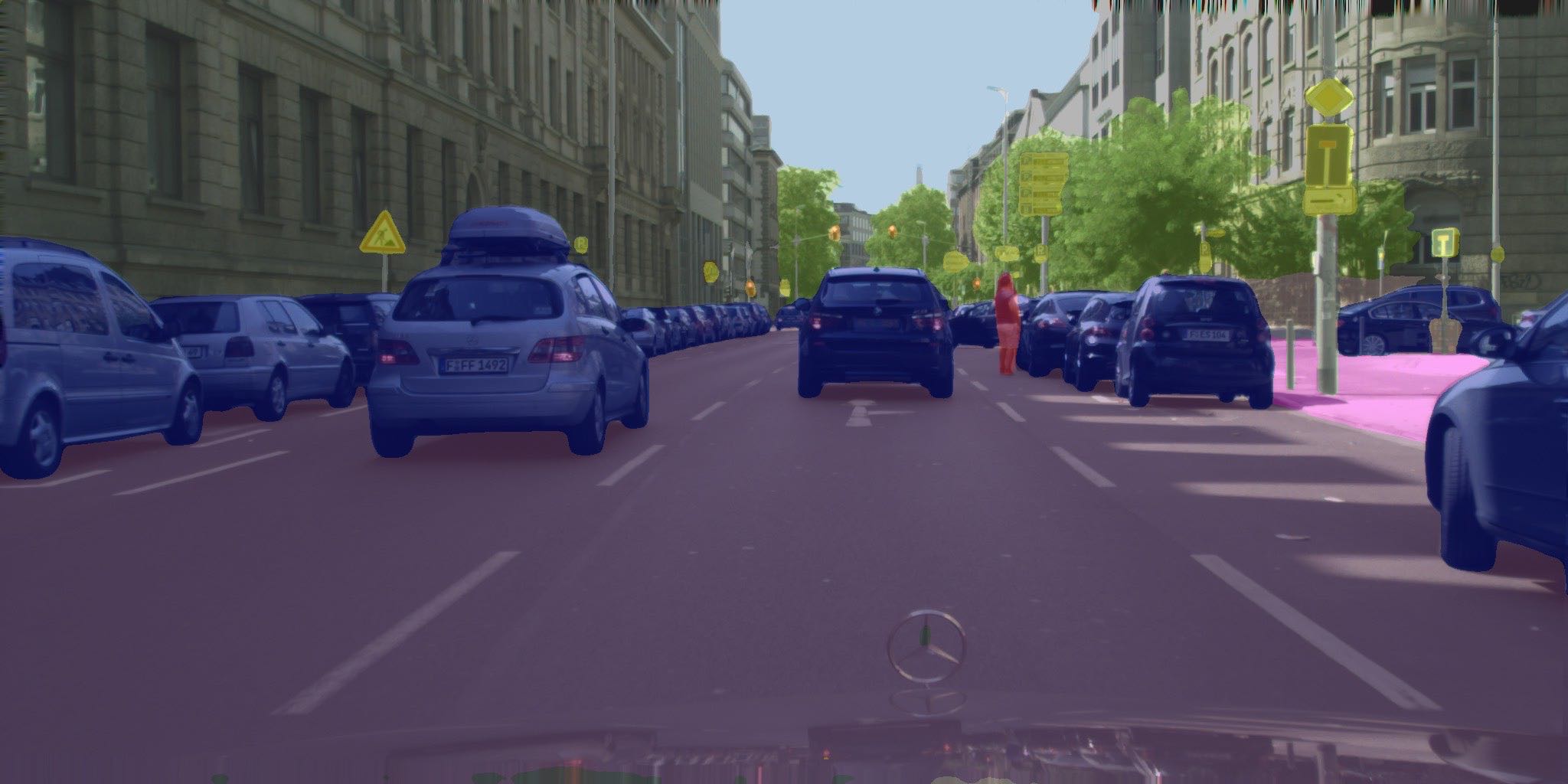} &
     \includegraphics[width=.42\textwidth]{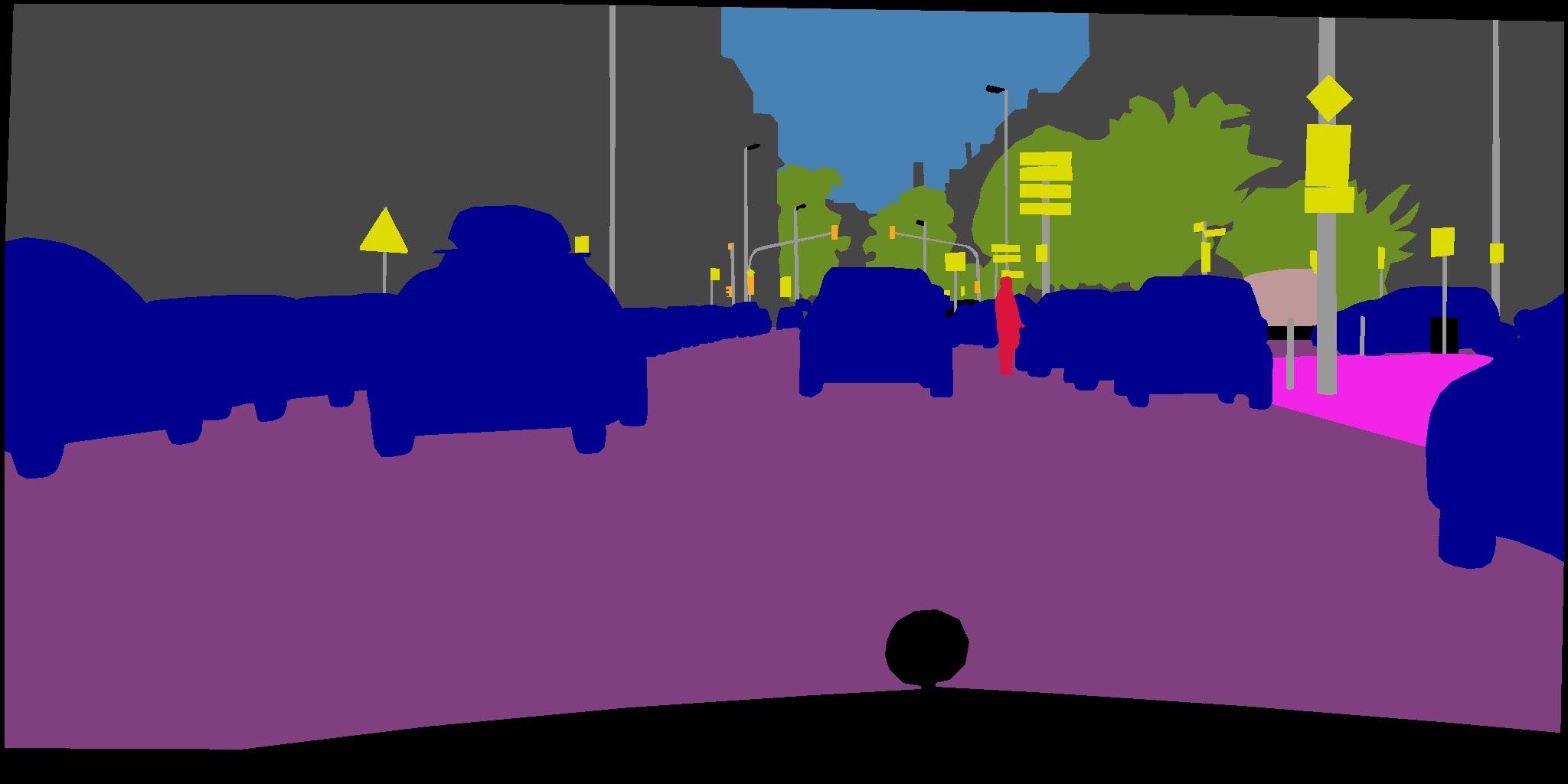}\\
 \includegraphics[width=.42\textwidth]{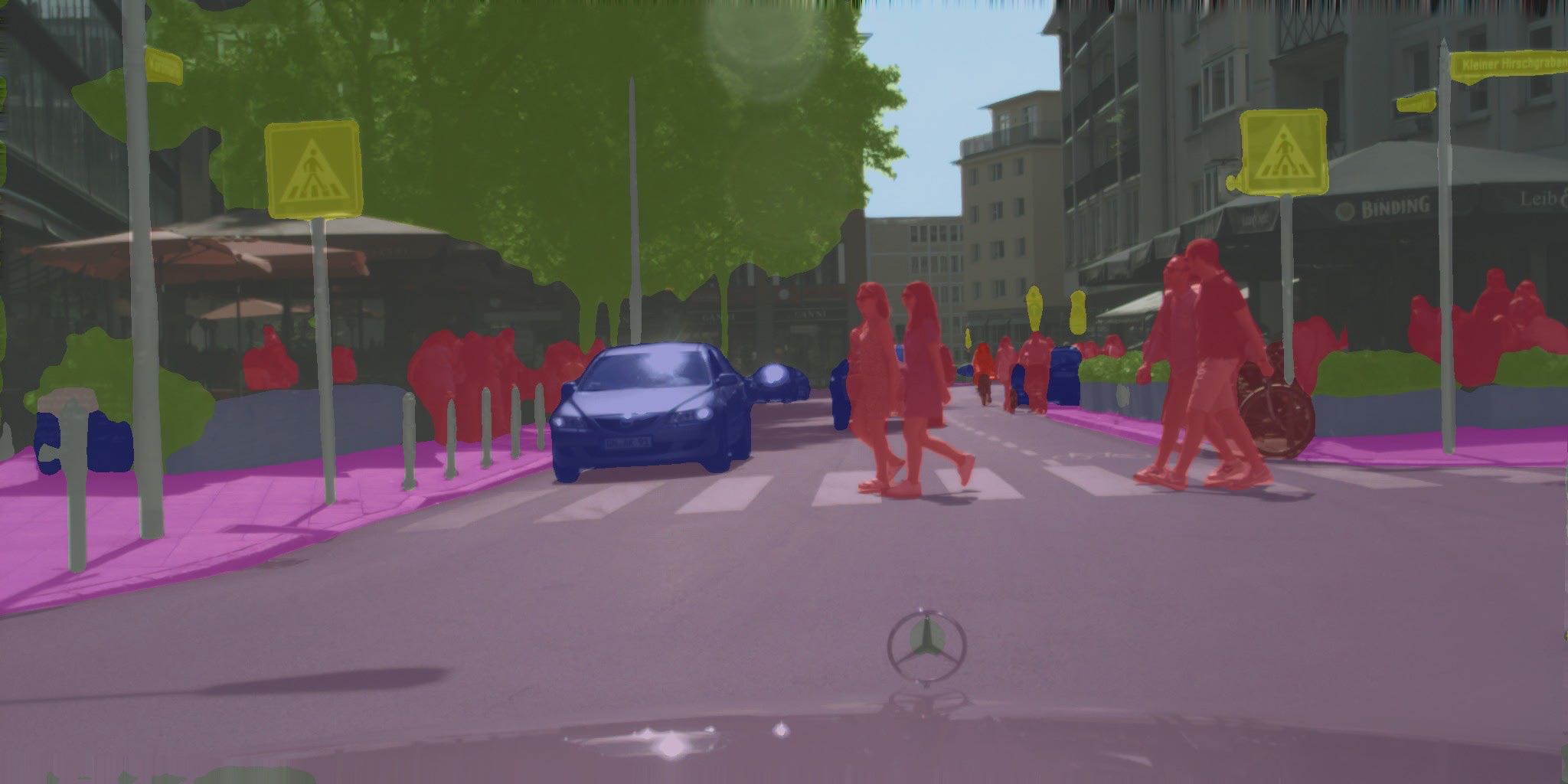} &
       \includegraphics[width=.42\textwidth]{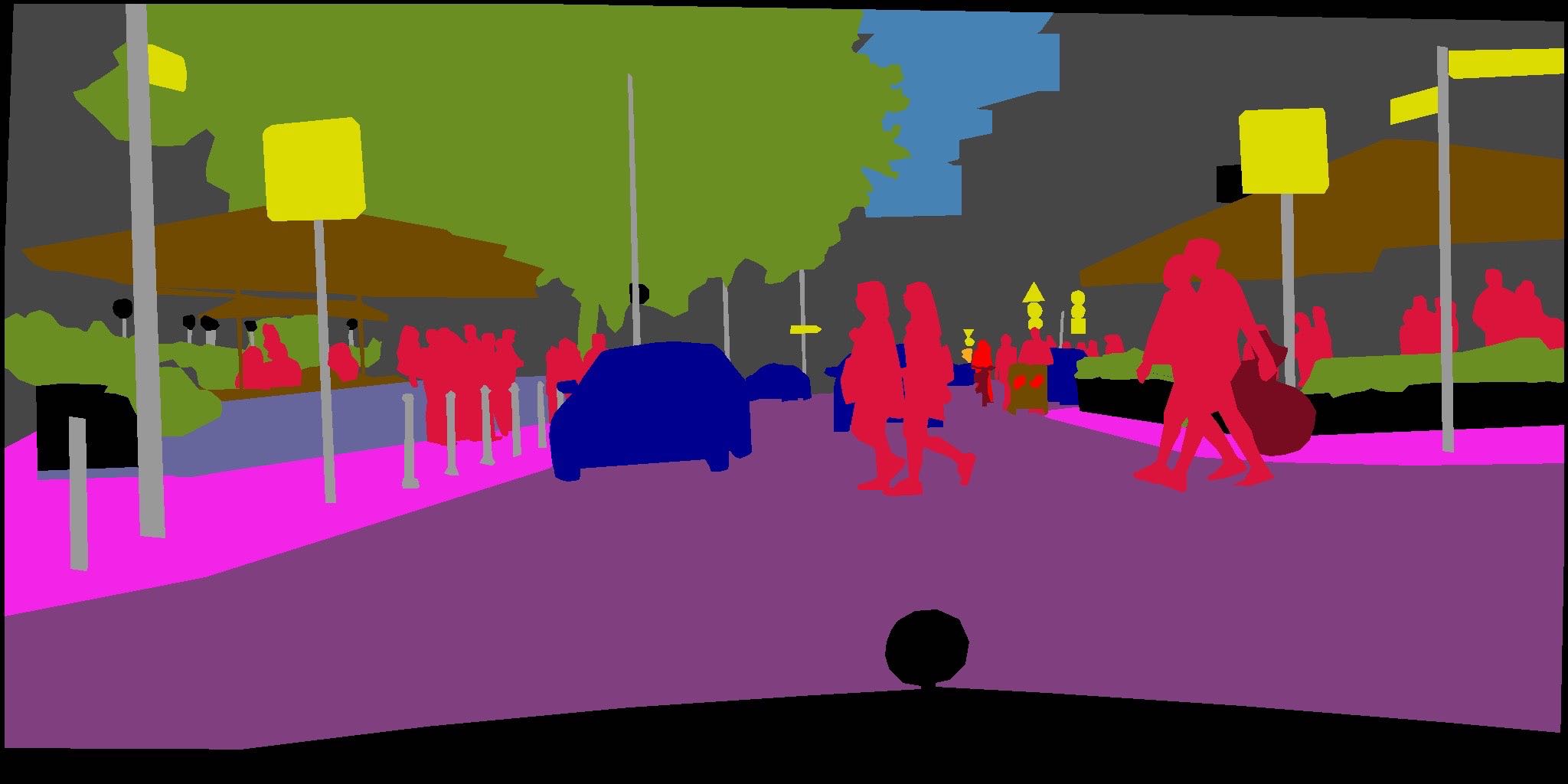} \\  
  
  \includegraphics[width=.42\textwidth]{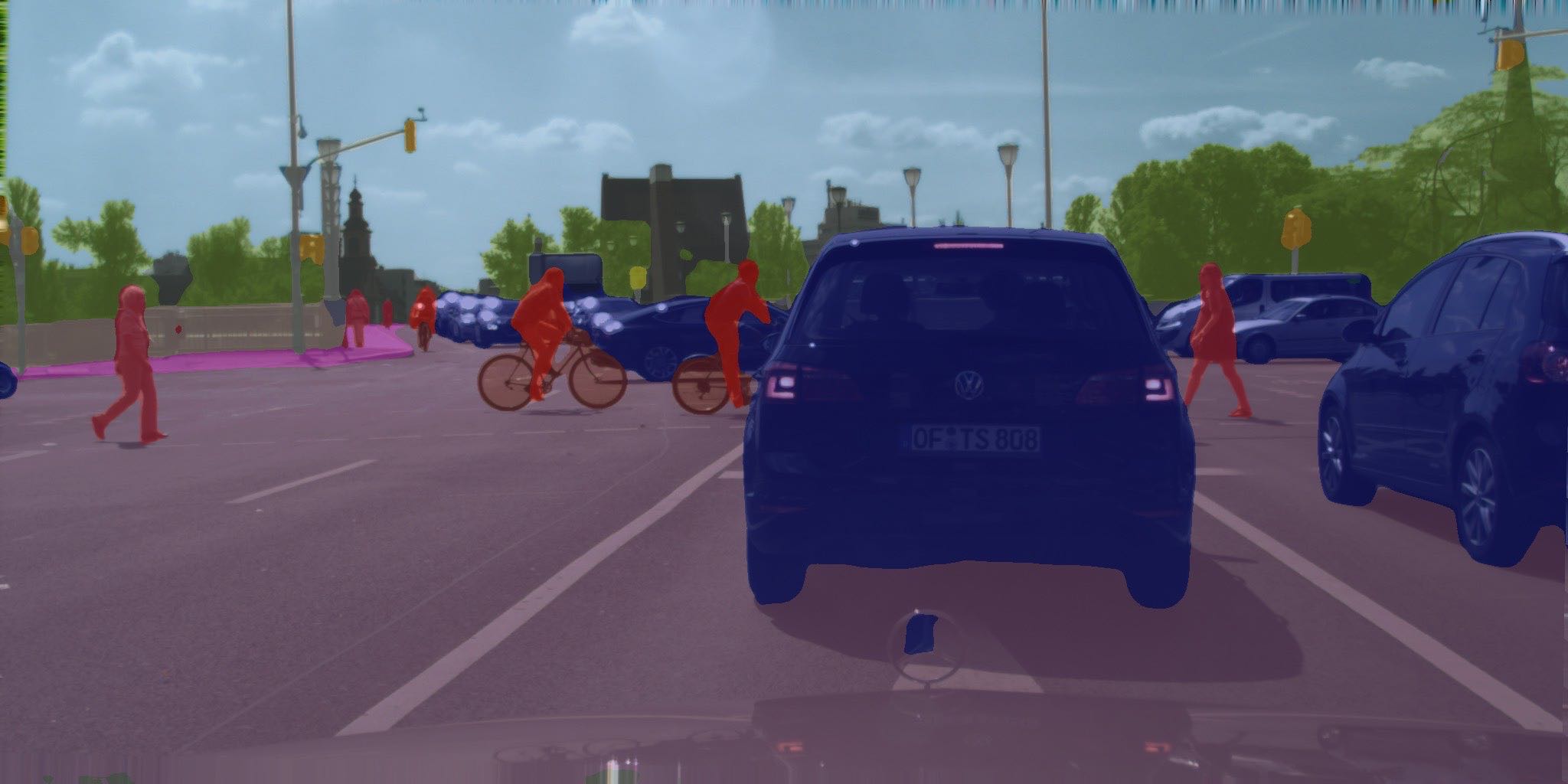} &
       \includegraphics[width=.42\textwidth]{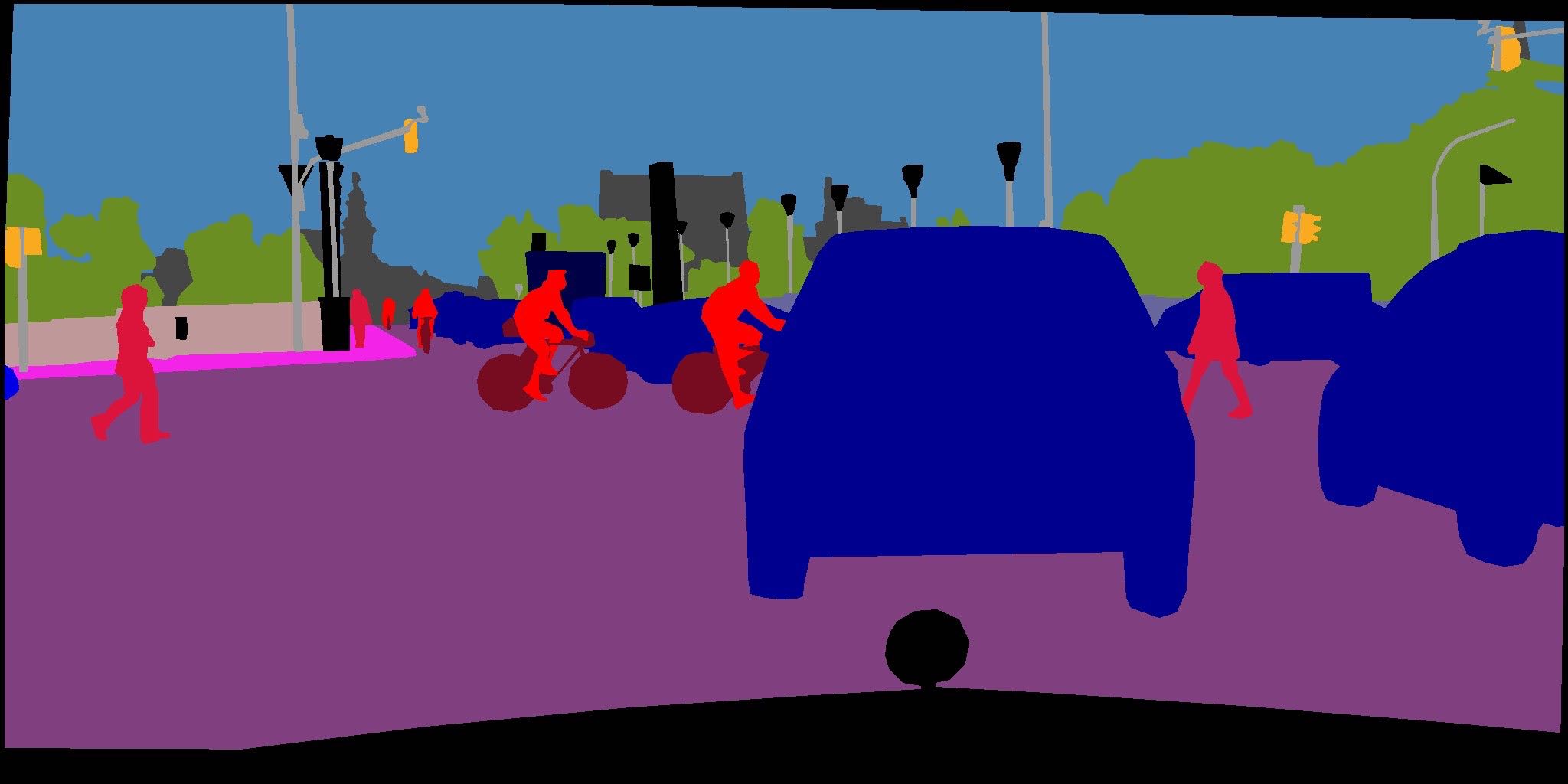}\\
 \includegraphics[width=.42\textwidth]{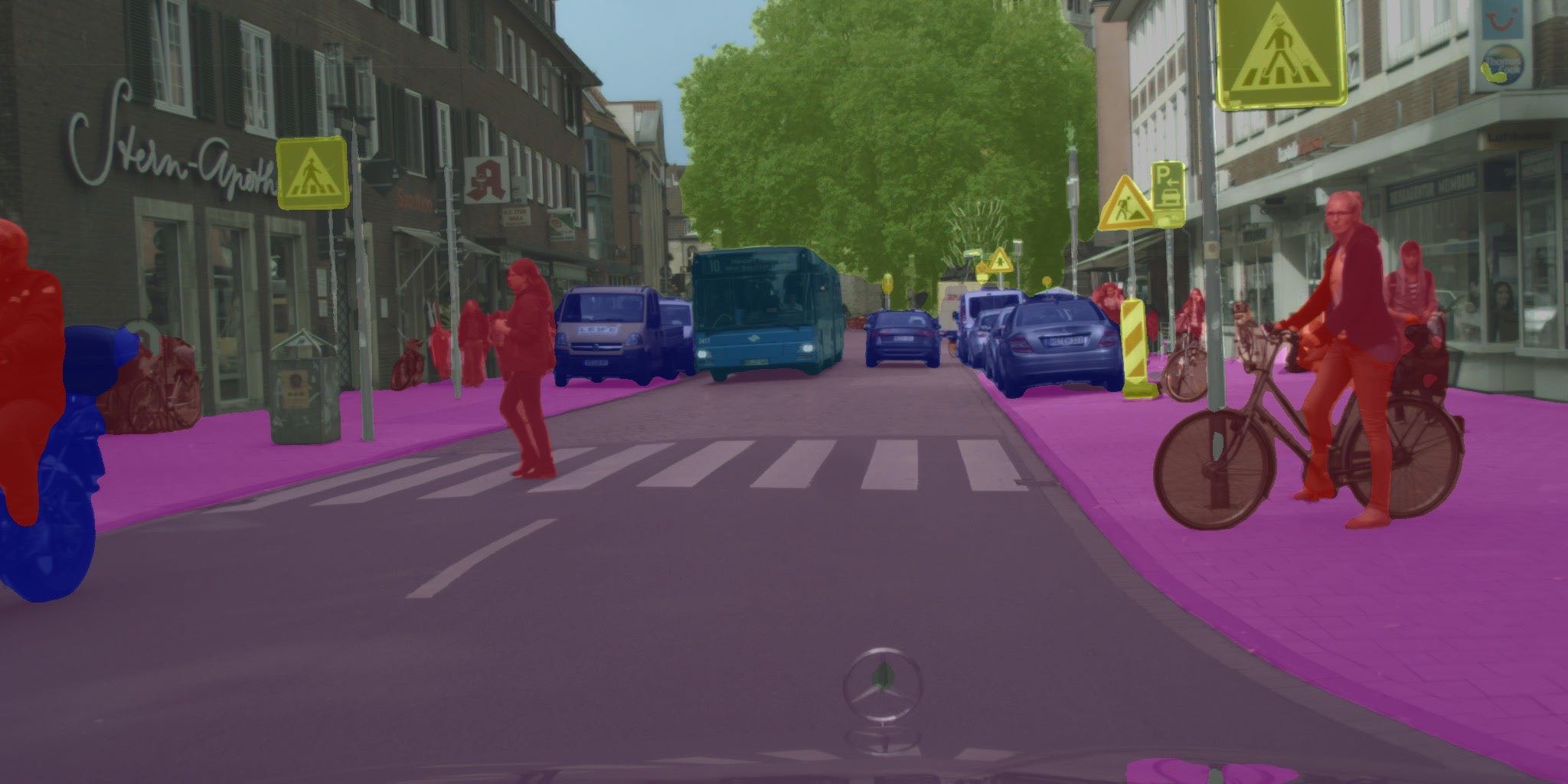} &
         \includegraphics[width=.42\textwidth]{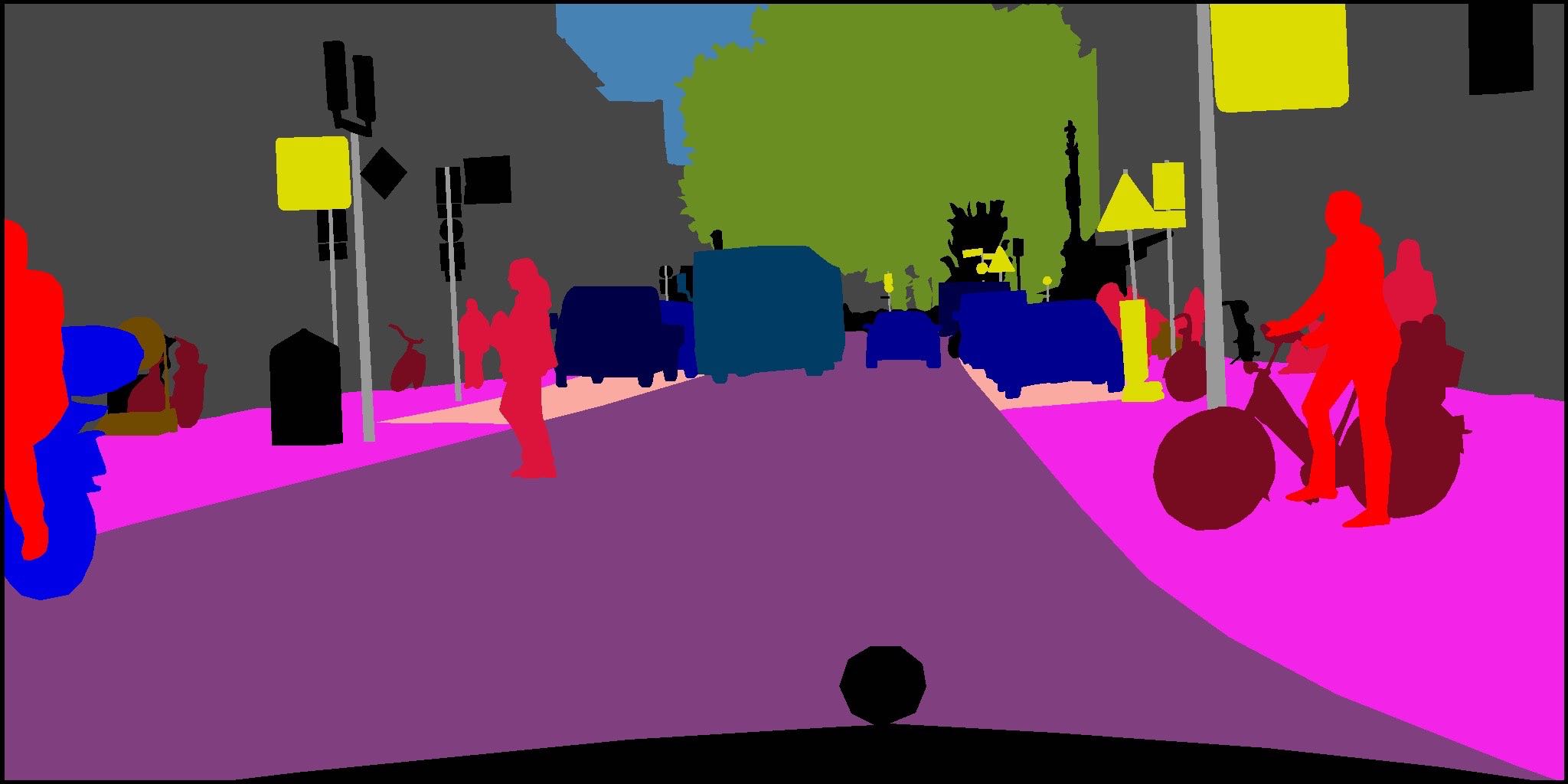} 
\end{tabular}
  \caption{Visualization results of zoom-out dilated ResNet-101 on Cityscapes validation set. Left images are our semantic segmentation predicitons overlay on RGB images. Right images are ground truth segmenations.}\label{fig:cityscapesvis}  
\end{figure}

\subsubsection{Results on Cityscapes dataset} 

Cityscapes is a rich urban scene understanding dataset~\cite{cityscapes}. It contains 5000 high quality pixel-level segmentation
annotations of complex urban scene images. These 5000 annotated images are split to 2975, 500, 1525 training, validation and test sets respectively. In addition to 
5000 finely annotated images, 20000 images are provided with coarse annotations. Segmentation annotations include 30 object classes. However,
some of these classes are too rare and are excluded from standard benchmark.  Following the standard benchmark, we evaluate our zoom-out model on semantic segmentation of 19 object categories. For the following experiments we train our models for 160 epochs on the 2975 finely annotated training images. For the zoom-out models we set the zoom-out features spatial size to be 1/8 of the input image size. We follow the training procedure and parameters of~\cite{you2019torchcv}. Table~\ref{tab:cityscapes} shows the results of various architectures based on ResNet-101 backbone with pyramid pooling module~\cite{PSPNet}. In Table~\ref{tab:cityscapes} similar to the results on PASCAL, zoom-out achieves similar accuracy to dilated convolution based model and once it is used in conjunction with dilated convolution it significantly improves the results. Figure~\ref{fig:cityscapesvis} visualizes our results on on Cityscapes validation set.

\begin{table}[!th]
  \centering
{\small
\setlength{\tabcolsep}{12pt}
  \begin{tabular}{l|	c |l}

Method		& Crop size  	& Mean IoU\\
\hline
dilated ResNet-101 &  $512 \times 512$  & 76.5\\
\hline
zoom-out ResNet-101 &  $512 \times 512$  &   76.6\\
zoom-out dilated ResNet-101 &  $512 \times 512$  &  78.3\\
\hline
  \end{tabular}}
  \caption{Semantic segmentation results on Cityscapes validation set.}
  \label{tab:cityscapes}
\end{table}

\section{Summary}
\label{sec:zoomconc}

We presented zoom-out for learning and extracting rich representations. This chapter focused on the application of zoom-out architecture in semantic image segmentation. Despite apparent simplicity of our method and lack of explicit representation of the
structured nature of the segmentation task, we showed that we can far
surpass previous state of the art which was based on hand-crafted features with a random field or structured support vector machine model of considerable complexity. We were the first ones to show that segmentation, just like image classification, detection and other recognition tasks, can benefit from the advent of deep convolutional networks. 
Our further experiments with modern architectures such as DenseNet and ResNet in Section~\ref{sec:zoomfurther} revealed that zoom-out can achieve similar accuracy 
to recently proposed methods for semantic segmentation like dilated/atrous convolution while being faster and having less memory consumption. Furthermore, we showed that zoom-out can be applied to networks with dilated convolution as well and increase the accuracy. Having a powerful backbone network will significantly improve results as we observed a significant boost in accuracy by replacing CNN-S with VGG-16 and replacing VGG-16 with ResNet or DenseNet. In that regard, zoom-out can be easily applied to different choices of backbone CNN architectures. In the following we briefly mention some of the work/applications that have utilized zoom-out.

\paragraph{Abdominal lymph node detection}
Shin \textit{et al}.~\cite{shinetal} incorporate zoom-out multi-scale feature extraction strategy for thoraco-abdominal lymph node (LN) detection. Zoom-out multi-scale feature extraction strategy led to their best performance model. 
\paragraph{Pancreas segmentation}
Roth \textit{et al}.~\cite{rothseg} utilized zoom-out features for pancreas segmentation in abdominal computed tomography (CT) scans. They computed zoom-out features for each superpixel and then assigned each superpixel a probability of being pancreas. Farag  \textit{et al}~\cite{Faragseg} also used similar approach for pancreas segmentation.
\paragraph{Foreground extraction}
Kolkin \textit{et al}.~\cite{Nicksen} utilized zoom-out features to compute a rich feature representation for each pixel. They proposed “squeezed zoom-out” architecture by adding squeeze module to zoom-out features, resulting in state-of-the-art performance in range of binary segmentation tasks.
\paragraph{Weakly-supervised segmentation}
Shimoda \textit{et al}~\cite{Shimoda} used zoom-out features for the task of weakly-supervised segmentation. 
\paragraph{Colorization}
Larsson \textit{et al}.~\cite{LMS:ECCV:2016} utilized zoom-out representations for the task of automatic colorization of grayscale images.
\paragraph{Aerial urban scene classification}
Santana \textit{et al}~\cite{Santana} used zoom-out features for aerial urban scene classification.
\paragraph{Optical remote sensing change detection.}
Amin \textit{et al}~\cite{Amin} utilized zoom-out architecture to extract rich superpixel descriptors for optical remote sensing change detection.

\chapter{Diverse Sampling for Weakly Supervised Learning of Semantic Segmentation} \label{chap:chapter2}

In this chapter we propose an approach for learning category-level semantic segmentation purely from image-level classification
tags indicating presence of categories. It exploits localization cues that emerge from training a modified zoom-out architecture tailored for classification tasks, to drive a weakly supervised process that automatically labels a sparse, diverse training set
of points likely to belong to classes of interest. Our approach has almost no hyperparameters,  and allows for very fast training of segmentation models. It obtains competitive results on the VOC 2012
segmentation benchmark. 

\section{Introduction}\label{sec:divsampintro}

Most recent progress
can be attributed to advances in
deep learning and to availability of large, manually
labeled data sets. However, the cost and complexity
of annotating segmentation are significantly higher than that for
classification; consequently, we have orders of magnitude more images
and categories in classification data sets such as ImageNet~\cite{russakovsky2014imagenet} or Places2~\cite{zhou2014learning}, than in
segmentation data sets, such as VOC~\cite{everingham2015pascal} or
MS-COCO~\cite{lin2014microsoft}.

Given this gap, and the objective difficulty in rapidly closing it,
many researchers have considered weakly supervised
segmentation, where the goal is still pixel-level labeling at test
time, but only spatially coarse annotations are available at training
time. Common examples of such annotations include partial
annotations, in which only a subset of pixels is labeled; bounding
boxes, where a square with an associated label is drawn around objects
of interest; and image tags, where labels provide no spatial information
and simply indicate whether or not a particular class is present
\emph{somewhere} in the image. We focus on this last, arguably
weakest, level of per-image supervision.

There is mounting evidence that this task, while difficult, is not hopeless. 
Units sensitive to object localization have been shown to emerge as part of the
representations learned by CNNs trained for
image classification~\cite{zhou2015object}. Furthermore, some
localization methods demonstrate the utility of features learned by
classification CNNs by using them to achieve competitive results~\cite{oquab2015object,zhou2015object}.

Our method is inspired by recent work~\cite{russakovsky2015what} 
demonstrating that reasonable segmenation accuracy could
be achieved with very few point-wise labels provided by human annotators. In this Chapter we propose an automatic version of this
idea, replacing human annotators with an automatic labeling
procedure. Our approach starts by learning noisy localization
networks separately for each foreground class, trained solely with image-level classification tags, and using a novel multiple
instance learning loss (global softmax, Section~\ref{sec:method}) adapted for the segmentation task. By combining the localization evidence provided by
 these networks with a novel diversity sampling procedure, we obtain a
 sparse, informative, and accurate set of labeled pixels. We can use
 these samples to
 rapidly train a fully convolutional multi-class pixel-wise label predictor
 operating on hypercolumn/zoomout representation of image
 features~\cite{hariharan2015hypercolumns,mostajabi2015feedforward} in less than 3 minutes. 

In contrast to much previous work, our approach is simple and
modular. It almost
entirely lacks hyper-parameters like thresholds and weighting
coefficients. It also allows for easy addition of new
foreground classes or incorporation of more image examples for some classes,
without the need to retrain the entire system. We also avoid complex
integration with externally trained components, other than the basic
ImageNet-pretrained neural network we use to extract pixel
features. Despite this simplicity we obtain results on the VOC 2012 data
set that improve upon
most of previous work on image-level weak supervision of
segmentation.

\section{Background}\label{sec:related}

The availability of
training data with manually annotated segmentation
masks, in particular VOC~\cite{everingham2015pascal} and recently MS-COCO~\cite{lin2014microsoft}, has been instrumental in these developments. However,
recent work has shown that training on weaker annotations, such as
partial labeling~\cite{russakovsky2015what}, object bounding boxes~\cite{papandreou2015weakly,dai2015boxsup,khoreva2016weakly} or other
cues such as object sizes~\cite{pathak2015constrained}, can
produce results very close to those using strongly supervised training.
However, closing this gap with strongly-supervised methods trained exclusively on image-level
tags, which is the regime we consider in this Chapter, remains more
challenging.

Our work was in part inspired by the experiments
in~\cite{russakovsky2015what} showing that very sparse point-wise
supervision allows training reasonable CNN-based segmentation
models. Our approach aims to replace manual annotation with
a point-wise supervision obtained automatically from image-level tags. Similarly to other recent work, we obtain
this supervision by leveraging the recently established observation:
CNNs trained for image classification tasks appear to contain internal
representations tuned to object
localization~\cite{bergamo2014self,zhou2015object,oquab2015object,cinbis2015weakly}. These
representations have been combined
with a pooling strategy to obtain objects' heat maps, which can be used
to train a segmentation model, 
often with a variant of
the Expectation-Maximization
algorithm~\cite{papandreou2015weakly,xu2015learning} or with multiple instance learning
(MIL)~\cite{pathak2014fully,pinheiro2015image,kim2016deconvolutional,kolesnikov2016seed}.

There is a body of work related to exploring and exploiting
diversity in learning and vision. Most relevant to our work is
the DivMBest algorithm~\cite{batra12diverse,yadollahpour2013discriminative} which somewhat resembles our procedure
for greedy diverse sampling of points described in
Section~\ref{sec:sampling}. The form of the diversity-infused
objective and the context are quite different, however: DivMBest is used to sample assignments in a
structured prediction settings for a single input example, whereas we
aim to sample examples (image points); in applications of DivMBest the
diverse samples are typically fed to a post-processing stage like
reranking whereas in our case, the diverse sample is directly used as
a training set for a segmentation algorithm.

Some recent methods combine image-level supervision with additional
components, such as object proposals~\cite{pinheiro2015image}, saliency~\cite{wei2015stc} or objectness
measures~\cite{russakovsky2015what}. Most of these components require
localization annotations, such as bounding boxes and/or segmentation
masks, which introduces a dependency on additional, often expensive
annotations beyond image-level tags.

In contrast, our approach is simple and modular. The modularity of our method comes from the way that we compute objects' heat maps. We train a separate classifier for each object class. This means for adding a new object class we simply train a classifier for that class without the need to retrain the entire model. Our approach does not require any external
systems, other than the initial CNN~\cite{simonyan2014very} pretrained on the ImageNet
classification task~\cite{russakovsky2014imagenet}, making the
entire pipeline independent of any requirements beyond image-level annotations.

The most established
benchmark for this task is VOC 2012 dataset, with the
standard performance measure being intersection over union averaged
over 21 classes (mIoU). In
Section~\ref{sec:exp} we show that despite its simplicity and
efficiency, our approach
outperforms most competing methods on this benchmark.

\section{Automatic point-wise supervision for segmentation}\label{sec:method}
The basis of our bootstrapping method is the localization maps
obtained for each of the foreground classes. These maps are sampled
for each class, and the resulting sparse set of
point-wise labels on the training images is used to train the final
segmentation network. We describe these steps in detail below.

\subsection{Learning localization with image-level tags}\label{sec:localization}
 We start by extracting an
image feature map using a pretrained fully convolutional
network. Then for each foreground class $c$, we 
construct a per-location localization network on top of these features, which
outputs two scores per-location $i$: $S(c,i)$ for foreground, and
$\bar{S}(c,i)$ for background (which in this case means anything
other than the foreground class at hand). 

An obvious next step now is
to convert these pixel-wise scores into image-level foreground/background probability score, using
some sort of pooling scheme; this can then be used to compute
image-level classification log-loss and backpropagate it to the
localization network. We consider two such schemes.

\paragraph{The per-pixel softmax model} We can convert the scores into
per-location posterior probabilities using the standard softmax model, and apply max pooling over the
resulting probability map:
\begin{equation}
  \label{eq:softmax}
p(c)\,=\,\max_i\frac{\exp S(c,i)}{\exp S(c,i)\,+\,\exp \bar{S}(c,i)}.
\end{equation}
This can be interpreted as requiring that for images containing
foreground class, the network assign at least one location high probability
while for images without the foreground, all locations must have low
probability. The background scores $\bar{S}$ do not have a direct
meaning other than to normalize the probabilities. 

\paragraph{The global softmax model} An alternative is to apply
max-pooling \emph{separately} for the two score maps, and convert the
maxima to the image-level probability:
\begin{equation}
  \label{eq:globalsoftmax}
  p(c)\,=\,\frac{\max_i \exp S(c,i)}{\max_j \exp S(c,j)\,+\,\max_l\exp
    \bar{S}(c,l)}.
\end{equation}
This model no longer is equivalent to the per-location softmax, and in
fact does not provide per-location probability map. It specifically
encourages the background scores to be high for images without the
foreground. It also routes the gradient of the loss via two locations
in each image, instead of one with~\eqref{eq:softmax}, and therefore may facilitate faster training.

It is worth noting that this approach does not include an explicit
``background localization'' model. Background is defined separately
for each foreground class as its complement, and jointly as the
complement of all foreground classes. Adding another foreground class
would require only training one new localization model for that class;
the definition of segmentation background would then automatically be
updated, and reflected in the sampling process described below.

\subsection{Sampling strategies}\label{sec:sampling}
We now consider the goal of translating class-specific score maps to
supervisory signal for semantic segmentation. 

\begin{figure}[H]
  \centering
\begin{tabular}{c}
 \includegraphics[width=0.4\linewidth]{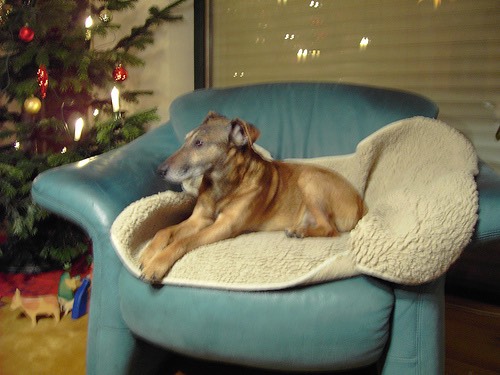}\\
\includegraphics[width=0.9\linewidth]{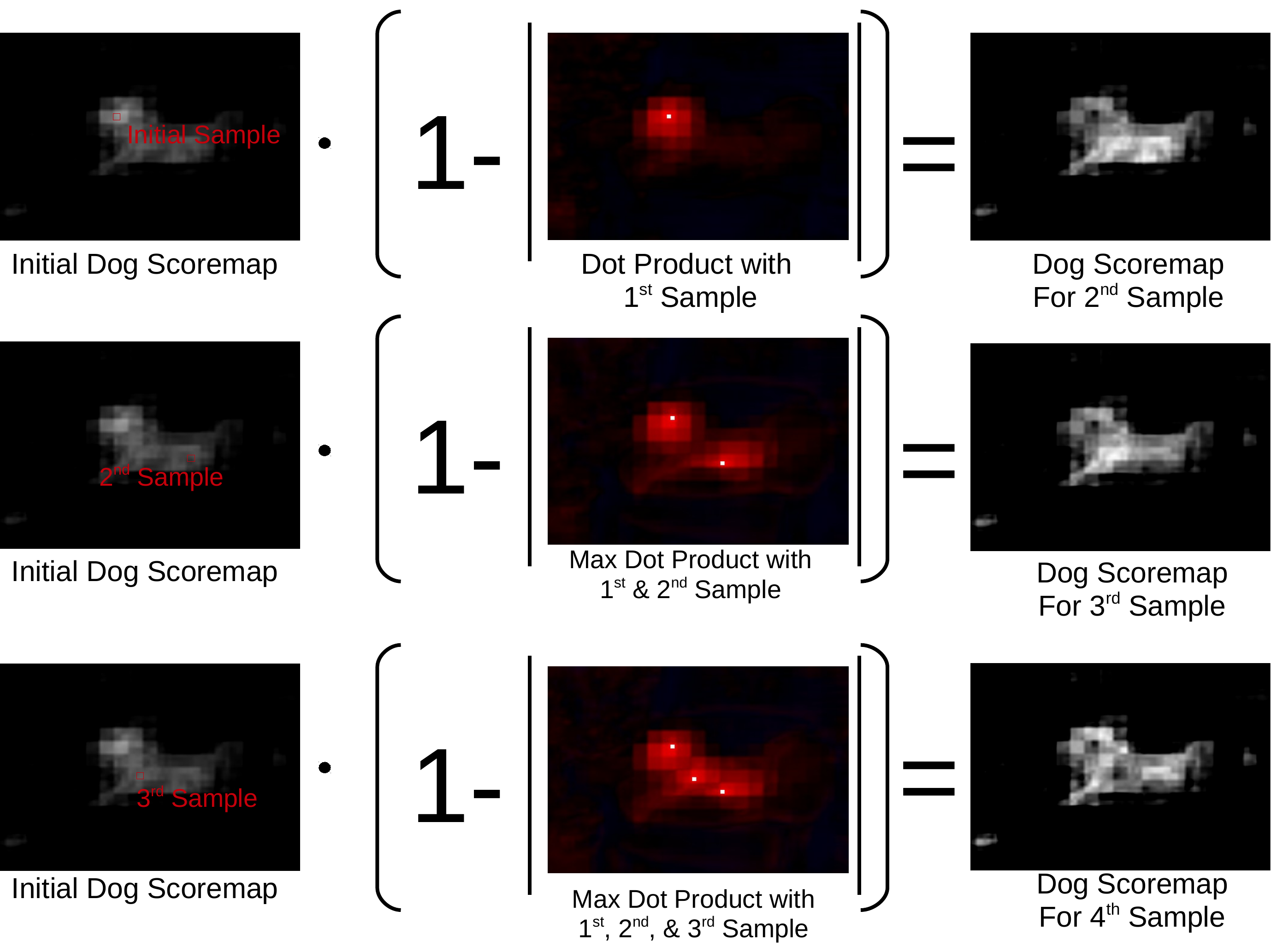}\\
\includegraphics[width=0.4\linewidth]{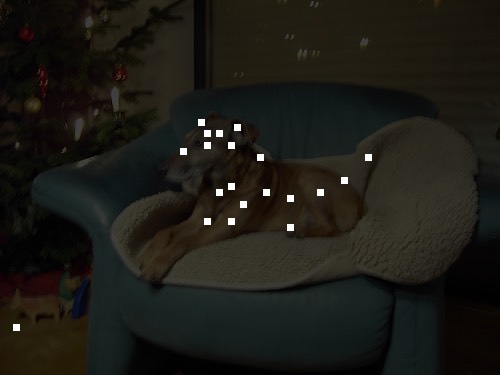} 

\end{tabular}
\caption{Our diverse sampling procedure. Left: input image, labeled as containing class
  \texttt{dog}. Middle: first four steps of diverse sampling procedure
  for \texttt{dog}, starting with the map
  $S(\cdot,\text{\texttt{dog}})$. Right: sample of 20 diverse
points.}
\label{fig:divsample}  
\end{figure}

Our general framework
for this will be to select a sparse set of locations from the training
images, for which we will assign class labels. The segmentation
predictor is
then trained by learning to map the image features for these
locations to the assigned class labels.

Let $S(i,c)$ be the score at spatial index $i$
for a class $c$ produced by our image-level classification
model. One approach would be to densely label $y_i=
\argmax_{c}S(i,c)$. Background requires a separate treatment: it is
not a ``real'' class, rather it's defined by not being one of the
foreground classes. Hence, we do not have a separate model for it, and
instead can assign background labels to pixels in which no foreground
class attains a sufficient score: $y_i=bg$ if $\max_cS(i,c)<\tau$. 
This simplistic strategy has two problems: (1)
 some classes may have systemically lower scores than others, and
(2) it is unclear how to optimally set the value of $\tau$.

However, our hypothesis is that while the scoremaps provide only coarse
localization, and an inconsistent level of confidence across images
and classes, the maximum activations of a class scoremap when that
class is present appear to reliably correspond with pixels containing
the correct class. (We verified this qualitatively, on a few classes and
a number of
training images). So, an alternative approach is to label as $c$ the
$k$ locations
corresponding with highest scores for class $c$. However the size of objects varies widely
across images, and it isn't clear what $k$ should be. If $k$ is too
high, the labels will be very noisy. If $k$
is too low,  most of the pixels will be tightly clustered 
portions of each class, e.g.,  wheels of cars, or
 faces of people; training on such examples is much less effective
because many of the samples will highly correlated. 

The method we propose here alleviates these problems by relying on
\emph{diversity sampling}. Let $\mathbf{z}_i$ be the image feature vector at spatial
index $i$, normalized to unit norm, and let $F$ be the set of foreground classes present in the
image. For each class $c\in F$ we define
the  $k$th sampled location $x^c_k$ from that image by induction:
\begin{eqnarray}\label{eq:div-fg}
  x^c_1\,&=\,\argmax_i S(i,c),\\
  x^c_k\,&=\,\argmax_i\,\left\{S(i,c)\left[1-\max_{k'<k}
  \left| \mathbf{z}_i\cdot\mathbf{z}_{x^c_{k'}} \right|\right]\,\right\}.
\end{eqnarray}
where $\cdot$ denotes the dot product.
In other words, we aim to select points with the highest scores for
class $c$, but penalize them for similarity to
previously selected points for that class. This encourages selection of a set of examples
that jointly provides more information (due to diversity), and is likely
to be accurate (due to high scores). 

The definition of similarity as
dot product in feature space, rather than as spatial
proximity, is important. If features at two points 
are similar, they will appear as similar to the eventual pixel classifier
(since it operates on those features), regardless of how close they
are in the image; labeling only one
of them is sufficient. On the other hand, two points nearby may have
very different feature representations and thus it is beneficial to label both to give the pixel
classifier more information.

This approach also naturally leads to a threshold-free method for
selecting background points. We let the $k^{th}$ sample, $x^{bg}_k$
for be defined as:
\begin{equation}
  \label{eq:div-bg}
  x^{bg}_k\,=\,\argmin_i \max\left\{\,
    \max_{c\in F,k'}\left|\mathbf{z}_i\cdot\mathbf{z}_{x^c_{k'}}\right|\,,\,
    \max_{k''<k}\left|\mathbf{z}_i\cdot\mathbf{z}_{x^{bg}_{k''}}\right|
    \right\}
\end{equation}
where $k'$ ranges over indices of selected points for each foreground
class, and $k''$ ranges over points selected so far for the
background. This simply searches at each step for the image location
most dissimilar to any foreground points to maximize the chance
of correctly identifying background points, and to any other
background points selected so far to maximize diversity. Our diversity
sampling strategy is illustrated in Figures~\ref{fig:divsample},~\ref{fig:sampling}.

\subsection{Training semantic segmentation using point-wise supervision}
Once the automatically point-wise labels are obtained, we are ready to train
the segmentation as a per-pixel fully convolutional multi-class
(including background) classifier
network, with receptive field of 1$\times$1. This can be done by using
the standard convnet training machinery, with zero-masking applied to
the loss in locations where no labels are available; whether to
fine-tune the underlying network that extracts the visual features per
location is a choice.

\begin{figure*}[th]

    \setlength{\tabcolsep}{0.25em}
  \centering
  \begin{tabular}{c   c   c    c}
  \centering
\includegraphics[width=.22\textwidth]{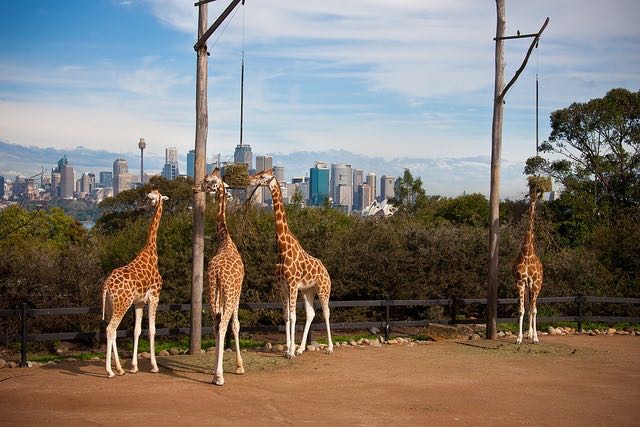} &
  	\includegraphics[width=.22\textwidth]{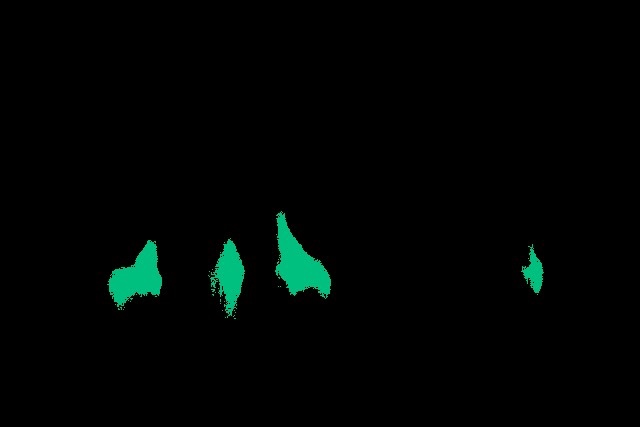}&
	\includegraphics[width=.22\textwidth]{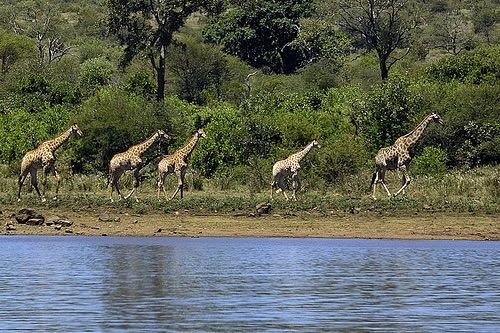} &
    \includegraphics[width=.22\textwidth]{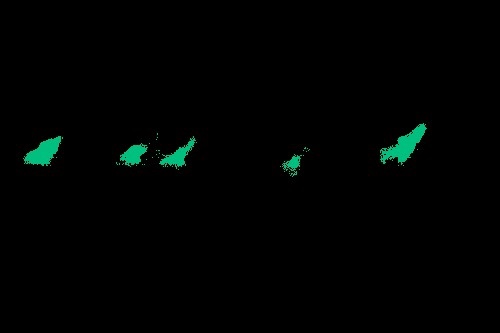}\\
\includegraphics[width=.22\textwidth]{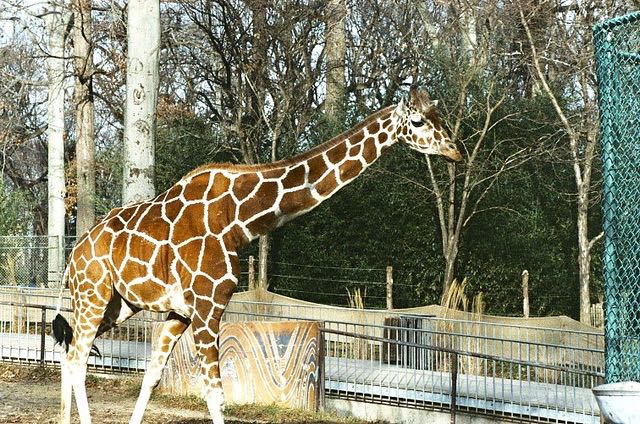} &
  	\includegraphics[width=.22\textwidth]{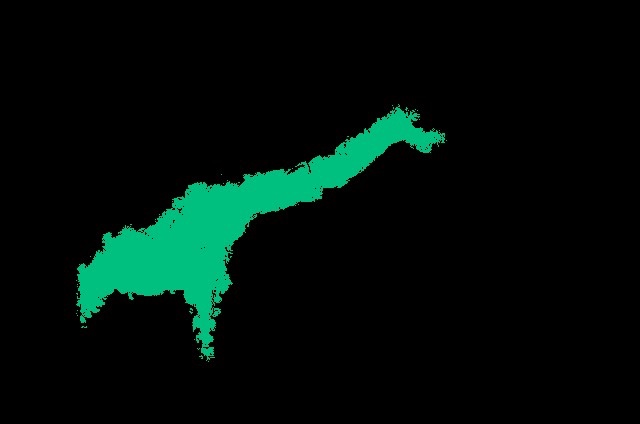}&
	\includegraphics[width=.22\textwidth]{figures/SegextraImsupdate/COCO_train2014_000000018792.jpg} &
  	\includegraphics[width=.22\textwidth]{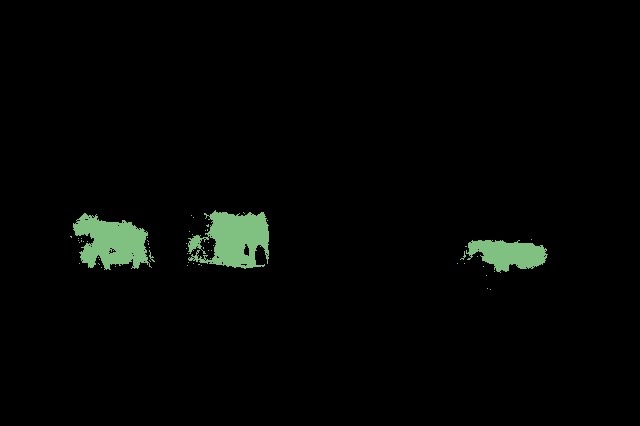}\\

\includegraphics[width=.22\textwidth]{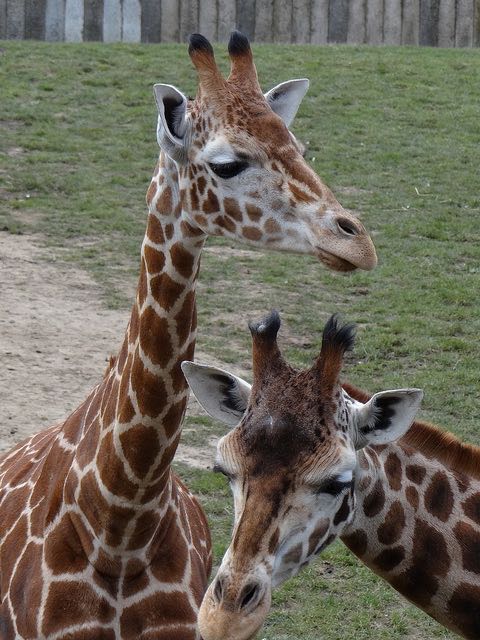} &
  	\includegraphics[width=.22\textwidth]{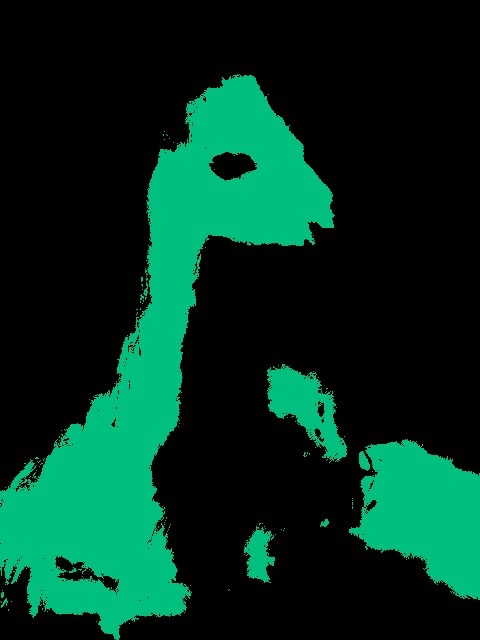}&
	\includegraphics[width=.22\textwidth]{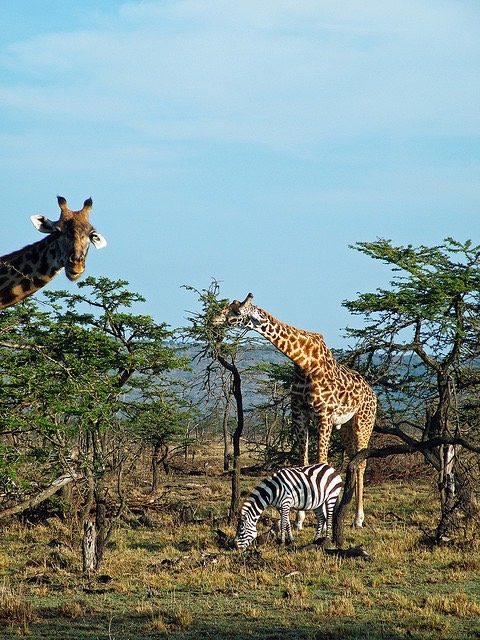} &
  	\includegraphics[width=.22\textwidth]{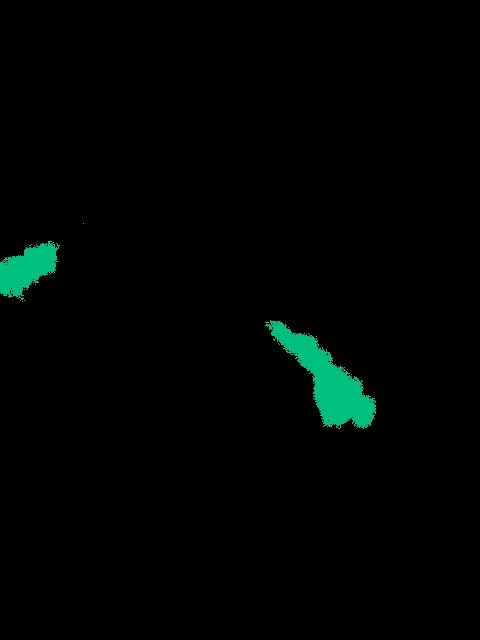}\\

\includegraphics[width=.22\textwidth]{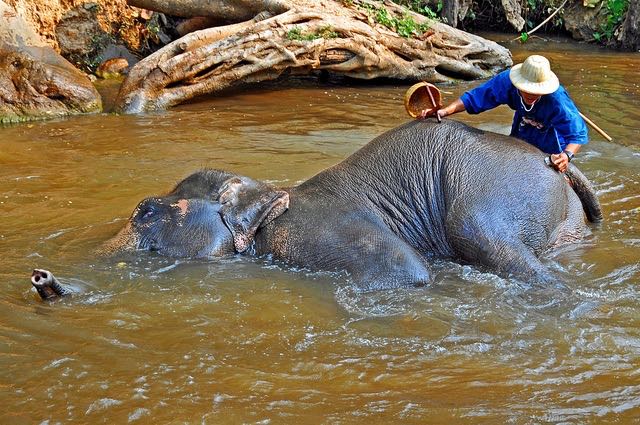} &
 	 \includegraphics[width=.22\textwidth]{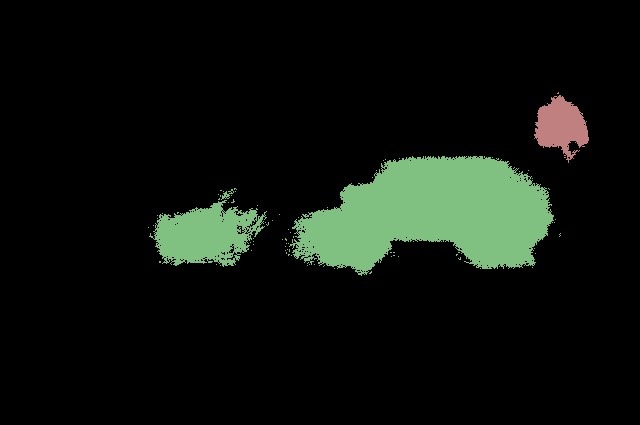}&
	\includegraphics[width=.22\textwidth]{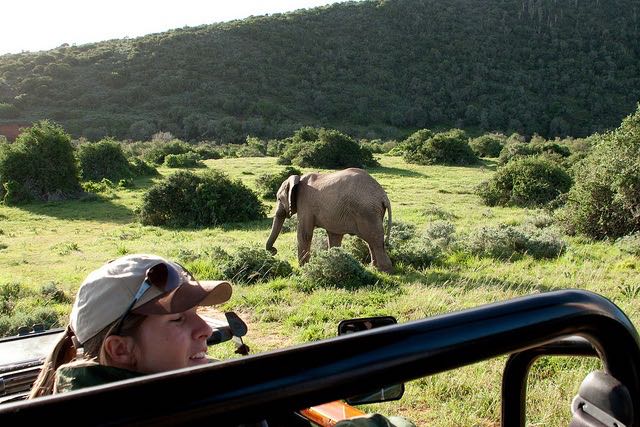} &
  	\includegraphics[width=.22\textwidth]{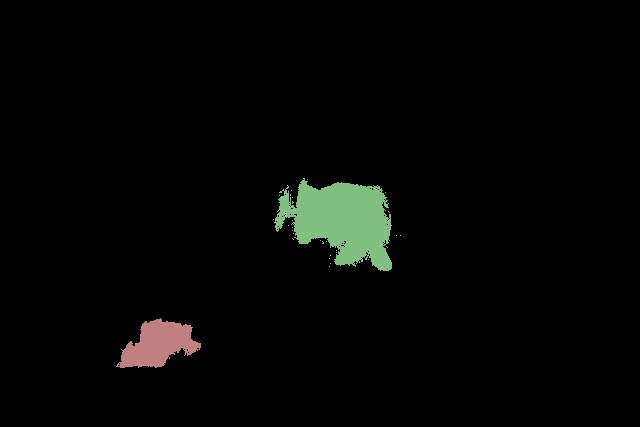}\\
\includegraphics[width=.22\textwidth]{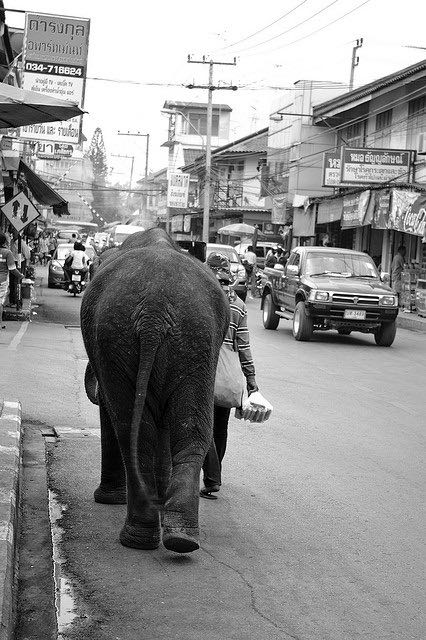} &
  	\includegraphics[width=.22\textwidth]{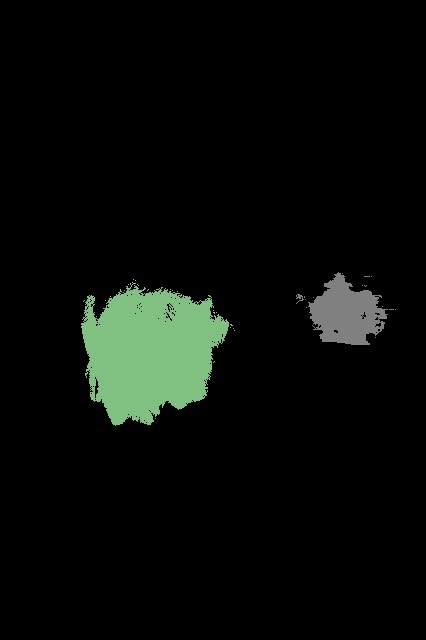}&
	\includegraphics[width=.22\textwidth]{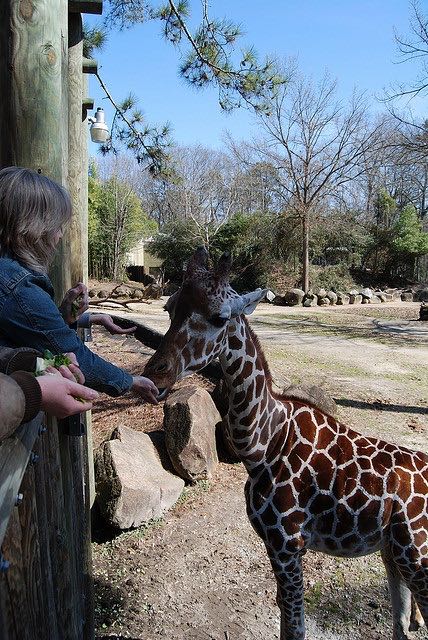} &
  	\includegraphics[width=.22\textwidth]{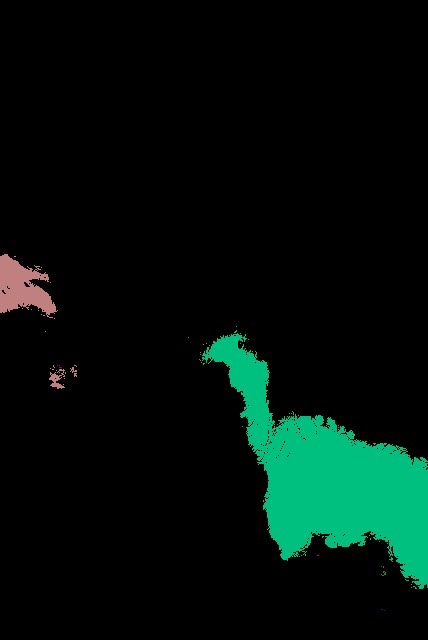}\\

 \includegraphics[width=.22\textwidth]{figures/SegextraImsupdate/COCO_train2014_000000045280.jpg} &
   	\includegraphics[width=.22\textwidth]{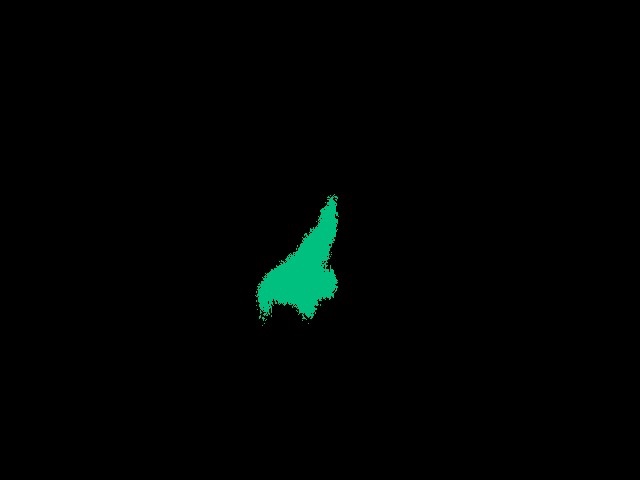}&
	\includegraphics[width=.22\textwidth]{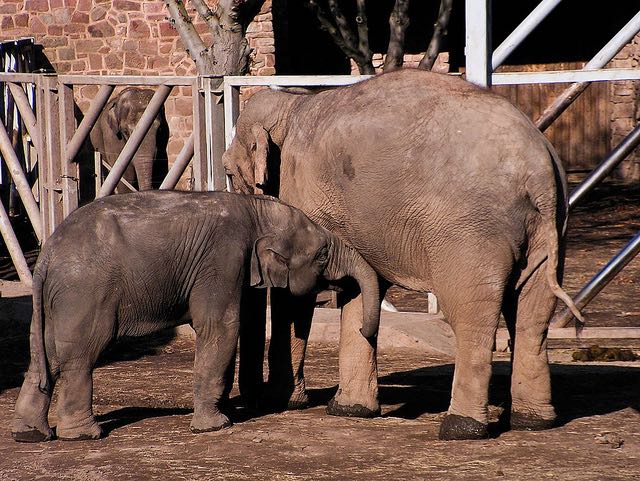} &
   	\includegraphics[width=.22\textwidth]{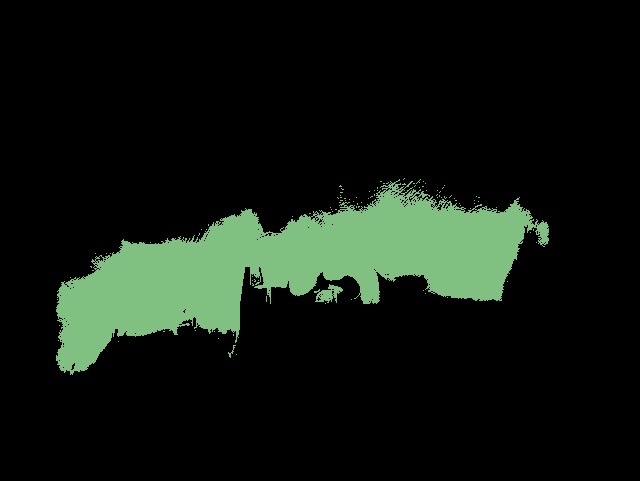}
  \end{tabular}
  \captionof{figure}{Example segmentations on extra classes from MS-COCO dataset added to the segmentation model trained on VOC dataset.}
  \label{fig:extraseg}

\end{figure*}

\newcommand{\setupSampIm}[3]{ input & #2 & #3 &  background\\  \includegraphics[width=.19\textwidth]{figures/SampIms/#1}&  \includegraphics[width=.19\textwidth]{figures/SampIms/#1_#2_inMap}&  \includegraphics[width=.19\textwidth]{figures/SampIms/#1_#3_inMap}&   \includegraphics[width=.19\textwidth]{figures/SampIms/#1_background_inMap}\\   &\includegraphics[width=.19\textwidth]{figures/SampIms/#1_#2_output}&  \includegraphics[width=.19\textwidth]{figures/SampIms/#1_#3_output}&  \includegraphics[width=.19\textwidth]{figures/SampIms/#1_background_output}\\ }

\begin{figure*}[!th]
  \centering
  \begin{tabular}{cccc}
\setupSampIm{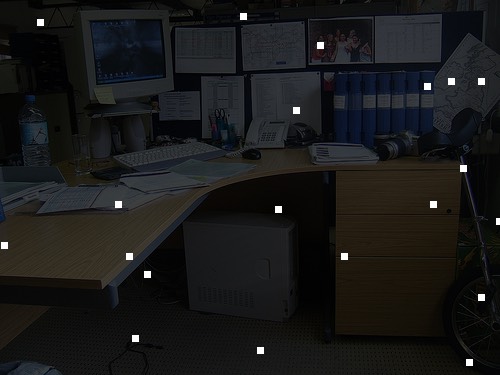}{bottle}{tvmonitor}
\setupSampIm{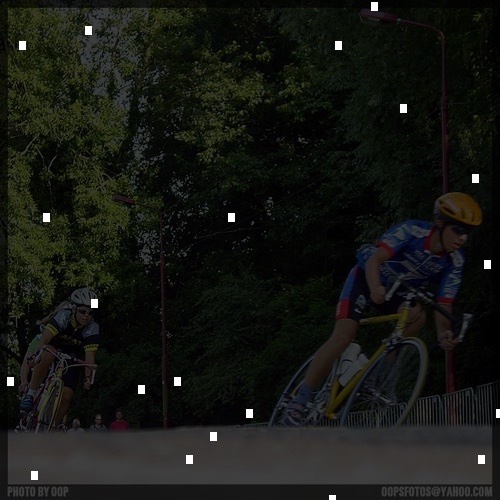}{person}{bicycle}
\setupSampIm{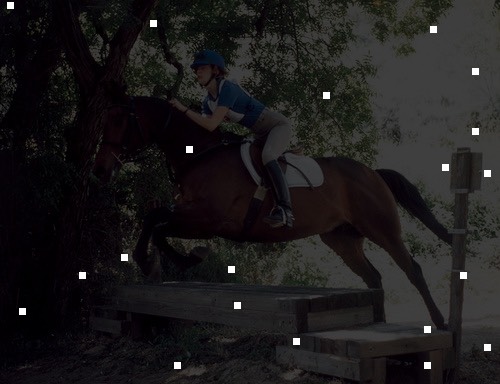}{person}{horse}
  \end{tabular}
  \caption{Examples of diverse sampling outputs. For each foreground class, we
    show the localization score map from global softmax model, and the
  selected 20 points. For background, the map shows the max over dot
  products with any selected foreground points.}
  \label{fig:sampling}
\end{figure*}

\section{Experiments}\label{sec:exp}
In order to compare our method to other work on segmentation, we
conduct all of our experiments on the VOC 2012 data set. For training
images (10,582 images in the train-aug set) we discard all annotations except for image-level labels indicating which of
the 20 foreground classes is present in each image. We evaluate
various versions of our method, as well as its components
individually, on the validation set, and finally use the models chosen based
on these experiments to obtain results on the test set from the VOC
evaluation server. All experiments were done in Torch7, using Adam~\cite{kingma2015adam}
update rule when training networks.

\subsection{Experimental setup}
\paragraph{Pixel features}
As the base CNN we use the VGG-16 network trained on ImageNet and
publicly available from the authors of~\cite{simonyan2014very}, from
which we remove all layers above pool.
This network is run in the fully convolutional mode on input images
resized to 256$\times$336 pixels. Then each of the 13 feature maps
(outputs of all convolutional layers, with pooling applied when
available) is resized to 1/4 of the input resolution, and concatenated
along feature dimensions. This produces a tensor in which each
location on the coarse 64$\times$84 grid has a 4,224-dimensional
feature vector. This closely follows the zoom-out extraction
protocol in~\cite{mostajabi2015feedforward}, but without superpixel pooling.

When computing dot products in diversity
sampling~\eqref{eq:div-fg},\eqref{eq:div-bg} we normalize zoom-out feature vectors
to unit norm in two stages: each feature dimension is normalized to be
zero-mean, unit variance over the entire
training set, then each feature vector is scaled to be unit Euclidean norm.
\paragraph{Localization models} For each class, the fully
convolutional localization
network consists of a 1$\times$1 convolutional layer with 1024 units,
followed by ReLU and the 1$\times$1 convolutional layer with 2 units,
which outputs the score maps $S$ for the foreground and $\bar{S}$ for background. At
training time, for the global softmax model~\eqref{eq:globalsoftmax}
this is followed by global max pooling layer and the softmax
layer, while for the per-pixel softmax~\eqref{eq:softmax} the order of
softmax and max pooling is reversed. 

For each class, we train the network on all positive examples for that
class (images that contain it) and a randomly sampled equal
number of negative examples, with
batch size of 1 image, learning rate $10^{-4}$ which after 2 epochs is
decreased to $10^{-5}$ for one additional epoch
and momentum 0.9.

We experimented with adding higher layer features (fc7 from
VGG-16) to the input to localization networks, but found that it makes
localization worse: it is too easy for the network to determine
presence of objects from these complex, translation invariant
features. We do however bring these features back when training the
final segmentation model, described next.
\paragraph{Segmentation model}
To provide image-level priors, which have been reported to improve
segmentation results in both fully supervised~\cite{mostajabi2015feedforward} and weakly
supervised~\cite{pinheiro2015image} settings, we augment the zoom-out feature map with the 
global features (layer fc7 of VGG-16, pooled over the entire
image and replicated for all locations). The combined
feature map (8320 features per location) is fed to a 1$\times$1
convolutional layer with 512 units, followed by ReLU, and the 21-channel
prediction layer, followed by the softmax loss layer. We train this
network on the selected set of points pooled over all training images, using batch size 100 (note:
this means 100 sample points, not 100 images!), fixed learning rate
$10^{-6}$, and momentum 0.9, for two epochs. With these settings,
typical time to train the final segmentation model is less than 3 minutes
on a single Titan X GPU.

\subsection{Evaluation of model components}
We start by evaluating components of our approach on the validation set.
\paragraph{Localization model} As shown in Table~\ref{tab:locmodels},
the global softmax model~\eqref{eq:globalsoftmax} obtains
significantly better results than the per-pixel
softmax~\eqref{eq:softmax}. Therefore we choose it for all the
subsequent experiments.

\begin{table}[H]
\centering
\setlength\tabcolsep{4pt}
 \begin{tabular}[c]{| l | l |}
\hline
    Model  &  mIoU\\
\hline
Pixel softmax  &  38.0 \\
Global softmax  &  40.6\\
\hline
  \end{tabular}
  \captionof{table}{Comparison of localization models on VOC 2012 validation  set}
  \label{tab:locmodels}
\end{table}
\paragraph{Bootstrapping by localization maps} We could attempt using the score maps obtained by localization
network directly as the predicted segmentation maps. Specifically, we
considered assigning each pixel to the highest scoring class (after
normalizing the scores so that for each class on average images with
the foreground present have the highest score of 1), or to the
background if the highest foreground score is below
threshold. This results in poor segmentation accuracy: the highest
validation mIoU after sweeping the
threshold values was 25.66, for threshold of 0.2. 

We also made an attempt to use
these score-based segmentation maps as the source of bootstrapping directly,
without sampling. That is, we can train the segmentation network in
the usual fully-supervised way, giving it the score map-based
dense segmentation labels as if it were ground truth. The results were
very poor; in large fraction of the pixels the localization models are
uncertain, and densely localized labels will force the segmentation model to make a
decision in those uncertain points as well, leading to a very noisy labeling.
\paragraph{Effect of sampling strategy}
For our diverse sampling method, we need to set the value of
$k$. Table~\ref{tab:miou-vs-k} shows the effect this value has on VOC 2012
validation mean IoU. The optimal $k$ among those tested is 20, but the
behavior is stable across a large range of values. 

We also compared alternative sampling strategies, namely selecting the
top $k$ scoring points for each class, or using diversity but in
spatial domain instead of in feature domain. Table~\ref{tab:sampling}
shows that the diversity sampling using feature similarity is indeed superior
to those. Figure~\ref{fig:sampling} shows a few qualitative examples
of diverse sampling outputs. Notably, sampling for background is
usually quite accurate, even though it is oblivious to the actual
class scores and is entirely driven by diversity with respect to the
foreground and within background.

\begin{table}[H]
\centering
\setlength\tabcolsep{4pt}
  \begin{tabular}{|l|l|l|l|l|l|}
\hline
    $k$&1 & 5&10&20 & 50\\\hline
    mIoU & 35.1& 37.2&39.3  &  40.6 & 40.4\\
\hline
  \end{tabular}
    \captionof{table}{mIoU on VOC 2012 validations set as a function of $k$ points in
      diverse sampling, with global softmax model}
    \label{tab:miou-vs-k}

\end{table}

\begin{table}[H]
\centering
\setlength\tabcolsep{4pt}
  \begin{tabular}{| l | l |}
\hline
    Sampling  &  mIoU\\
\hline
 Dense &  15.0\\
Spatial &  33.4\\
Top $k=20$ &  30.7\\
 Diverse, $k=20$ &  40.6\\
\hline
  \end{tabular}
  \captionof{table}{Comparison of point sampling strategies}
  \label{tab:sampling}

\end{table}

\setlength{\tabcolsep}{4pt}

\newcommand{\setupSegEx}[1]{\includegraphics[width=.18\textwidth]{figures/SegIms/#1} &%
\includegraphics[width=.18\textwidth]{figures/SegIms/#1_GT}&%
\includegraphics[width=.18\textwidth]{figures/SegIms/#1_base_heatmap}&%
\includegraphics[width=.18\textwidth]{figures/SegIms/#1_our_model}&%
\includegraphics[width=.18\textwidth]{figures/SegIms/#1_our_model_crf}\\}

\begin{figure*}[!th]
  \centering
  \begin{tabular}{ccccc}
    input & GT & loc-map & diverse 20 & diverse 20 with CRF\\
\setupSegEx{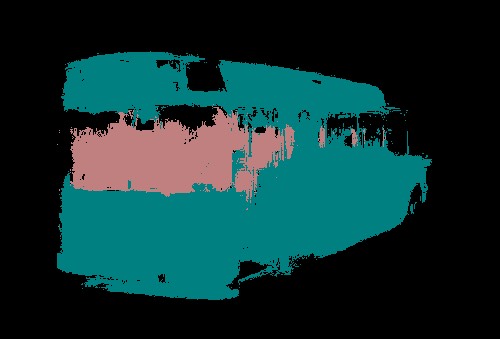}
    \setupSegEx{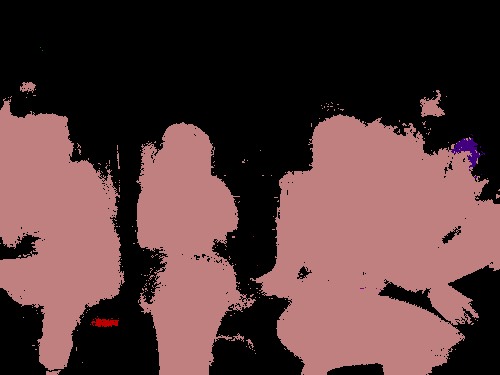}
    \setupSegEx{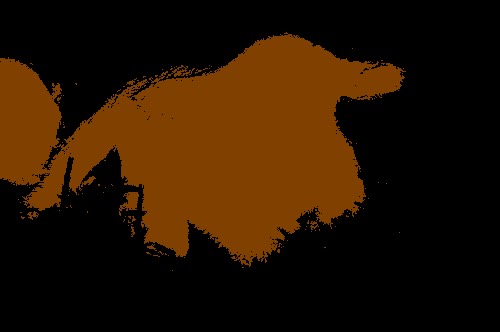}
    \setupSegEx{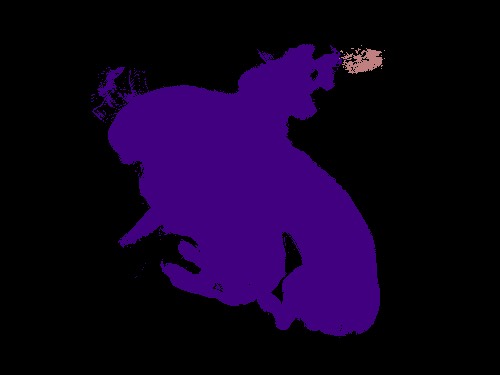}
\setupSegEx{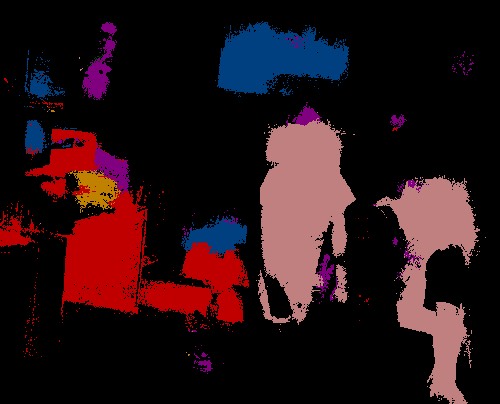}
\setupSegEx{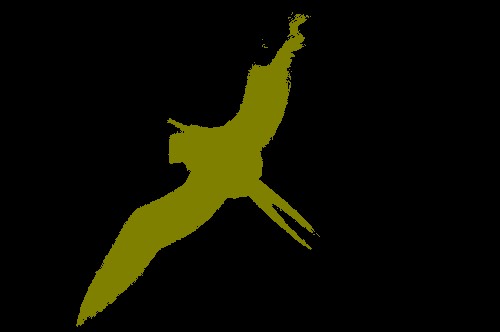}
\setupSegEx{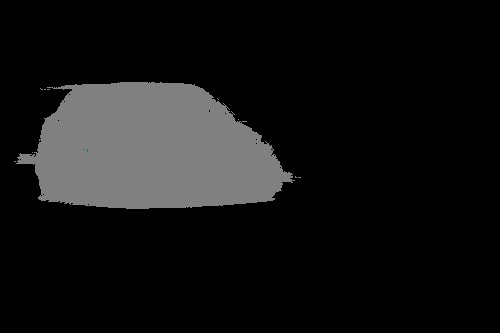}
\setupSegEx{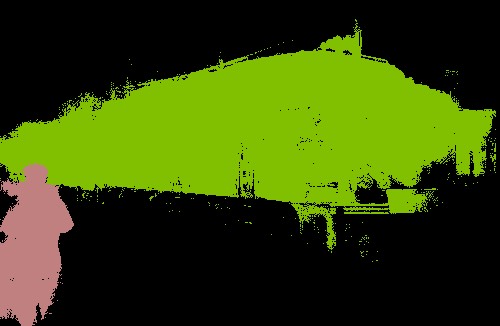}
  \end{tabular}
  \caption{Examples of segmentations learned through our
    weakly supervised approach. From left: input image, ground truth,
    thresholded localization score maps, segmentation learned with diverse $k=20$ sampling, CRF post-processing.}
  \label{fig:examples}
\end{figure*}

\subsection{Final segmentation results}

Based on the preliminary experiments on the validation above, we identify
the optimal setting for our approach: global softmax localization model, with
diverse sampling, $k=20$. We report the results of
this method in Table~\ref{tab:dvcomparison}. We also report the results of applying fully connected CRF ~\cite{koltun2011efficient} with author's default parameters on top of our predictions.



\begin{table}[H]
\setlength\tabcolsep{4pt}
  \centering
  \footnotesize
  \begin{tabular}{|l|l|l|l|}
    \hline

Method & VOC2012 validation  & VOC2012 test  & comments\\
\hline
\textcolor{black}{DeepLab-CRF\cite{chen2015semantic}} & \textcolor{black}{68.7} &
                                                     \textcolor{black}{71.6}
                                        & \textcolor{black}{fully supervised}\\
\textcolor{black}{FCRN\cite{wu2016high}} & \textcolor{black}{74.8} &
                                                     \textcolor{black}{77.3}
                                        & \textcolor{black}{fully supervised}\\
\textcolor{black}{BoxSup\cite{dai2015boxsup}} & \textcolor{black}{62.0} &
                                                     \textcolor{black}{64.6}
                                        & \textcolor{black}{bounding box-supervised}\\

\textcolor{black}{Bbox-Seg\cite{papandreou2015weakly}} & \textcolor{black}{60.6} &
                                                     \textcolor{black}{62.2}
                                        & \textcolor{black}{bounding box-supervised}\\
\textcolor{black}{1 point~\cite{russakovsky2015what}} &
\textcolor{black}{35.1} &
                                                     \textcolor{black}{}&
\textcolor{black}{manual annotation, 1 pt/class}\\
\textcolor{black}{1 point+Obj~\cite{russakovsky2015what}}&
\textcolor{black}{42.7} &
                                                     \textcolor{black}{}&
\textcolor{black}{+ objectness prior}\\
\textcolor{black}{STC\cite{wei2015stc}} & \textcolor{black}{49.8} &\textcolor{black}{51.2} & \textcolor{black}{externally trained saliency model}\\
\hline
MIL-sppx~\cite{pinheiro2015image} & 36.6 & 35.8 & superpixel smoothing\\
CCCN~\cite{pathak2015constrained} & 35.3 & 35.6 &\\
EM-Adapt~\cite{papandreou2015weakly} & 38.2 & 39.6 &\\
SEC\cite{kolesnikov2016seed}& 50.7 & 51.5 &\\
Ours & 40.6& 41.2& no post-processing\\
Ours & 45.2 & 46.0  & CRF post-processing \\
\hline
  \end{tabular}
  \caption{Comparison of competitive segmentation methods, supervised with
    image-level tags. For reference, top of the table includes methods trained
  with stronger supervision on VOC 2012 data, or on additional
  data. 
}
  \label{tab:dvcomparison}
\end{table}
The top portion of the table contains representative results for
stronger supervision scenarios, for reference; these are not directly
comparable to our results. Among the methods trained on the same data and in
the same regime as ours, our results are the highest. It is
interesting that we obtain similar results to those with
\emph{manual} annotation of a single point per class per image~\cite{russakovsky2015what} (better than their results without
objectness prior),
although our point selection is fully automatic.

STC~\cite{wei2015stc}, achieves test mIoU of 51.2. It is trained
on 40,000 additional images, collected in a carefully designed
procedure to make them easy to learn from. STC also uses an externally
trained saliency mechanism, which requires mask annotations to
train. 

SEC~\cite{kolesnikov2016seed} is trained solely on VOC data which has higher accuracy, achieving 51.5 mIoU on the test set. However, their approach is considerably more complex than ours,
employing hyperparameters that determine various thresholds, and the
time to train the final segmentation system with SEC is almost two
orders of magnitude higher than ours. SEC also uses significantly larger field of view for the
underlying segmentation network than in our experiments (378 vs. 224
for us), and results reported in~\cite{kolesnikov2016seed} suggest
that this may be very important.
\paragraph{Adding new object classes on the fly} 
One of the key characteristics of our model is modularity. Consider that we want to add new classes like Giraffe and Elephant which are not part of VOC dataset. We train localization model for new classes with only image level tags from MS-COCO dataset for Giraffe and Elephant. Since we have trained the localization model for each class separately, there is no need to re-train the other classes localization models. The segmentation training data for new classes is the sparse set of diverse points extracted from localization model output. Sparsity significantly speeds the segmentation training and diversity leads to high quality segmentation output. It takes less than 3 minutes to re-train the segmentation model with additional classes, without hurting its accuracy on VOC segmentation benchmark. Hence, it is practical to add new classes on the fly. Qualitative examples of segmentation results for the new classes are shown in Figure~\ref{fig:extraseg}.

\section{Summary}\label{sec:conclusions}

We have proposed an approach to learning category-level semantic
segmentation when the only annotation available is tags indicating the
presence (or absence) of each foreground class in each image. Our approach is based on a form of
bootstrapping, in which a sparse, visually diverse set of points in
training images is
labeled based on class-specific localization maps, predicted from the
image tags. 

Among the appealing properties of our method are its simplicity,
near absence of hyperparameters (and insensitivity to the only
hyperparameter, the sample size), modularity (easy to
update the model with new classes and/or examples), lack of reliance
on complex external components requiring strong supervision, and last
but not least, competitive empirical performance and speed.

\chapter{Label Embedding}\label{chap:labelemb}
In the first part of this chapter we introduce custom regularization functions for use in
supervised training of deep neural networks. Our technique
is applicable when the ground-truth labels themselves exhibit internal structure; we derive a regularizer by learning an autoencoder over the set of annotations. In the second part of this chapter, we explore how well we can predict depth maps directly from semantic segmentation maps and propose a two-part architecture for joint prediction of depth and semantic segmentation.

\section{Introduction}
\label{sec:introduction}

\begin{figure}[!th]
   \begin{center}
      \includegraphics[width=1\linewidth]{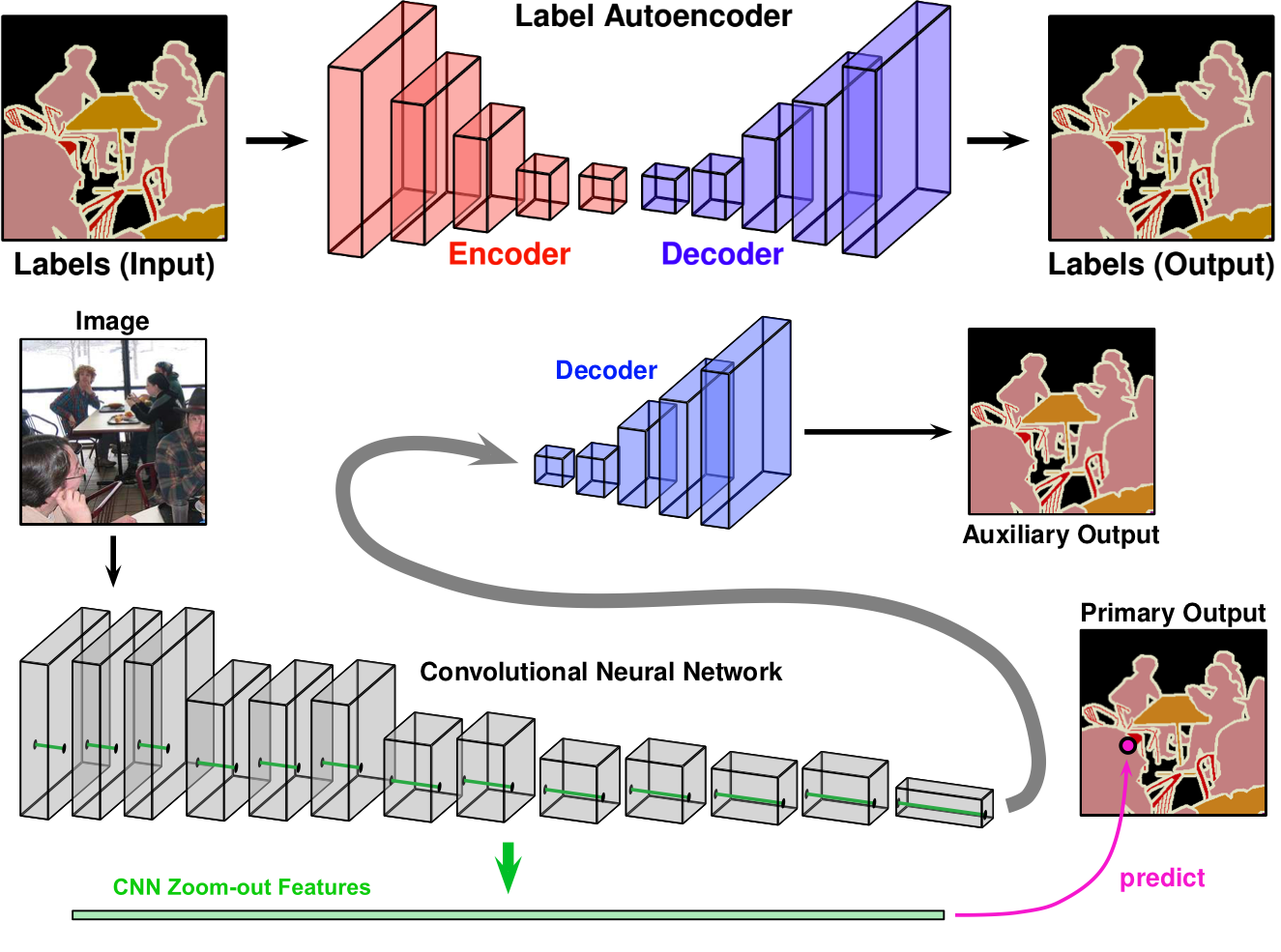}
   \end{center}
   \vspace{-0.01\linewidth}
   \caption{
      Exploiting label structure when training semantic segmentation.
      Top:
      An initial phase looks only at the ground-truth annotation of training
      examples, ignoring the actual images. We learn an autoencoder that
      approximates an identity function over segmentation label maps.  It is
      constrained to compress and reconstitute labels by passing them through
      a bottleneck connecting an encoder (red) and decoder (blue).
      Bottom:
      The second phase trains a standard CNN
      for semantic segmentation using zoom-out features for per-pixel output.  However, we attach an auxiliary branch
      (and loss) that also predicts segmentation by passing through the decoder
      learned in the first phase.  After training, we discard this decoder
      branch, making the architecture appear standard.
   }
   \label{fig:overview}
   \vspace{-0.015\linewidth}
\end{figure}

The recent successes of supervised deep learning rely on the availability of
large-scale datasets with associated annotations for training.  In computer
vision, annotation is a sufficiently precious resource that it is commonplace
to pretrain systems on millions of labeled ImageNet~\cite{ImageNet} examples.
These systems absorb a useful generic visual representation ability during
pretraining, before being fine-tuned to perform more specific tasks using
fewer labeled examples.

Current state-of-the-art semantic segmentation methods~\cite{DeepParsing,
DeepLab,PSPNet} follow such a strategy.  Its necessity is driven by the high
relative cost of annotating ground-truth for spatially detailed
segmentations~\cite{PASCAL,COCO}, and the accuracy gains achievable by
combining different data sources and label modalities during training.  A
collection of many images, coarsely annotated with a single label per image
(ImageNet~\cite{ImageNet}), is still quite informative in comparison to a
smaller collection with detailed per-pixel label maps for each image (PASCAL~\cite{PASCAL} or COCO~\cite{COCO}).

We show that detailed ground-truth annotation of this latter form contains
additional information that existing schemes for training deep convolutional
neural networks (CNNs) fail to exploit.  By designing a new training procedure,
we are able to capture some of this information, and as a result increase
accuracy at test time.

Our method is orthogonal to recent efforts, discussed in
Section~\ref{sec:related}, on learning from images in an unsupervised or
self-supervised manner~\cite{
CtxEncoder,
Jigsaw,
ZIE:ECCV:2016,
LMS:ECCV:2016,
LMS:CVPR:2017,
ZIE:CVPR:2017,
DKD:ICLR:2017}.
It is not dependent upon the ability to utilize an external pool of data.
Rather, our focus on more efficiently utilizing provided labels makes our
contribution complementary to these other learning techniques.  Experiments
show gains both when training from scratch, and in combination with pretraining
on an external dataset.

Our innovation takes the form of a regularization function that is itself
learned from the training set labels.  This yields two distinct training
phases.  The first phase models the structure of the labels themselves by
learning an autoencoder.  The second phase follows the standard network
training regime, but includes an auxiliary task of predicting the output via
the decoder learned in the first phase.  We view this auxiliary branch as a
regularizer; it is only present during training.  Figure~\ref{fig:overview}
illustrates this scheme.

Section~\ref{sec:method} further details our approach and the intuition behind
it.  Our regularizer can be viewed as a requirement that the system understand
context, or equivalently, as a method for synthesizing context-derived labels
at coarser spatial resolution.  The auxiliary branch must predict this more
abstract, context-sensitive representation in order to successfully interface
with the decoder.

Experiments, covered in Section~\ref{sec:experiments}, focus on the PASCAL
semantic segmentation task.  We take baseline CNN architectures, the
established VGG~\cite{VGG} network and the state-of-the-art DenseNet~\cite{
DenseNet}, and report performance gains from enhancing them with our custom
regularizer during training.  Section~\ref{sec:experiments} also provides
ablation studies, explores an alternative regularizer implementation, and
visualizes representations learned by the label autoencoder.

Results demonstrate performance gains under all settings in which we applied
our regularization scheme: VGG or DenseNet, with or without data augmentation,
and with or without ImageNet pretraining.  Performance of a very deep DenseNet,
with data augmentation and ImageNet pretraining, is still further improved with
use of our regularizer during training.  Together, these results indicate that
we have discovered a new and generally applicable method for regularizing
supervised training of deep networks.  Moreover, our method has no cost at
test time; it produces networks architecturally identical to baseline designs.

Section~\ref{sec:conclusion} discusses implications of our demonstration that
it is possible to squeeze more benefit from detailed label maps when training
deep networks.  Our results open up a new area of inquiry on how best to build
datasets and design training procedures to efficiently utilize annotation.

\section{Related work}
\label{sec:related}

The abundance of data, but more limited availability of ground-truth
supervision, has sparked a flurry of recent interest in developing
unsupervised or weakly supervised methods for training deep neural networks.  Here, the idea is
to utilize a large reserve of unlabeled data in order to prime a deep network
to encode generally useful visual representations.  Subsequently, that network
can be fine-tuned on a novel target task, using actual ground-truth
supervision on a smaller dataset.  Pretraining on ImageNet~\cite{ImageNet}
currently yields such portable representations~\cite{DeCAF}, but lacks the
ability to scale without requiring additional human annotation effort.

Recent research explores a diverse array of data sources and tasks for
self-supervised learning.  In the domain of images, proposed tasks include
inpainting using context~\cite{CtxEncoder}, solving jigsaw puzzles~\cite{
Jigsaw}, colorization~\cite{ZIE:ECCV:2016,LMS:ECCV:2016,LMS:CVPR:2017},
cross-channel prediction~\cite{ZIE:CVPR:2017}, and learning a bidirectional
variant~\cite{DKD:ICLR:2017} of generative adversarial networks (GANs)~\cite{
GAN}.  In the video domain, recent works harness temporal coherence~\cite{
MCW:ICML:2009,JG:CVPR:2016}, co-occurrence~\cite{IZKA:ICLRworkshop:2016}, and
ordering~\cite{MZH:ECCV:2016}, as well as tracking~\cite{WG:ICCV:2015},
sequence modeling~\cite{SMS:ICML:2015}, and motion grouping~\cite{
PGDDH:CVPR:2017}.  Owens~\etal.~\cite{OWMFT:ECCV:2016} explore cross-modality
self-supervision, connecting vision and sound.  Agrawal~\etal.~\cite{
ANAML:NIPS:2016} and Nair~\etal.~\cite{NCAIAML:ICRA:2017} examine settings in
which a robot learns to predict the visual effects of its own actions.

Training a network to perform ImageNet classification or a self-supervised
task, in addition to the task of interest, can be viewed as a kind of implicit
regularization constraint.  Zhang~\etal.~\cite{ZLL:ICML:2016} explore explicit
auxiliary reconstruction tasks to regularize training.  However, they focus on
encoding and decoding image feature representations.  Our approach differs
entirely in the source of regularization.

Specifically, by autoencoding the structure of the target task labels, we
utilize a different reserve of information than all of the above methods.  We
design a new task, but whereas self-supervision formulates the new task on
external data, we derive the new task from the annotation.  This separation of
focus allows for possible synergistic combination of our method with
pretraining of either the self-supervised or supervised (ImageNet) variety.
Section~\ref{sec:experiments} tests the latter.

Another important distinction from recent self-supervised work is that, as
detailed in Section~\ref{sec:method}, we use a generic mechanism, based on
an autoencoder, for deriving our auxiliary task.  In contrast, the vast
majority of effort in self-supervision has relied on using domain-specific
knowledge to formulate appropriate tasks.  Inpainting~\cite{CtxEncoder},
jigsaw puzzles~\cite{Jigsaw}, and colorization~\cite{ZIE:ECCV:2016,
LMS:ECCV:2016,LMS:CVPR:2017} exemplify this mindset;
BiGANs~\cite{DKD:ICLR:2017} are perhaps an exception, but to date their results
compare less favorably~\cite{LMS:CVPR:2017}.

The work of Xie~\etal.~\cite{XHT:ECCV:2016} focuses on learning a
shallow corrective model that essentially denoises a predicted label map using
center-surround filtering.  In contrast, we build a deep model of label space.
Also, unlike~\cite{XHT:ECCV:2016}, our approach has no test-time cost, as we
impose it only as a regularizer during training, rather than as an ever-present
denoising layer.

Inspiration for our method traces back to the era of vision prior to the
pervasive use of deep learning.  It was once common to consider context as
important~\cite{Torralba:IJCV:2003}, reason about object parts, co-occurrence,
and interactions~\cite{DRF:IJCV:2011}, and design graphical models to capture
such relationships~\cite{STFW:ICCV:2005}.  We refer to only a few sample
papers as fully accounting for a decade of computer vision research is not
possible here.  In the following section, we open a pathway to pull such
thinking about compositional scene priors into the modern era: simply
learn, and employ, a deep model of label space.

\section{Method}
\label{sec:method}

\begin{figure}[t]
   \setlength\fboxsep{0pt}
   \begin{center}
      \begin{minipage}[t]{0.01\linewidth}
         \vspace{0pt}
      \end{minipage}
      \hfill
      \begin{minipage}[t]{0.45\linewidth}
         \vspace{-1pt}
         \begin{center}
            \includegraphics[width=1.0\linewidth]{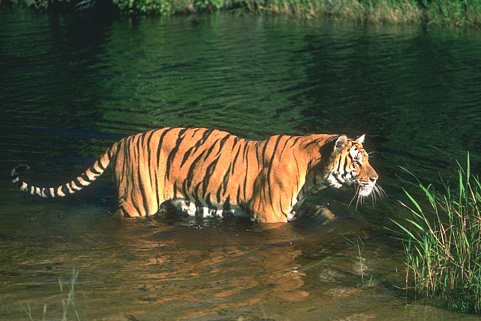}
         \end{center}
      \end{minipage}
      \hfill
      \begin{minipage}[t]{0.45\linewidth}
         \vspace{0pt}
         \begin{center}
            \begin{overpic}[width=1.0\linewidth,grid=false]{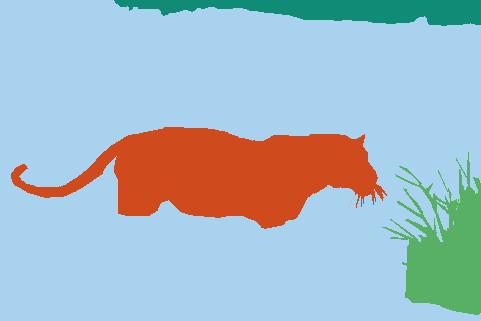}
               \put(40,30){\textcolor{white}{\footnotesize{\textsf{\textbf{cat}}}}}
               \put(80,10){\textcolor{white}{\footnotesize{\textsf{\textbf{grass}}}}}
               \put(8,44){\textcolor{black}{\scriptsize{\textsf{tail}}}}
               \put(12,42){\color{black}\vector(0,-4){12}}
               \put(80,52){\textcolor{black}{\scriptsize{\textsf{ear}}}}
               \put(82,50){\color{black}\vector(-1,-2){5}}
               \put(50,49){\textcolor{black}{\scriptsize{\textsf{head}}}}
               \put(58,47){\color{black}\vector(1,-1){15}}
               \put(27,10){\textcolor{black}{\scriptsize{\textsf{body}}}}
               \put(35,15){\color{black}\vector(1,1){10}}
            \end{overpic}
         \end{center}
      \end{minipage}
      \hfill
      \begin{minipage}[t]{0.01\linewidth}
         \vspace{0pt}
      \end{minipage}
   \end{center}
   \vspace{-0.01\linewidth}
   \caption{
      Informative structure in annotation.
      The shape of labeled semantic regions hints at unlabeled parts (black
      arrows).  Object co-occurrence provides a prior on scene composition.
   }
   \label{fig:tiger}
   \vspace{-0.03\linewidth}
\end{figure}

Figure~\ref{fig:tiger} explains the intuition behind the
regularization scheme outlined in Figure~\ref{fig:overview}.  Suppose we want
to train a CNN to recognize and segment cats, but our limited training set
consists only of tigers.  It is conceivable that the CNN will learn an
equivalence between black and orange striped texture and the cat category, as
such association suffices to classify every pixel on a tiger.  It thus
overfits to the tiger subclass and fails when tested on images of house cats.
This behavior could arise even if trained with detailed supervision of the
form shown in Figure~\ref{fig:tiger}.

Yet, the semantic segmentation ground-truth suggests to any human that texture
should not be the primary criterion.  There are no stripes in the annotation.
Over the entire training set, regions labeled as cat share a distinctive shape
that deforms in a manner suggestive of unlabeled parts (e.g. head, body, tail,
ear).  The presence or absence of other objects in the scene may also provide
contextual cues as to the chance of finding a cat.  How can we force the CNN to
notice this wealth of information during training?

We could consider treating the ground-truth label map as an image, and
clustering local patches.  The patch containing the skinny tail would fall in a
different cluster than that containing the pointy ear.  Adding the cluster
identities as another semantic label, and requiring the CNN to predict them,
would force the CNN to differentiate between the tail and ear by developing a
representation of shape.  This clustering approach is reminiscent of
Poselets~\cite{BM:ICCV:2009,BMBM:ECCV:2010}.

Following this strategy, we would need to hand-craft another scheme for
capturing object co-occurrence relations, perhaps by clustering descriptors
spanning a larger spatial extent.  We would prefer a general means of capturing
features of the ground-truth annotations, and one not limited to a few
hand-selected characteristics.  Fortunately, deep networks are a suitable
general tool for building the kind of abstract feature hierarchy we desire.

\begin{figure}[!th]
   \begin{center}
      \setlength\tabcolsep{6.5pt}
      \begin{tabular}{@{}c|c|c@{}}
      \footnotesize{Layers} & \footnotesize{DenseNet-67} & \footnotesize{DenseNet-121}\\
      \Hline
         & &\\[-0.01pt]
         Convolution
         &
            $\begin{bmatrix}
               3 \times 3\ \text{conv, stride 2}\\
               3 \times 3\ \text{conv}\\
               3 \times 3\ \text{conv}\\
            \end{bmatrix} \times 1$
         &
            $\begin{matrix}
               7 \times 7\ \text{conv,}\\
               \text{stride 2}\\
            \end{matrix}$
            \vspace{0.01\linewidth}
         \\[11pt]
      \hline
         Pooling &
            \multicolumn{2}{c}{$3 \times 3$ max pool, stride 2}\\
      \hline
         & &\\[-0.01pt]
         Dense Block (1)
         &
            $\begin{bmatrix}
               1 \times 1\ \text{conv}\\
               3 \times 3\ \text{conv}\\
            \end{bmatrix} \times 6$
         &
            $\begin{bmatrix}
               1 \times 1\ \text{conv}\\
               3 \times 3\ \text{conv}\\
            \end{bmatrix} \times 6$
            \vspace{0.01\linewidth}
         \\[5pt]
      \hline
         \multirow{2}{*}{Transition Layer (1)}
            & \multicolumn{2}{c}{$1 \times 1$ conv}\\
            \cline{2-3}
            & \multicolumn{2}{c}{$2 \times 2$ average pool, stride 2}\\
      \hline
         & &\\[-0.01pt]
         Dense Block (2)
         &
            $\begin{bmatrix}
               1 \times 1\ \text{conv} \\
               3 \times 3\ \text{conv}\\
            \end{bmatrix} \times 8$
         &
            $\begin{bmatrix}
               1 \times 1\ \text{conv} \\
               3 \times 3\ \text{conv}\\
            \end{bmatrix} \times 12$
            \vspace{0.01\linewidth}
         \\[5pt]
      \hline
         \\[-8pt]
         \multirow{2}{*}{Transition Layer (2)}
            & \multicolumn{2}{c}{$1 \times 1$ conv}\\
            \cline{2-3}
            & \multicolumn{2}{c}{$2 \times 2$ average pool, stride 2}\\
      \hline
         & &\\[-0.01pt]
         Dense Block (3)
         &
            $\begin{bmatrix}
               1 \times 1\ \text{conv} \\
               3 \times 3\ \text{conv}\\
            \end{bmatrix} \times 8$
         &
            $\begin{bmatrix}
               1 \times 1\ \text{conv} \\
               3 \times 3\ \text{conv}\\
            \end{bmatrix} \times 24$
            \vspace{0.01\linewidth}
         \\[5pt]
      \hline
         \multirow{2}{*}{Transition Layer (3)}
            & \multicolumn{2}{c}{$1 \times 1$ conv}\\
            \cline{2-3}
            & \multicolumn{2}{c}{$2 \times 2$ average pool, stride 2}\\
      \hline
         & &\\[-0.01pt]
         Dense Block (4)
         &
            $\begin{bmatrix}
               1 \times 1\ \text{conv}\\
               3 \times 3\ \text{conv}\\
            \end{bmatrix} \times 8$
         &
            $\begin{bmatrix}
               1 \times 1\ \text{conv}\\
               3 \times 3\ \text{conv}\\
            \end{bmatrix} \times 16$
            \vspace{0.01\linewidth}
         \\[5pt]
      \hline
      \end{tabular}
      \caption{
        DenseNet architectural specifications.
      }
      \label{fig:densenet}
   \end{center}
\end{figure}

\begin{figure}
   \begin{center}
      \includegraphics[width=1\linewidth]{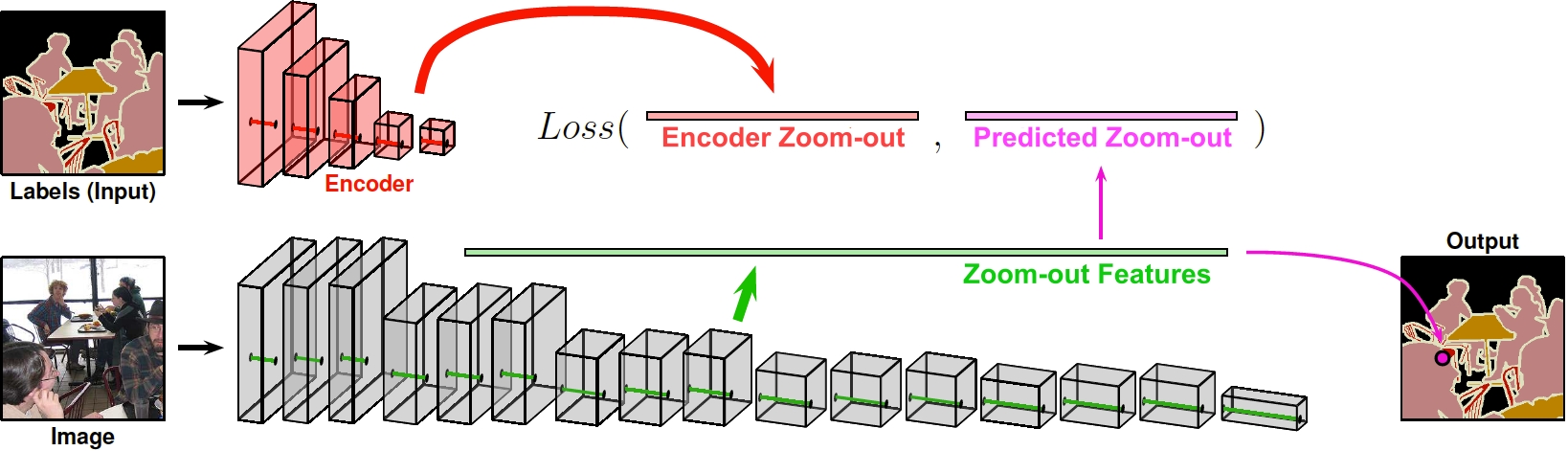}
   \end{center}
   \vspace{-0.01\linewidth}
   \caption{
      Alternative regularization scheme.
      Instead of predicting a representation to pass through the decoder, as
      in Figure~\ref{fig:overview}, we can train with an auxiliary regression
      problem.  We place a loss on directly predicting activations produced by
      the hidden layers of the encoder.
   }
   \label{fig:alternative}
   \vspace{-0.015\linewidth}
\end{figure}

\subsection{Modeling labels}
\label{sec:modeling}

As shown in Figure~\ref{fig:overview}, we train an autoencoder
on the ground-truth label maps.  This autoencoder consumes a semantic
segmentation label map as input and attempts to reconstruct it as output.  By being required to pass through a small bottleneck representation,
the job of the autoencoder is nontrivial.  It must compress the label map into
the bottleneck representation.  This compression constraint will (ideally)
force the autoencoder to discover and implicitly encode parts and contextual
relationships.

Ground-truth semantic segmentation label maps are simpler than real images, so
this autoencoder need not have as high capacity as a network operating on
natural images.  We use a relatively simply autoencoder architecture,
consisting of a mirrored encoder and decoder, with no skip connections.  The
encoder is a sequence of five $3 \times 3$ convolutional layers, with
$2 \times 2$ max-pooling between them.  The decoder uses $2\mathrm{x}$
upsampling followed by $3 \times 3$ convolution.  As a default, we set each
layer to have $32$ channels.  We also experiment with some higher-capacity
variants:
\begin{itemize}
   \setlength{\itemsep}{-2pt}
   \item{conv1: $32$-channels;~~conv2-5: $128$ channels each}
   \item{conv1: $32$;~~conv2-4: $128$;~~conv5: $256$ channels}
\end{itemize}
These channel progressions are for the encoder; the decoder uses the same in
reverse order.  We refer to these three autoencoder variants by the number
of channels in their respective bottleneck layers ($32$, $128$, or $256$).

\subsection{Baseline CNN architectures}
\label{sec:architectures}

Convolutional neural networks for image classification gradually reduce
spatial resolution with depth through a series of pooling layers~\cite{
AlexNet,VGG,ResNet,DenseNet}.  As the semantic segmentation task requires
output at fine spatial resolution, some method of preserving or recovering
spatial resolution must be introduced into the architecture.  One option is
to gradually re-expand spatial resolution via upsampling~\cite{UNet,SegNet}.
Other approaches utilize some form of skip-connection to forward spatially
resolved features from lower layers of the network to the final layer~\cite{
FCN,Hypercolumns,ZoomOut}.  Dilated~\cite{YK:ICLR:2016} or atrous
convolutions~\cite{DeepLab} can also be mixed in.  Alternatively, the
basic CNN architecture can be reformulated in a multigrid setting~\cite{
KMY:CVPR:2017}.

Our goal is to examine the effects of a regularization scheme in isolation
from major architectural design changes. Hence, we choose zoom-out CNN architectures as a primary basis for experimentation,
as it is minimally separated from the established classification
networks in design space. It also offers the added advantage of having
readily available ImageNet pretrained models, easing experimentation in this
setting.

We consider zoom-out variants of VGG-16~\cite{VGG} and DenseNet~\cite{
DenseNet}. Refer to Section~\ref{sec:zoomfurther} for architecture details.
VGG-16 is widely used, while DenseNet~\cite{DenseNet} represents the latest
high-performance evolution of ResNet~\cite{ResNet}-like designs.  We use
67-layer and 121-layer DenseNets with the architectural details specified in
Figure~\ref{fig:densenet}. The 67-layer net uses a channel growth rate of
$48$, while the 121-layer network, the same as in~\cite{DenseNet}, uses a
growth rate of $32$.  We work with $256 \times 256$ input and output spatial
resolutions in both CNNs and our label autoencoder.

\subsection{Regularization via label modeling}
\label{sec:regularization}

As shown by the large gray arrow in Figure~\ref{fig:overview}, we impose our
regularizer by connecting a CNN (e.g., VGG or DenseNet) to the decoder portion of
our learned label autoencoder.  Importantly, the \emph{decoder parameters
are frozen during this training phase}.  The CNN now has two tasks, each with
an associated loss, to perform during training.  As usual, it must predict
semantic segmentation using zoom-out features. It must also predict the same
semantic segmentation via an auxiliary path through the decoder.
Backpropagation from losses along both paths influences CNN parameter updates.
Though they participate in one of these paths, parameters internal to the
decoder are never updated.

We connect VGG-16 or DenseNet to the decoder by predicting input for the
decoder from the output of the penultimate CNN layer prior to global pooling.
This is the second-to-last convolutional layer, and is selected because its
spatial resolution matches that of the expected decoder input.  The prediction
itself is made via a new $1 \times 1$ convolutional layer, dedicated for that
purpose.

If the label autoencoder learns useful abstractions, requiring the CNN
to work through the decoder ensures that it learns to work with those
abstractions.  The zoom-out pathway allows the CNN to make direct
predictions, while the decoder pathway ensures that the CNN has
``good reasons'' or a high-level abstract justification for its predictions.

Assuming autoencoder layers gradually buildup good abstractions, there
exist alternative methods of connecting it as a regularizer.
Figure~\ref{fig:alternative} diagrams one such alternative.  Here, we ask the
CNN to directly predict the feature representation built by the label encoder.
Encoder parameters are, of course, frozen here.  An auxiliary layer attempts to
predict the encoder zoom-out features from the CNN zoom-out features at the corresponding
spatial location.  The CNN must also still solve the original semantic
segmentation task.

As Section~\ref{sec:experiments} shows, this alternative scheme works well,
but not quite as well as using the decoder pathway.  Using the decoder is
also appealing for more reasons than performance alone.  Defining an auxiliary
loss in terms of decoder semantic segmentation output is more interpretable
than defining it in terms of mean square error (MSE) between two zoom-out
features.  Moreover, the decoder output is visually interpretable; we can see
the semantic segmentation predicted by the CNN via the decoder.

\section{Experiments}
\label{sec:experiments}

The PASCAL dataset~\cite{PASCAL} serves as our experimental testbed.  We
follow standard procedure for semantic segmentation, using the official PASCAL
2012 training set, and reporting performance in terms of mean intersection over
union (mIoU) on the validation set (as validation ground-truth is publicly
available).  We explore both our decoder- and encoder-based regularization
schemes in combination with multiple choices of base network, data
augmentation, and pretraining.  When applying the encoder as a regularizer,
we task the CNN with predicting the concatenation of the encoder's activations
in its conv1 and conv3 layers.

\subsection{Setup}
\label{sec:setup}

All experiments are done in PyTorch~\cite{PyTorch}, using the Adam~\cite{Adam} update
rule when training networks.  Models trained from scratch use a batch size of
$12$ and learning rate of $10^{-4}$ which after $80$ epochs decreased to
$10^{-5}$ for an additional $20$ epochs.  For the case of ImageNet
pretrained models, we normalize zoom-out features such that they have
zero mean and unit variance.  We keep the deep network weights frozen and
train the classifier for $10$ epochs with learning rate of $10^{-4}$.
Then we decrease the learning rate to $10^{-5}$ and train end-to-end for
additional $40$ epochs.

Data augmentation, when used, includes: a random horizontal flip and a crop of random size in the
(0.08 to 1.0) of the original size and a random aspect ratio of 3/4 to 4/3 of
the original aspect ratio, which is finally resized to create a $256 \times 256$
image Pretrained models are based
on the PyTorch torchvision library~\cite{torchvision}.

We use cross-entropy loss on auxiliary regularization branches, except where
indicated by a superscript $\dagger$ in results tables.  For these experiments,
we use MSE loss.

\begin{table}[!th]
   \definecolor{gray}{rgb}{0.33,0.33,0.33}
   \begin{center}
      \setlength\tabcolsep{3.5pt}
      \begin{tabular}{@{}l|c|c|>{\columncolor{white}[\tabcolsep][0pt]}r@{}}
      Architecture                        & Data-Aug?                & Auxiliary Regularizer                               & mIoU\\%
      \Hline%
                                          & \cellcolor{gray!15}{no}  & \cellcolor{gray!15}\textcolor{gray}{\textbf{none}}  & \cellcolor{gray!15}\textcolor{gray}{\textbf{37.3}}\\
      \multirow{2}{*}{VGG-16}             & no                       & Encoder (conv1 \& conv3)$^\dagger$                  & 41.1\\
      \multirow{2}{*}{%
         \footnotesize{~~-zoom-out}}   & no                       & \textbf{Decoder (32 channel)}$^\dagger$             & \textbf{42.4}\\
                                          \hhline{~---}
                                          & \cellcolor{gray!15}{yes} & \cellcolor{gray!15}\textcolor{gray}{\textbf{none}}  & \cellcolor{gray!15}\textcolor{gray}{\textbf{55.2}}\\
                                          & yes                      & \textbf{Decoder (128 channel)}                      & \textbf{57.1}\\
      \Hline
      \multirow{2}{*}{%
         VGG-16\footnotesize{-FCN8s}}     & \cellcolor{gray!15}{yes} & \cellcolor{gray!15}\textcolor{gray}{\textbf{none}}  & \cellcolor{gray!15}\textcolor{gray}{\textbf{51.5}}\\
                                          & yes                      & \textbf{Decoder (128 channel)}                      & \textbf{54.1}\\
      \Hline
                                          & \cellcolor{gray!15}{no}  & \cellcolor{gray!15}\textcolor{gray}{\textbf{none}}  & \cellcolor{gray!15}\textcolor{gray}{\textbf{40.5}}\\
                                          & no                       & Encoder (conv1 \& conv3)$^\dagger$                  & 44.0\\
                                          & no                       & \textbf{Decoder (32 channel)}$^\dagger$             & \textbf{45.2}\\
      \multirow{1}{*}{DenseNet-67}        & no                       & Decoder (128 channel)                               & 42.5\\
                                          \hhline{~---}
      \multirow{1}{*}{%
         \footnotesize{~~-zoom-out}}   & \cellcolor{gray!15}{yes} & \cellcolor{gray!15}\textcolor{gray}{\textbf{none}}  & \cellcolor{gray!15}\textcolor{gray}{\textbf{58.8}}\\
                                          & yes                      & Decoder (32 channel)                                & 59.4\\
                                          & yes                      & \textbf{Decoder (128 channel)}                      & \textbf{60.6}\\
                                          & yes                      & Decoder (256 channel)                               & 59.8\\
      \Hline
      \end{tabular}
      \caption{
         PASCAL mIoU without ImageNet pretraining.
         In each experimental setting (choice of architecture, and presence or
         absence of data augmentation), training with any of our regularizers
         improves performance over the baseline (shown in gray).
      }
      \label{tab:pascal_scratch}
   \end{center}
\end{table}

\begin{table}[!th]
   \begin{center}
      \setlength\tabcolsep{7.5pt}
      \begin{tabular}{@{}l|c|c|>{\columncolor{white}[\tabcolsep][0pt]}r@{}}
      Architecture                        & Data-Aug                 & Auxiliary Regularizer                               & mIoU\\
      \Hline
      &\cellcolor{gray!15}{}
      &\cellcolor{gray!15}{}
      &\cellcolor{gray!15}{}\\[-9pt]
                                          & \cellcolor{gray!15}{yes} & \cellcolor{gray!15}\textcolor{gray}{\textbf{none}}  & \cellcolor{gray!15}\textcolor{gray}{\textbf{58.8}}\\
      \multirow{1}{*}{DenseNet-67}        & yes                      & \textbf{Decoder (128 channel)}                      & \textbf{60.6}\\
      \multirow{1}{*}{%
         \footnotesize{~~-zoom-out}}   & yes                      & Unfrozen Decoder                                    & 60.2\\
                                          & yes                      & Random Init.~Decoder                                & 58.8\\
      \Hline
      \end{tabular}
      \caption{
         Ablation study.
         PASCAL mIoU deteriorates if the decoder parameters are not held fixed
         while training the main CNN.
      }
      \label{tab:pascal_ablation}
   \end{center}
\end{table}

\begin{table}[!th]
   \begin{center}
      \setlength\tabcolsep{6pt}
      \begin{tabular}{@{}l|c|c|>{\columncolor{white}[\tabcolsep][0pt]}r@{}}
      Architecture                        & Data-Aug?                & Auxiliary Regularizer                               & mIoU\\
      \Hline
      &\cellcolor{gray!15}{}
      &\cellcolor{gray!15}{}
      &\cellcolor{gray!15}{}\\[-9pt]
      \multirow{1}{*}{VGG-16}             & \cellcolor{gray!15}{no}  & \cellcolor{gray!15}\textcolor{gray}{\textbf{none}}  & \cellcolor{gray!15}\textcolor{gray}{\textbf{67.1}}\\
      \multirow{1}{*}{%
         \footnotesize{~~-zoom-out}}   & no                       & \textbf{Decoder (32 channel)}                       & \textbf{68.8}\\
      \Hline
      &\cellcolor{gray!15}{}
      &\cellcolor{gray!15}{}
      &\cellcolor{gray!15}{}\\[-9pt]
      \multirow{1}{*}{DenseNet-121}       & \cellcolor{gray!15}{yes} & \cellcolor{gray!15}\textcolor{gray}{\textbf{none}}  & \cellcolor{gray!15}\textcolor{gray}{\textbf{71.6}}\\
      \multirow{1}{*}{%
         \footnotesize{~~-zoom-out}}   & yes                      & \textbf{Decoder (128 channel)}                      & \textbf{71.9}\\
      \Hline
      &\cellcolor{gray!15}{}
      &\cellcolor{gray!15}{}
      &\cellcolor{gray!15}{}\\[-9pt]
      \multirow{1}{*}{ResNet-101}         & \cellcolor{gray!15}{yes} & \cellcolor{gray!15}\textcolor{gray}{\textbf{none}}  & \cellcolor{gray!15}\textcolor{gray}{\textbf{75.4}}\\
      \multirow{1}{*}{%
         \footnotesize{~~-PSPNet}}        & yes                      & \textbf{Decoder (128 channel)}                      & \textbf{75.9}\\
      \Hline
      \end{tabular}
      \caption{
         PASCAL mIoU with ImageNet pretraining.
      }
      \label{tab:pascal_imagenet}
   \end{center}
\end{table}

\begin{table}[!th]
   \begin{center}
      \setlength\tabcolsep{7.5pt}
      \begin{tabular}{@{}l|c|c|>{\columncolor{white}[\tabcolsep][0pt]}r@{}}
      Architecture                        & Data-Aug                 & Auxiliary Regularizer                               & mIoU\\
      \Hline
      &\cellcolor{gray!15}{}
      &\cellcolor{gray!15}{}
      &\cellcolor{gray!15}{}\\[-9pt]
      \multirow{1}{*}{DenseNet-67}        & \cellcolor{gray!15}{yes} & \cellcolor{gray!15}\textcolor{gray}{\textbf{none}}  & \cellcolor{gray!15}\textcolor{gray}{\textbf{72.3}}\\
      \multirow{1}{*}{%
         \footnotesize{~~-zoom-out}}   & yes                      & \textbf{Decoder (128 channel)}                      & \textbf{73.6}\\

      \Hline
      \end{tabular}
      \caption{
         PASCAL mIoU with COCO pretraining.
      }
      \label{tab:pascal_coco}
   \end{center}
\end{table}

\begin{figure}
   \begin{center}
      \includegraphics[width=0.75\linewidth]{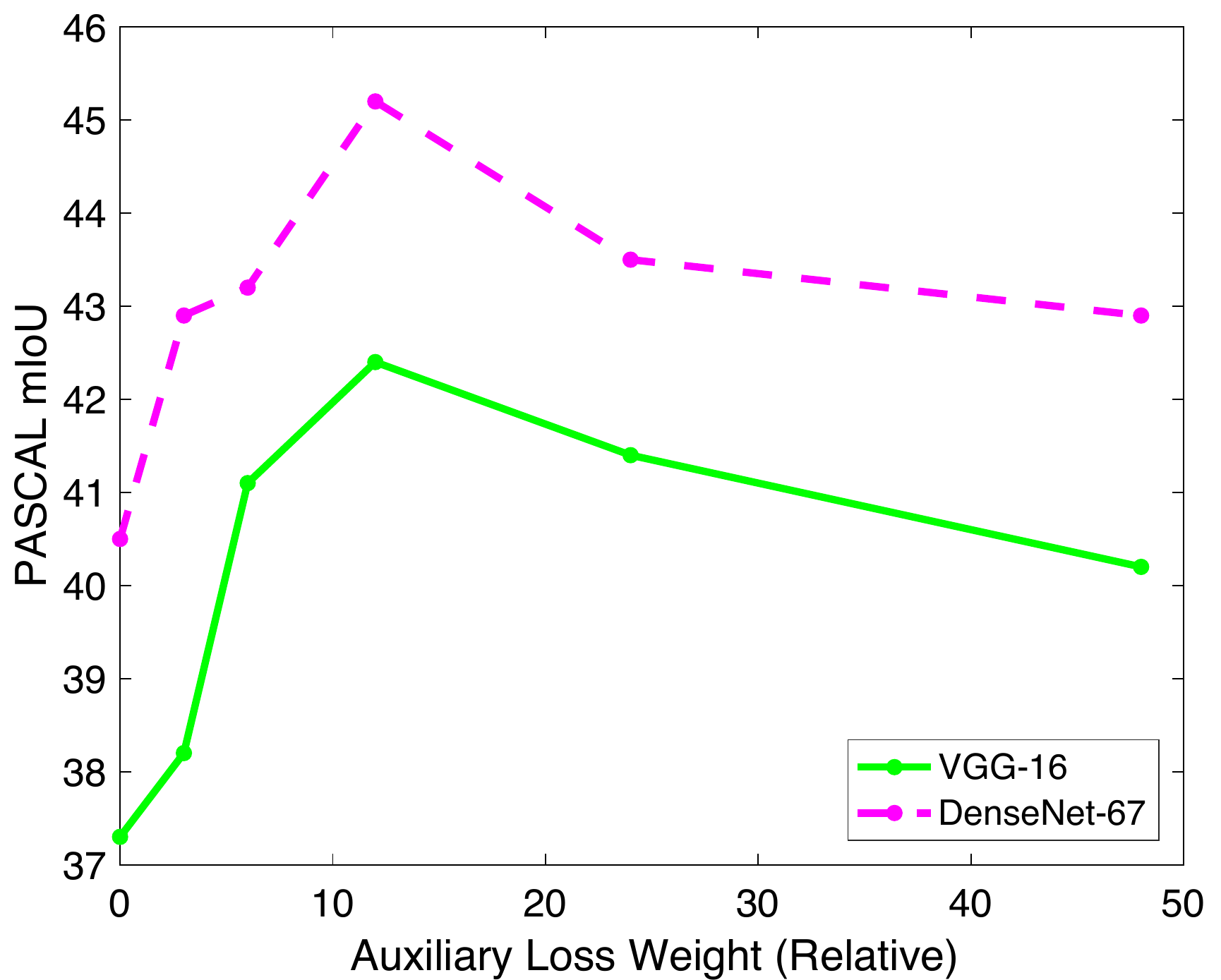}
   \end{center}
   \vspace{-0.03\linewidth}
   \caption{
      Auxiliary loss weighting.
      We plot validation performance as a function of the relative weight of the
      losses on the auxiliary vs primary output branches when training with the
      setup in Figure~\ref{fig:overview}. Performance is mIoU on PASCAL, without
      ImageNet pretraining or data augmentation.  Note that any nonzero
      weight on the auxiliary loss (any regularization) improves over
      the baseline.
   }
   \label{fig:weighting}
   \vspace{-0.035\linewidth}
\end{figure}

\begin{figure*}
   \setlength\fboxsep{0pt}
   \begin{center}
   \begin{minipage}[t]{0.98\linewidth}
   \begin{center}
      \begin{minipage}[t]{0.19\linewidth}
         \vspace{0pt}
         \begin{center}
         \begin{minipage}[t]{0.98\linewidth}
            \vspace{0pt}
            \begin{center}
               \fbox{\includegraphics[width=1.00\linewidth]{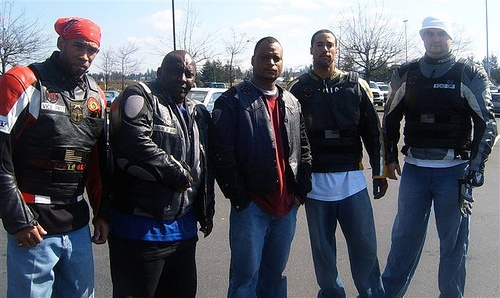}}\\
               \vspace{0.01\linewidth}
               \fbox{\includegraphics[width=1.00\linewidth]{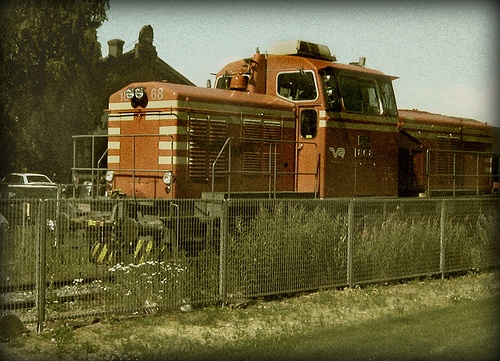}}\\
               \vspace{0.01\linewidth}
               \fbox{\includegraphics[width=1.00\linewidth]{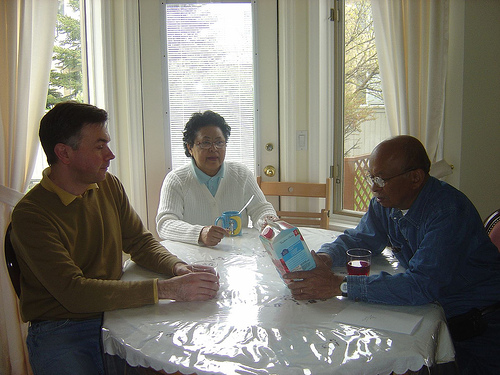}}\\
               \vspace{0.01\linewidth}
               \fbox{\includegraphics[width=1.00\linewidth]{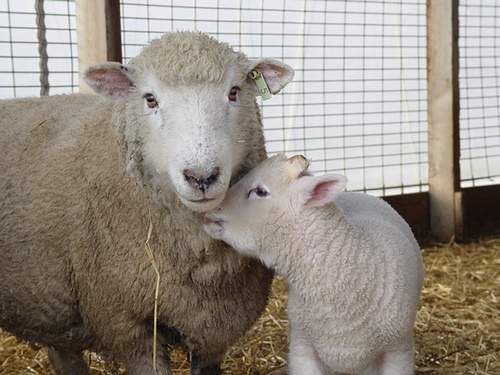}}\\
               \vspace{0.01\linewidth}
               \fbox{\includegraphics[width=1.00\linewidth]{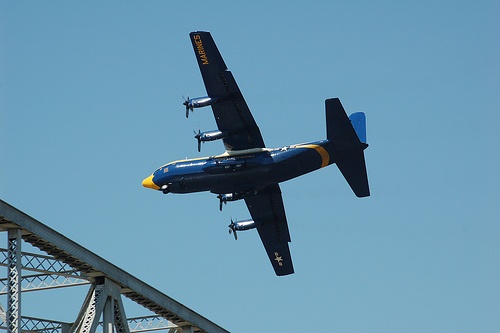}}\\
               \vspace{0.01\linewidth}
               \fbox{\includegraphics[width=1.00\linewidth]{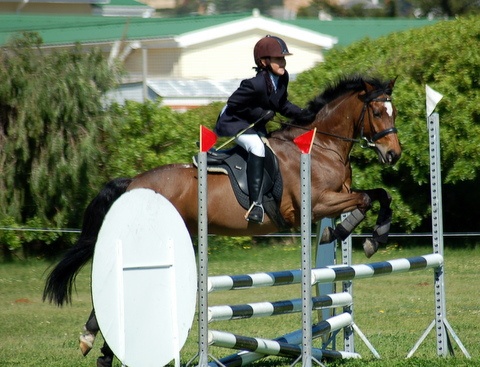}}\\
               \vspace{0.01\linewidth}
               \fbox{\includegraphics[width=1.00\linewidth]{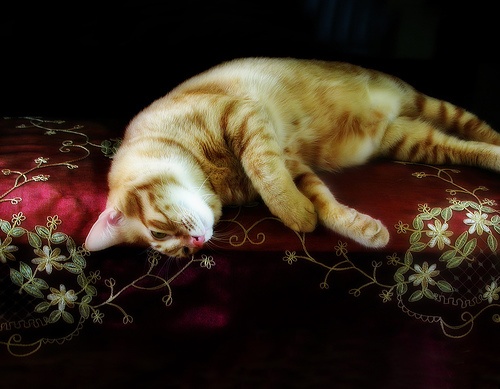}}\\
               \vspace{0.01\linewidth}
               \fbox{\includegraphics[width=1.00\linewidth]{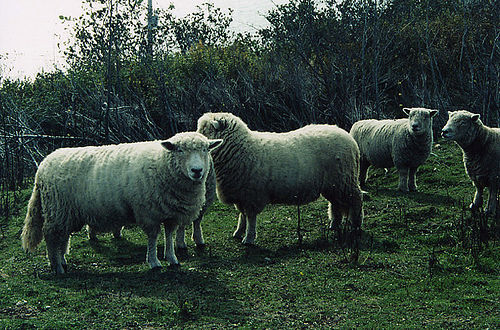}}\\
               \vspace{0.01\linewidth}
               \scriptsize{\textbf{\textsf{Image}}}
            \end{center}
         \end{minipage}
         \end{center}
      \end{minipage}
      \hfill
      \begin{minipage}[t]{0.38\linewidth}
         \vspace{0pt}
         \begin{center}
         \begin{minipage}[t]{0.49\linewidth}
            \vspace{0pt}
            \begin{center}
               \fbox{\includegraphics[width=1.00\linewidth]{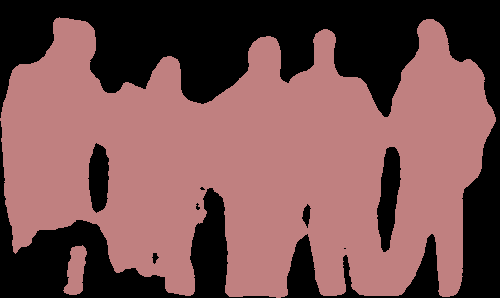}}\\
               \vspace{0.01\linewidth}
               \fbox{\includegraphics[width=1.00\linewidth]{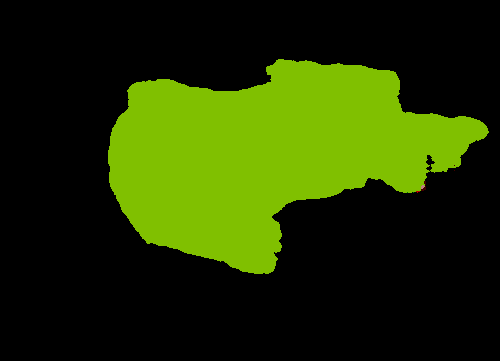}}\\
               \vspace{0.01\linewidth}
               \fbox{\includegraphics[width=1.00\linewidth]{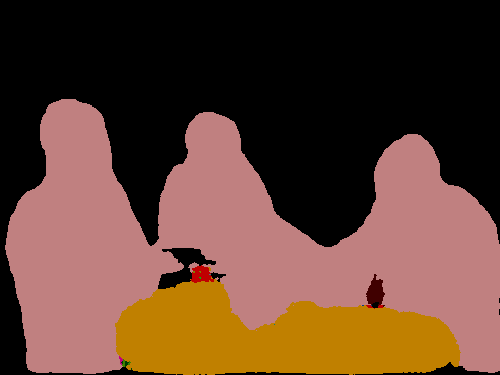}}\\
               \vspace{0.01\linewidth}
               \fbox{\includegraphics[width=1.00\linewidth]{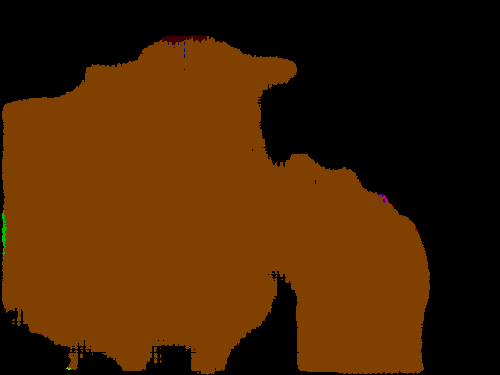}}\\
               \vspace{0.01\linewidth}
               \fbox{\includegraphics[width=1.00\linewidth]{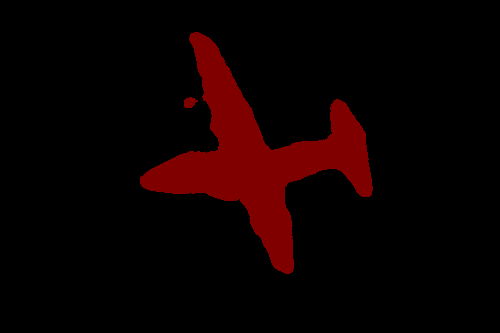}}\\
               \vspace{0.01\linewidth}
               \fbox{\includegraphics[width=1.00\linewidth]{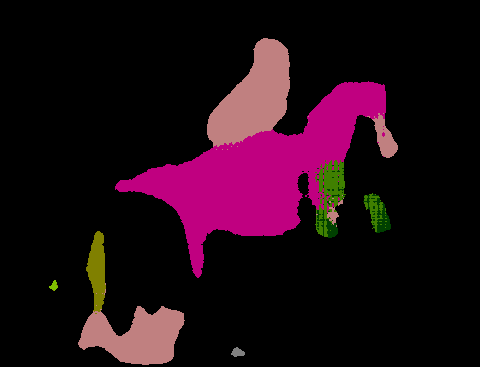}}\\
               \vspace{0.01\linewidth}
               \fbox{\includegraphics[width=1.00\linewidth]{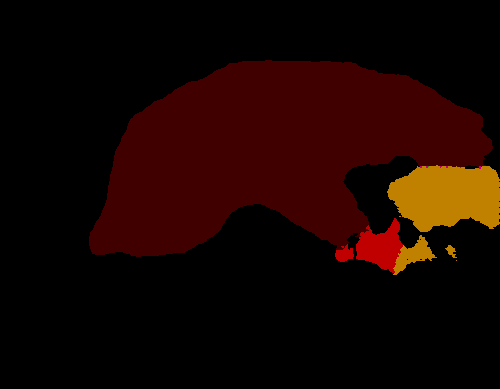}}\\
               \vspace{0.01\linewidth}
               \fbox{\includegraphics[width=1.00\linewidth]{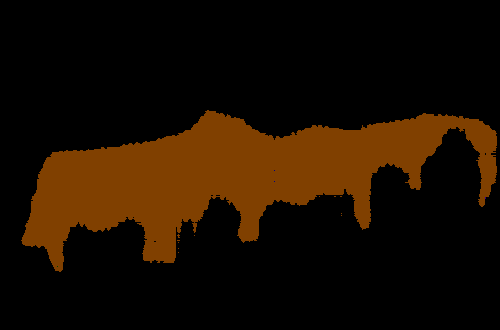}}\\
               \vspace{0.01\linewidth}
               \scriptsize{\textbf{\textsf{Auxiliary Output}}}
            \end{center}
         \end{minipage}
         \begin{minipage}[t]{0.49\linewidth}
            \vspace{0pt}
            \begin{center}
               \fcolorbox{red}{white}{\includegraphics[width=1.00\linewidth]{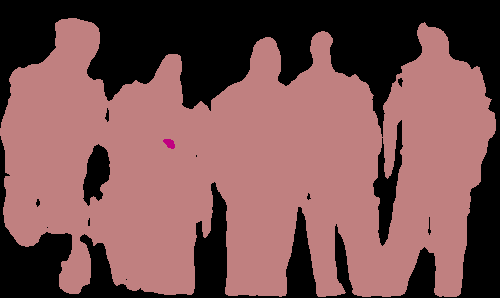}}\\
               \vspace{0.01\linewidth}
               \fcolorbox{red}{white}{\includegraphics[width=1.00\linewidth]{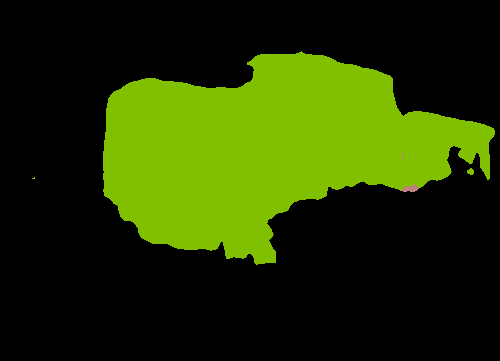}}\\
               \vspace{0.01\linewidth}
               \fcolorbox{red}{white}{\includegraphics[width=1.00\linewidth]{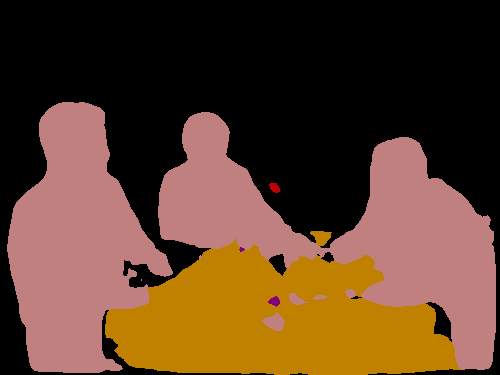}}\\
               \vspace{0.01\linewidth}
               \fcolorbox{red}{white}{\includegraphics[width=1.00\linewidth]{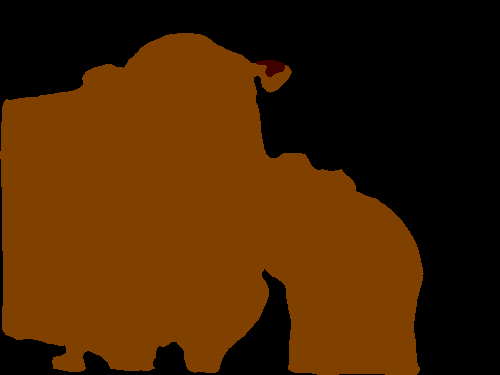}}\\
               \vspace{0.01\linewidth}
               \fcolorbox{red}{white}{\includegraphics[width=1.00\linewidth]{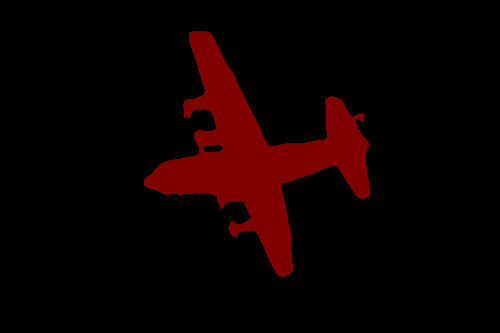}}\\
               \vspace{0.01\linewidth}
               \fcolorbox{red}{white}{\includegraphics[width=1.00\linewidth]{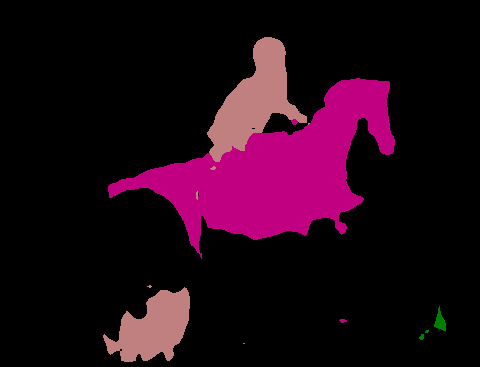}}\\
               \vspace{0.01\linewidth}
               \fcolorbox{red}{white}{\includegraphics[width=1.00\linewidth]{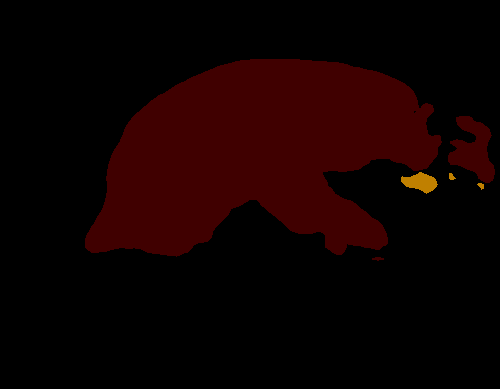}}\\
               \vspace{0.01\linewidth}
               \fcolorbox{red}{white}{\includegraphics[width=1.00\linewidth]{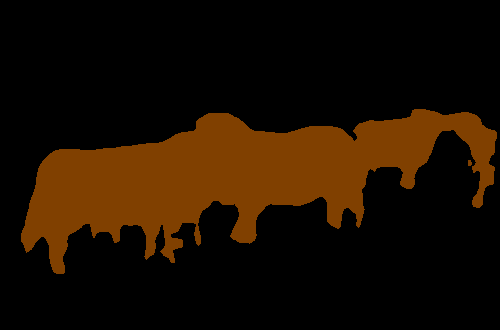}}\\
               \vspace{0.01\linewidth}
               \scriptsize{\textbf{\textsf{\textcolor{red}{Primary Output}}}}
            \end{center}
         \end{minipage}\\
         \vspace{1pt}
         \rule{0.975\linewidth}{0.5pt}\\
         \scriptsize{\textbf{\textsf{Our System: DenseNet-67 trained with regularizer}}}
         \end{center}
      \end{minipage}
      \hfill
      \begin{minipage}[t]{0.19\linewidth}
         \vspace{0pt}
         \begin{center}
         \begin{minipage}[t]{0.98\linewidth}
            \vspace{0pt}
            \begin{center}
               \fbox{\includegraphics[width=1.00\linewidth]{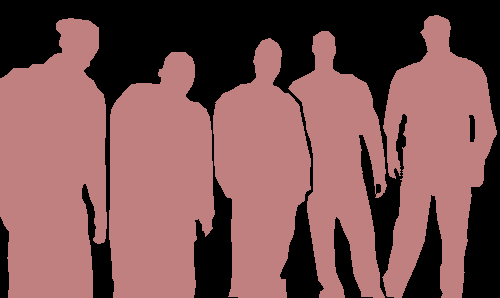}}\\
               \vspace{0.01\linewidth}
               \fbox{\includegraphics[width=1.00\linewidth]{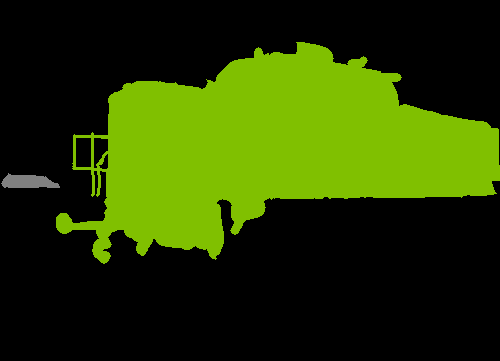}}\\
               \vspace{0.01\linewidth}
               \fbox{\includegraphics[width=1.00\linewidth]{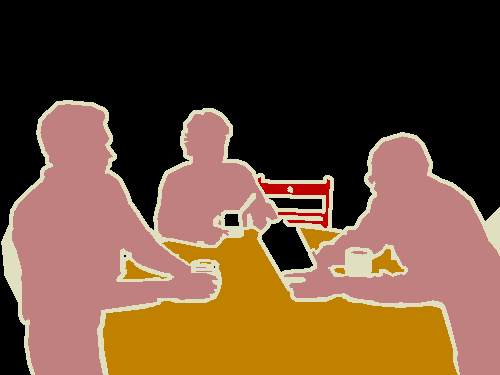}}\\
               \vspace{0.01\linewidth}
               \fbox{\includegraphics[width=1.00\linewidth]{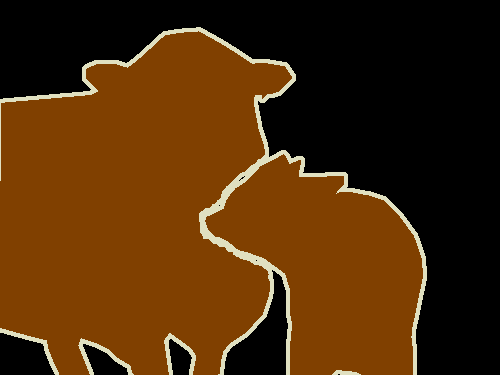}}\\
               \vspace{0.01\linewidth}
               \fbox{\includegraphics[width=1.00\linewidth]{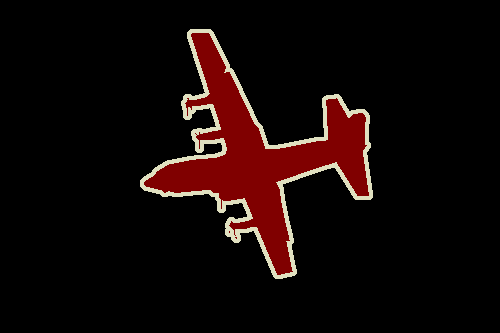}}\\
               \vspace{0.01\linewidth}
               \fbox{\includegraphics[width=1.00\linewidth]{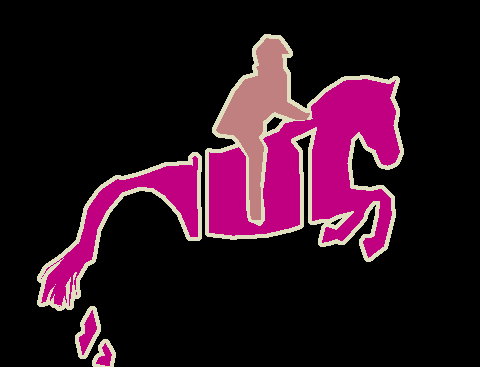}}\\
               \vspace{0.01\linewidth}
               \fbox{\includegraphics[width=1.00\linewidth]{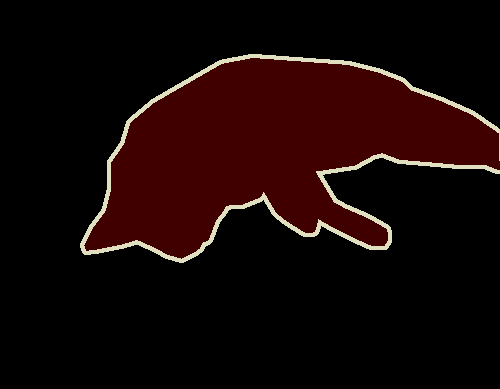}}\\
               \vspace{0.01\linewidth}
               \fbox{\includegraphics[width=1.00\linewidth]{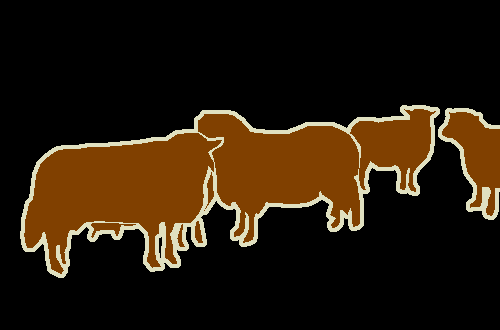}}\\
               \vspace{0.01\linewidth}
               \scriptsize{\textbf{\textsf{Ground-truth}}}
            \end{center}
         \end{minipage}
         \end{center}
      \end{minipage}
      \hfill
      \begin{minipage}[t]{0.19\linewidth}
         \vspace{0pt}
         \begin{center}
         \begin{minipage}[t]{0.98\linewidth}
            \vspace{0pt}
            \begin{center}
               \fbox{\includegraphics[width=1.00\linewidth]{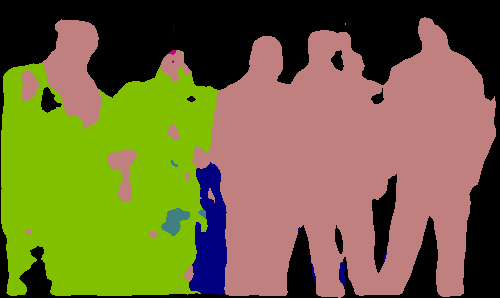}}\\
               \vspace{0.01\linewidth}
               \fbox{\includegraphics[width=1.00\linewidth]{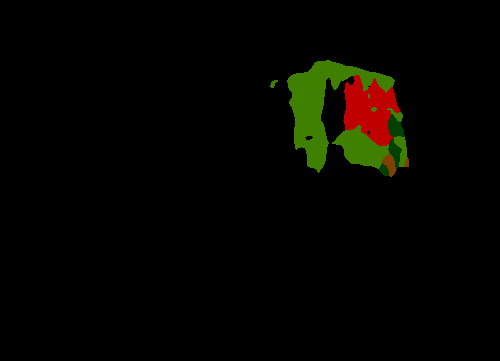}}\\
               \vspace{0.01\linewidth}
               \fbox{\includegraphics[width=1.00\linewidth]{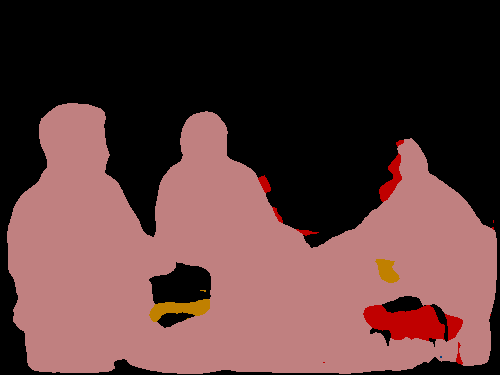}}\\
               \vspace{0.01\linewidth}
               \fbox{\includegraphics[width=1.00\linewidth]{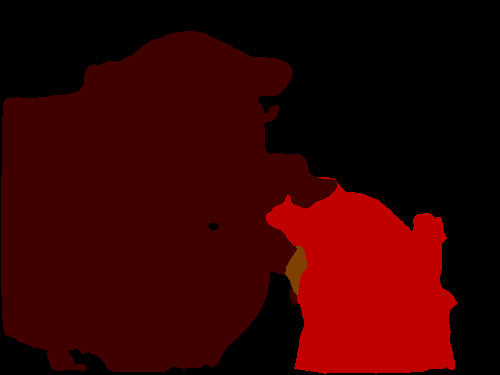}}\\
               \vspace{0.01\linewidth}
               \fbox{\includegraphics[width=1.00\linewidth]{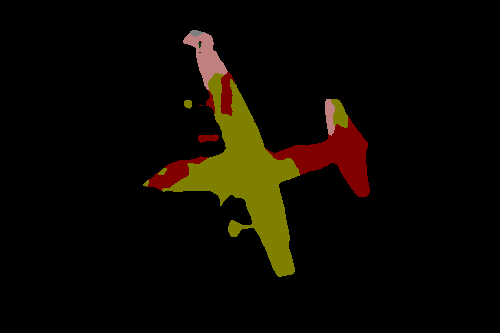}}\\
               \vspace{0.01\linewidth}
               \fbox{\includegraphics[width=1.00\linewidth]{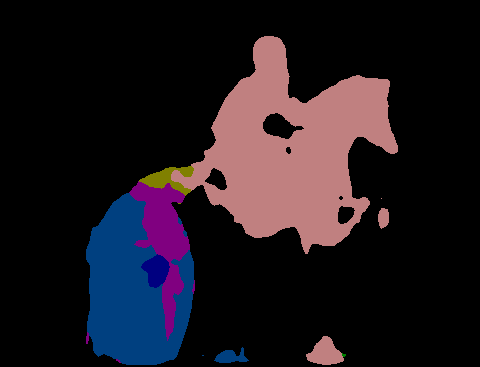}}\\
               \vspace{0.01\linewidth}
               \fbox{\includegraphics[width=1.00\linewidth]{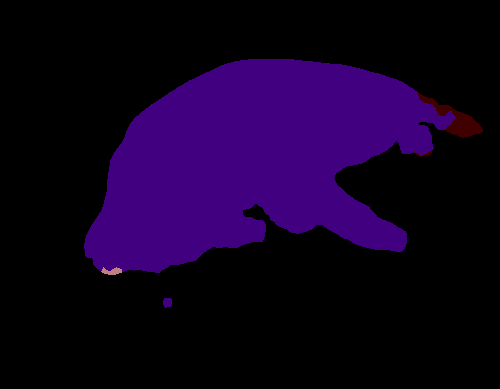}}\\
               \vspace{0.01\linewidth}
               \fbox{\includegraphics[width=1.00\linewidth]{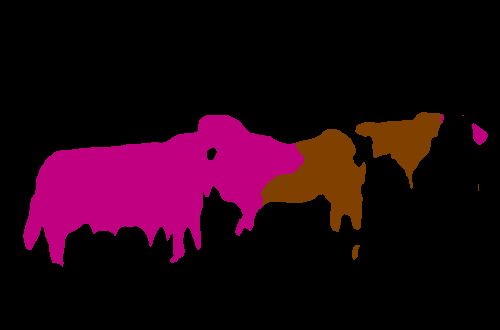}}\\
               \vspace{0.01\linewidth}
               \scriptsize{\textbf{\textsf{Baseline DenseNet-67}}}
            \end{center}
         \end{minipage}
         \end{center}
      \end{minipage}
   \end{center}
   \end{minipage}
   \end{center}
   \vspace{-0.03\linewidth}
   \begin{small}

   \caption{
      Semantic segmentation results on PASCAL.
      We show the output of a baseline 67-layer zoom-out DenseNet (rightmost
      column) compared to that of the same architecture trained with our
      auxiliary decoder branch as a regularizer (middle columns).  All examples
      are from the validation set.  While we can discard the auxiliary branch
      after training, we include its output here to display the decoder's
      operation.  Our network provides high-level signals to the decoder which,
      in turn, produces reasonable segmentations.  To best illustrate the
      effect of regularization, all results shown are for networks trained from
      scratch, without ImageNet pretraining or data augmentation.  This
      corresponds to the $40.5$ to $45.2$ jump in mIoU reported in
      Table~\ref{tab:pascal_scratch}, between the baseline and our primary
      output.
   }
      \end{small}
\label{fig:yoyo}
\end{figure*}

\subsection{Semantic Segmentation Results}
\label{sec:results}

Tables~\ref{tab:pascal_scratch},~\ref{tab:pascal_imagenet},
and~\ref{tab:pascal_coco} summarize the performance benefits of training with
our regularizer.  In the absence of pretraining or data augmentation, we boost
performance of both VGG-16 and DenseNet-67 by $5.1$ and $4.7$ mIoU,
respectively, which is more than a $10\%$ relative boost.  Regularization with
our decoder still improves mIoU (from $58.8$ to $60.6$) of DenseNet-67 trained
with data augmentation.  To further show the robustness of our regularization
scheme to the choice of architecture, we also experiment with an FCN~\cite{FCN}
version of VGG-16, as included in Table~\ref{tab:pascal_scratch}.

Table~\ref{tab:pascal_ablation} demonstrates the necessity of our two-phase
training procedure.  If we unfreeze the decoder and update its parameters in
the second training phase, test performance of the primary output deteriorates.
Likewise, if we skip the first phase, and train from scratch with an unfrozen,
randomly initialized decoder, the accuracy gain disappears.  Thus, the
regularization effect is due to a transfer of information from the learned
label model, rather than stemming from an architectural design of dual output
pathways.

Table~\ref{tab:pascal_imagenet} shows that our regularization scheme
synergizes with ImageNet pretraining.  It improves VGG-16 performance, and
even provides some benefit to a very deep 121-layer DenseNet pretrained on
ImageNet, while using data augmentation.  A baseline $71.6$ mIoU for DenseNet-121
is near state-of-the-art for networks that do not employ additional tricks
(\eg custom pooling layers~\cite{PSPNet}, use of multiscale, or post-processing
with CRFs~\cite{DeepLab}).  Our improvement to $71.9$ mIoU may be nontrivial.
Expanding trials in combination with pretraining, our regularizer improves
results when pretraining on COCO, as shown in Table~\ref{tab:pascal_coco}.

We also combine our regularizer with the latest network design for semantic
segmentation: dilated ResNet augmented with the pyramid pooling module of
PSPNet~\cite{PSPNet}.  We used the output of the pyramid pooling layer to
predict input for the decoder and semantic segmentation.
Table~\ref{tab:pascal_imagenet} shows gain over the corresponding PSPNet
baseline.

Beyond autoencoder architecture choice, application of our regularizer involves
one free parameter: the relative weight of the auxiliary branch loss
with respect to the primary loss.  Figure~\ref{fig:weighting} shows how
performance of the trained network varies with this parameter, when using our
32-channel bottleneck layer decoder with MSE loss on the auxiliary branch. Weighting is important, but the
optimal balance appears consistent when changing architecture from VGG-16 to DenseNet-67.

We have also run similar experiments with cross-entropy loss on the auxiliary
branch with the weight parameter in $[0,6]$.  Here, the weight parameter range
is changed due to the difference in the dynamic range of values between MSE
loss and cross-entropy loss.  Behaving similarly to Figure~\ref{fig:weighting},
relative weighting of $0.5$ achieves the highest accuracy.  We use this weight
value across all of the experiments using our decoder with 128-channel
bottleneck layer.  While the regularizer always provides a benefit, placing a
proper relative weight on the auxiliary loss is important.

Figure 5.6 visualizes the impact of training with our learned
label decoder as a regularizer.  Most notably, the network trained with
regularization appears to correct some global or large-scale semantic errors
in comparison to the baseline.  Contrast such behavior to CRF-based
post-processing, which typically achieves impact through fixing local mistakes.
Also notable is that our auxiliary output itself is quite reasonable.  This
suggests that the autoencoder training phase is successful in creating encoders
and decoders that model label structure.

\subsection{Label Model Introspection}
\label{sec:introspection}

\begin{figure*}
   \setlength{\fboxsep}{0pt}
   \setlength{\fboxrule}{1pt}
   \tiny
   \begin{center}
   \begin{minipage}[t]{0.45\linewidth}
      \vspace{0pt}
      \fcolorbox{red!10}{red!10}{
      \begin{tabular}{ccc}
      \vspace{-5pt}
      {\fcolorbox{red!10}{red!10}{~}}\\
      \hspace{-4.5pt}{\fcolorbox{mediumgreen}{white}{\includegraphics[width=2.1cm]{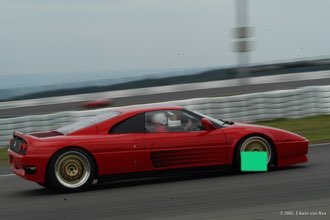}}}&\hspace{-10pt}
      {\fbox{\includegraphics[width=2.1cm]{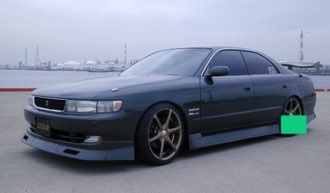}}}&\hspace{-10pt}
      {\fbox{\includegraphics[width=2.1cm]{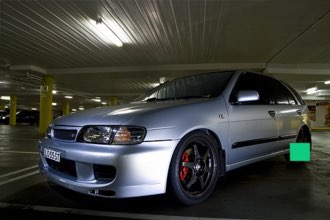}}}\\
      \vspace{3pt}
      \hspace{-4.5pt}{\fcolorbox{mediumgreen}{white}{\includegraphics[width=2.1cm]{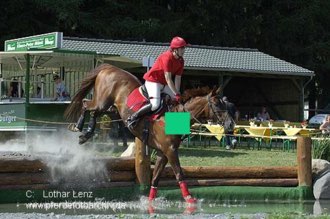}}}&\hspace{-10pt}
      {\fbox{\includegraphics[width=2.1cm]{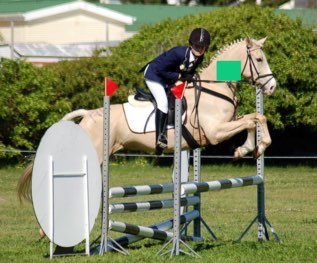}}}&\hspace{-10pt}
      {\fbox{\includegraphics[width=2.1cm]{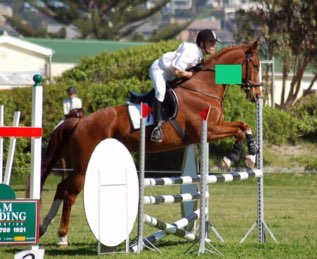}}}\\
      \vspace{3pt}
      \hspace{-4.5pt}{\fcolorbox{mediumgreen}{white}{\includegraphics[width=2.1cm]{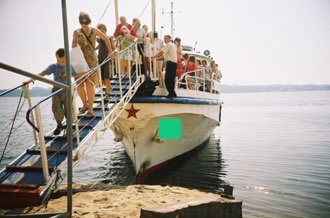}}}&\hspace{-10pt}
      {\fbox{\includegraphics[width=2.1cm]{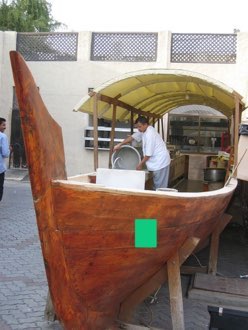}}}&\hspace{-10pt}
      {\fbox{\includegraphics[width=2.1cm]{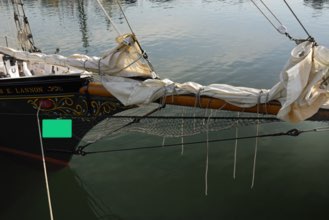}}}\\
      \vspace{3pt}
      \hspace{-4.5pt}{\fcolorbox{mediumgreen}{white}{\includegraphics[width=2.1cm]{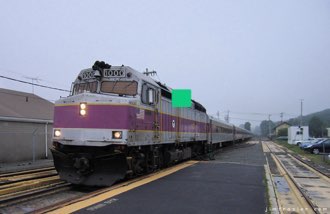}}}&\hspace{-10pt}
      {\fbox{\includegraphics[width=2.1cm]{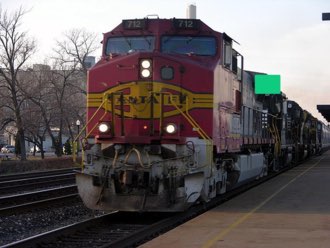}}}&\hspace{-10pt}
      {\fbox{\includegraphics[width=2.1cm]{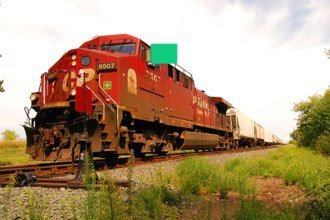}}}\\
      \vspace{3pt}
      \hspace{-4.5pt}{\fcolorbox{mediumgreen}{white}{\includegraphics[width=2.1cm]{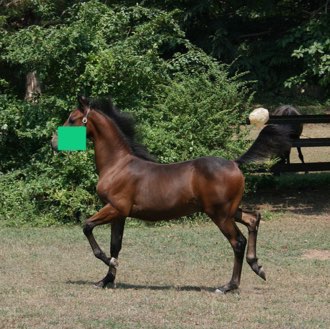}}}&\hspace{-10pt}
      {\fbox{\includegraphics[width=2.1cm]{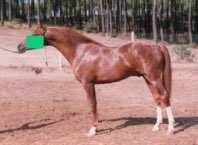}}}&\hspace{-10pt}
      {\fbox{\includegraphics[width=2.1cm]{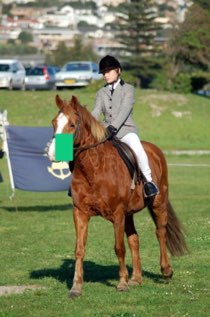}}}\\
      \vspace{3pt}
      \hspace{-4.5pt}{\fcolorbox{mediumgreen}{white}{\includegraphics[width=2.1cm]{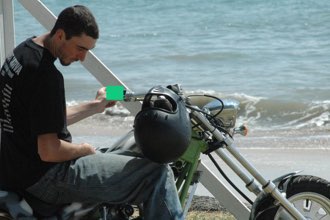}}}&\hspace{-10pt}
      {\fbox{\includegraphics[width=2.1cm]{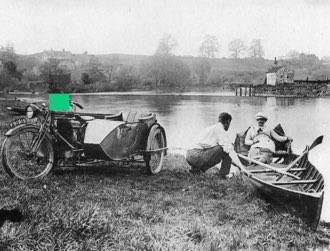}}}&\hspace{-10pt}
      {\fbox{\includegraphics[width=2.1cm]{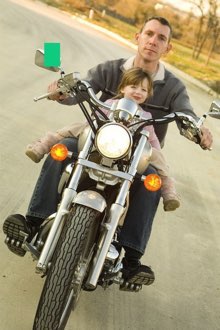}}}\\
      \vspace{3pt}
      \hspace{-4.5pt}{\fcolorbox{mediumgreen}{white}{\includegraphics[width=2.1cm]{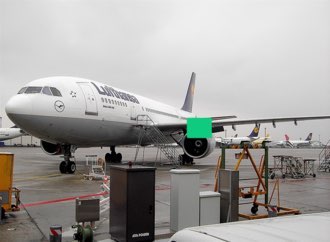}}}&\hspace{-10pt}
      {\fbox{\includegraphics[width=2.1cm]{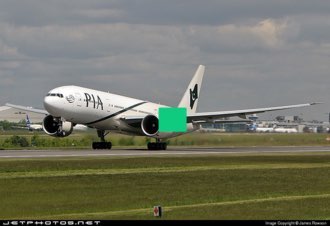}}}&\hspace{-10pt}
      {\fbox{\includegraphics[width=2.1cm]{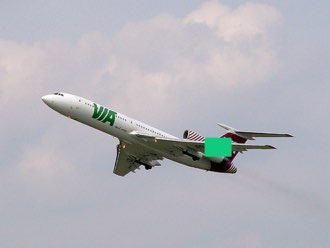}}}
      \end{tabular}}
   \end{minipage}
   \hfill
   \begin{minipage}[t]{0.5\linewidth}
      \vspace{15pt}
      \fcolorbox{gray!20}{gray!20}{%
      \begin{tabular}{ccc}
      \vspace{-5pt}
      {\fcolorbox{gray!20}{gray!20}{~}}\\
      \hspace{-4.5pt}{\fcolorbox{mediumgreen}{white}{\includegraphics[width=2.1cm]{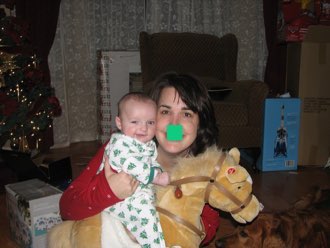}}}&\hspace{-10pt}
      {\fbox{\includegraphics[width=2.1cm]{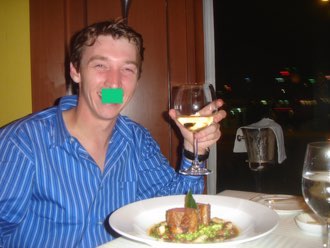}}}&\hspace{-10pt}
      {\fbox{\includegraphics[width=2.1cm]{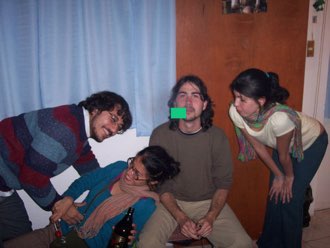}}}\\
      \vspace{3pt}
      \hspace{-4.5pt}{\fcolorbox{mediumgreen}{white}{\includegraphics[width=2.1cm]{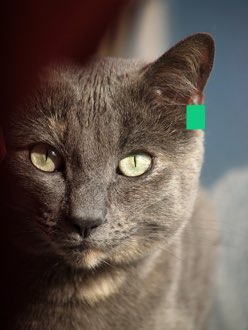}}}&\hspace{-10pt}
      {\fbox{\includegraphics[width=2.1cm]{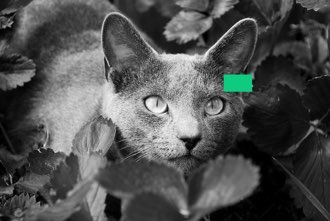}}}&\hspace{-10pt}
      {\fbox{\includegraphics[width=2.1cm]{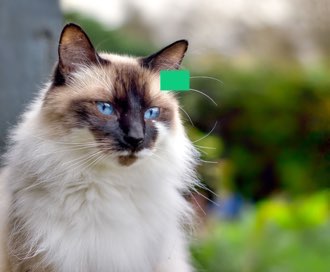}}}\\
      \vspace{3pt}
      \hspace{-4.5pt}{\fcolorbox{mediumgreen}{white}{\includegraphics[width=2.1cm]{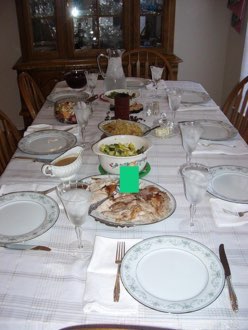}}}&\hspace{-10pt}
      {\fbox{\includegraphics[width=2.1cm]{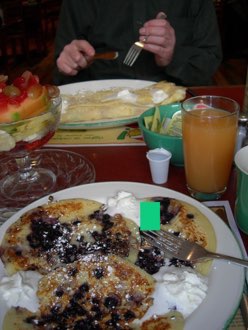}}}&\hspace{-10pt}
      {\fbox{\includegraphics[width=2.1cm]{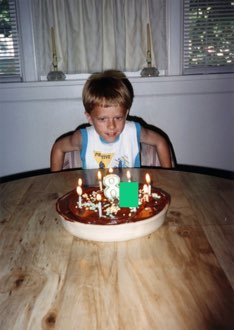}}}\\
      \vspace{3pt}
      \hspace{-4.5pt}{\fcolorbox{mediumgreen}{white}{\includegraphics[width=2.1cm]{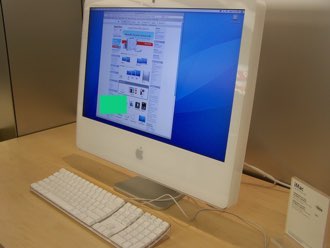}}}&\hspace{-10pt}
      {\fbox{\includegraphics[width=2.1cm]{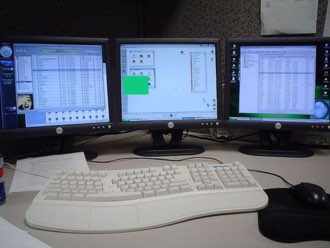}}}&\hspace{-10pt}
      {\fbox{\includegraphics[width=2.1cm]{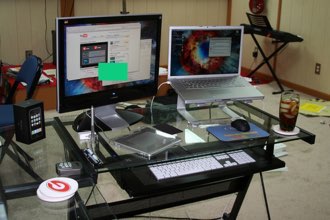}}}
      \end{tabular}}
      ~\\~\\~\\
      \fcolorbox{red!10}{red!10}{%
      \begin{tabular}{ccc}
      \vspace{-5pt}
      {\fcolorbox{red!10}{red!10}{~}}\\
      \hspace{-4.5pt}{\fcolorbox{mediumgreen}{white}{\includegraphics[width=2.1cm]{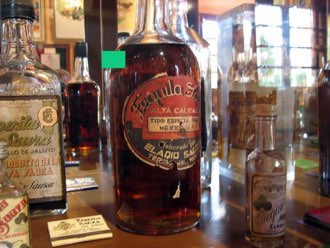}}}&\hspace{-10pt}
      {\fbox{\includegraphics[width=2.1cm]{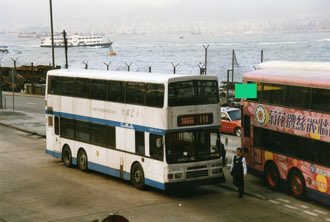}}}&\hspace{-10pt}
      {\fbox{\includegraphics[width=2.1cm]{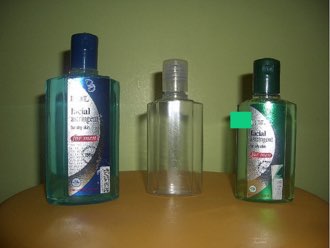}}}\\
      \vspace{3pt}
      \hspace{-4.5pt}{\fcolorbox{mediumgreen}{white}{\includegraphics[width=2.1cm]{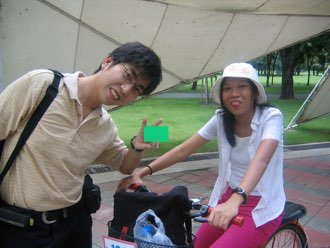}}}&\hspace{-10pt}
      {\fbox{\includegraphics[width=2.1cm]{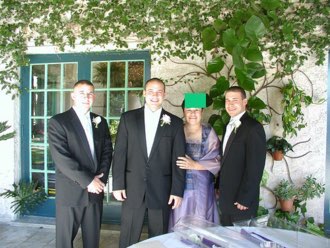}}}&\hspace{-10pt}
      {\fbox{\includegraphics[width=2.1cm]{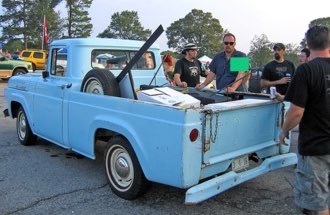}}}\\
      \vspace{3pt}
      \hspace{-4.5pt}{\fcolorbox{mediumgreen}{white}{\includegraphics[width=2.1cm]{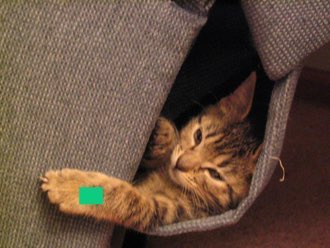}}}&\hspace{-10pt}
      {\fbox{\includegraphics[width=2.1cm]{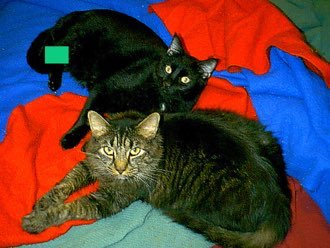}}}&\hspace{-10pt}
      {\fbox{\includegraphics[width=2.1cm]{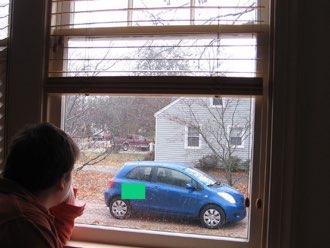}}}
      \end{tabular}}
   \end{minipage}
   \end{center}
   \vspace{-0.01\linewidth}
   \tiny
   \caption{
      Finding regions with similar representations.
      For each \textcolor{mediumgreen}{\textbf{query}} image (green border) and
      region (green rectangle), the next two images to the right are those in the
      validation set containing the nearest regions to the query region.  All
      query images are from the training set.  For examples on red background,
      search is conducted not by looking at images, but via matching features
      produced by the \textcolor{red}{\textbf{encoder}} run on ground-truth
      label maps.  The bottom-right shows failure cases, such as matching a
      cat's arm to the car rear door.  For examples on gray background, our
      \textcolor{darkgray}{\textbf{DenseNet-67 zoom-out CNN}} is used to
      predict the label space search representations from images.
   }
   \label{fig:nearest}
\end{figure*}

To further investigate what the autoencoder learns, we consider using the
bottleneck representation produced by the encoder as defining features by
which we can perform queries in label space.  Specifically, we pick a region
of a training image label and represent that region with features extracted
from bottleneck layer.  As the bottleneck layer is low-resolution, we are
selecting features at coarse, but corresponding spatial location.

Next, we perform nearest neighbor search over all regions in the validation
set and find the two closest regions to the query region.
Figure~\ref{fig:nearest} shows the results of this experiment.  Returned
regions not only have the same object class types as the query regions, but
also share similar shapes to that of the query.  This reveals that our
label autoencoder has learned to capture object shape characteristics.

We also repeat this experiment, except with queries starting from images.
Here the bottleneck representation is produced by a CNN, which was trained with
both zoom-out and decoder prediction pathways; the latter yields the
required features.  As shown in the top right of Figure~\ref{fig:nearest},
returned regions have similar context and shape to the query.

\section{Monocular depth and semantic segmentation}
\label{sec:segtodepth}

Monocular depth estimation is a crucial step towards scene geometry understanding from 2D images. The goal in monocular depth estimation is to predict the depth value of each pixel, given a single RGB image as input. In many practical applications such as autonomous driving or robotic surgery we require both semantic segmentation and depth predictions. Therefore, it is important to investigate how correlated these two tasks are. In this section first we explore how well we can predict depth maps directly from semantic segmentation maps, without using RGB images. We will show that it is possible to predict depth maps directly from semantic segmentation maps and surprisingly even with better accuracy comparing to predicting depth from RGB images. This indicates a strong correlation between these two tasks. Based on this observation, we propose an architecture in which we use segmentation as an intermediate representation for depth estimation.

Datasets like NYUv2~\cite{NYU} have orders of magnitude more depth data than annotated segmentation maps.  With the proposed architecture, we show that we are able to utilize additional depth data to gain noticeable boost in segmentation accuracy while maintaining depth prediction accuracy.

\subsection{Semantic segmentation to depth}
\begin{figure}
  \centering
\begin{tabular}{cc}
\includegraphics[width=.49\textwidth]{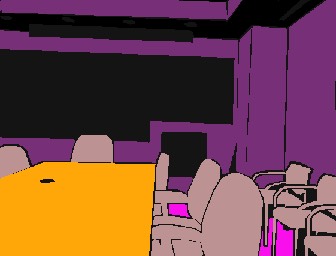} &
 \includegraphics[width=.49\textwidth]{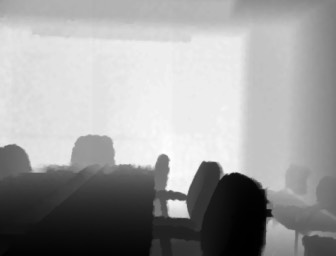}
\end{tabular}
  \caption{Left image shows semantic segmentation map and the right image is its coressponding depth map. Chairs that are occluded by table are farther from camera than the table. Also relative size of the chairs hints at their distance from the camera.}\label{fig:introseg2depthh}  
\end{figure}

Given a semantic segmentation mask we want to estimate depth values for all pixels. There are many clues such as relative objects' sizes and occlusion boundaries that can be leveraged. Figure~\ref{fig:introseg2depthh} shows an input semantic segmentation map and its corresponding depth map. 

Previous and recent work on jointly predicting semantic segmentation and depth~\cite{Wangdepth, Jiao_2018_ECCV, Mousavian3DV16, lindepth} have shown that these two tasks can benefit from joint training. In contrast, we predict depth maps from semantic segmentation maps rather than predicting them jointly from RGB images. In~\cite{Zamir_2018_CVPR} they predict depth maps from predicted segmentation maps. However, we predict depth maps from ground truth segmentation maps to evaluate how well we can predict depth maps from semantic segmentation maps. All of these methods assume that they have access to both ground truth semantic segmentation and depth for all of their training images. In contrast, we don't have such an assumption. In Section~\ref{unifiedmodel} we describe how we deal with unequal number of segmentation annotations and depth images.  

We evaluate our models on NYUv2 dataset~\cite{NYU}. NYUv2 is a widely used dataset for monocular depth estimation in indoor environments. It contains 464 indoor scenes collected with a Kinect RGBD camera. To form our training set, we sub-sample 60,000 RGB images with their corresponding depth maps uniformly from the raw part of the dataset. The labeled part of dataset consists of 1449 RGB-D images provided with semantic segmentation annotations. We follow the standard evaluation approach and use 40 categories for semantic segmentation task. 

For the following experiments, we train our models on the labeled part of the NYUv2. We follow the standard split of 795 training images and 654 test images. When we use segmentation as input, we have one binary mask channel for each object class. We use ResNet-50 with dilated convolution and apply pyramid pooling layer~\cite{PSPNet} to the features of last fully convolutional layer of ResNet. For further details regarding architecture design refer to Section~\ref{sec:zoomfurther}. Training images are resized to $256 \times 336$. For data augmentation we use random horizontal flip. We use the Adam~\cite{Adam} update
rule with learning rate of $10^{-4}$ for 60 epochs.

In order to evaluate our depth estimations we report our results based on the commonly used measures:

\begin{description}

   \item[$\bullet$] Root mean squared error (linear): $RMSE = \sqrt{\frac{1}{N} \sum_{i, j} |  y_{ij} - \hat{y}_{ij} |^2} $
   
   \item[$\bullet$] Root mean squared error (log): $RMSElog = \sqrt{\frac{1}{N} \sum_{i, j} |  log(y_{ij}) - log(\hat{y}_{ij}) |^2} $
   
   \item[$\bullet$] Squared relative difference: $srel = \frac{1}{N} \sum_{i, j} | y_{ij} - \hat{y}_{ij}|^2/\hat{y}_{ij}$
   
   \item[$\bullet$] Absolute relative difference: $rel = \frac{1}{N} \sum_{i, j} | y_{ij} - \hat{y}_{ij} |/\hat{y}_{ij}$
   
    \item[$\bullet$] Thresholded accuracy: Percentage of predicted depth $y$ such that $ max(y_i/\hat{y}_i, \hat{y}_i/y_i) < threshold $

\end{description}

Where $\hat{y}$ is the predicted depth map and $y$ is the ground truth depth map. Total number of pixels is $N$.

\begin{figure} \label{seg2depthfig}
   \setlength\fboxsep{0pt}
   \begin{center}
   \begin{minipage}[t]{0.98\linewidth}
   \begin{center}
      \begin{minipage}[t]{0.19\linewidth}
         \vspace{0pt}
         \begin{center}
         \begin{minipage}[t]{0.98\linewidth}
            \vspace{0pt}
            \begin{center}
               \fbox{\includegraphics[width=1.00\linewidth]{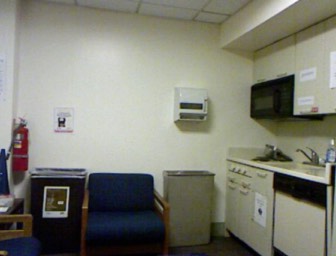}}\\
               \vspace{0.01\linewidth}
               \fbox{\includegraphics[width=1.00\linewidth]{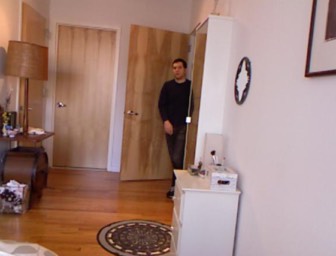}}\\
               \vspace{0.01\linewidth}
               \fbox{\includegraphics[width=1.00\linewidth]{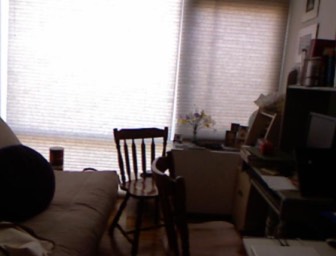}}\\
               \vspace{0.01\linewidth}
               \fbox{\includegraphics[width=1.00\linewidth]{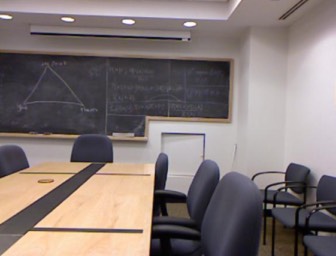}}\\
               \vspace{0.01\linewidth}
               \fbox{\includegraphics[width=1.00\linewidth]{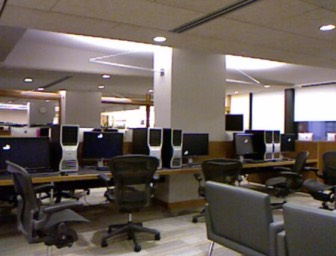}}\\
               \vspace{0.01\linewidth}
               \fbox{\includegraphics[width=1.00\linewidth]{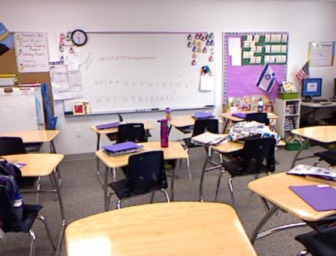}}\\
               \vspace{0.01\linewidth}
               \fbox{\includegraphics[width=1.00\linewidth]{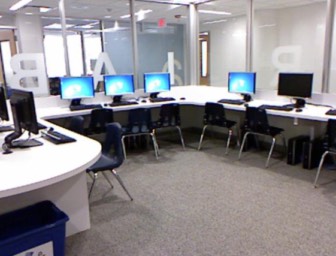}}\\
               \vspace{0.01\linewidth}
               \fbox{\includegraphics[width=1.00\linewidth]{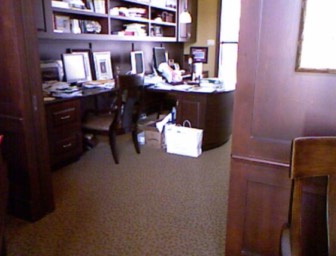}}\\
               \vspace{0.01\linewidth}
               \scriptsize{\textbf{\textsf{Image}}}
            \end{center}
         \end{minipage}
         \end{center}
      \end{minipage}
      \hfill
	  \begin{minipage}[t]{0.19\linewidth}
         \vspace{0pt}
         \begin{center}
         \begin{minipage}[t]{0.98\linewidth}
            \vspace{0pt}
            \begin{center}
               \fbox{\includegraphics[width=1.00\linewidth]{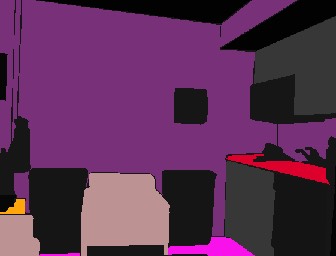}}\\
               \vspace{0.01\linewidth}
               \fbox{\includegraphics[width=1.00\linewidth]{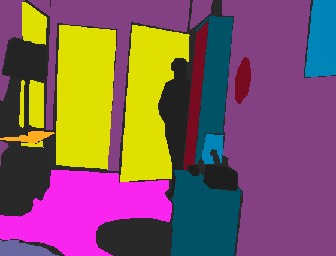}}\\
               \vspace{0.01\linewidth}
               \fbox{\includegraphics[width=1.00\linewidth]{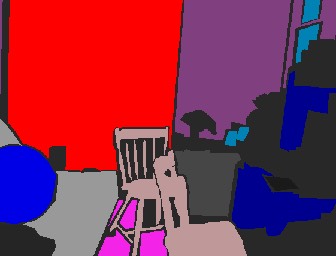}}\\
               \vspace{0.01\linewidth}
               \fbox{\includegraphics[width=1.00\linewidth]{NYU/nyugt/9_gt.png}}\\
               \vspace{0.01\linewidth}
               \fbox{\includegraphics[width=1.00\linewidth]{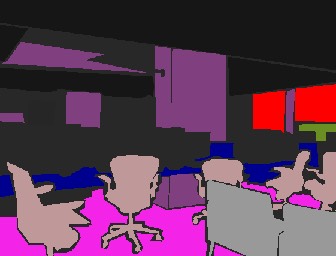}}\\
               \vspace{0.01\linewidth}
               \fbox{\includegraphics[width=1.00\linewidth]{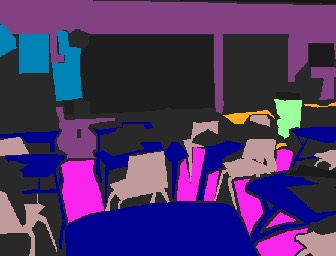}}\\
               \vspace{0.01\linewidth}
               \fbox{\includegraphics[width=1.00\linewidth]{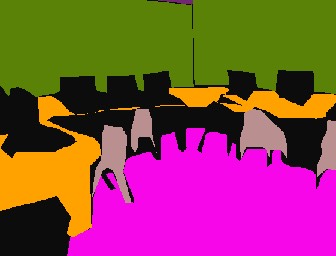}}\\
               \vspace{0.01\linewidth}
               \fbox{\includegraphics[width=1.00\linewidth]{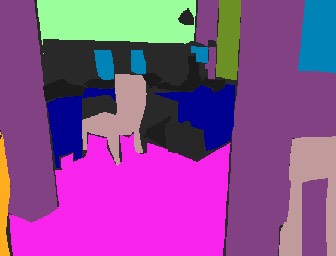}}\\
               \vspace{0.01\linewidth}
               \scriptsize{\textbf{\textsf{Ground Truth Segmentation}}}
            \end{center}
         \end{minipage}
         \end{center}
      \end{minipage}
		 \hfill
	  \begin{minipage}[t]{0.19\linewidth}
         \vspace{0pt}
         \begin{center}
         \begin{minipage}[t]{0.98\linewidth}
            \vspace{0pt}
            \begin{center}
               \fbox{\includegraphics[width=1.00\linewidth]{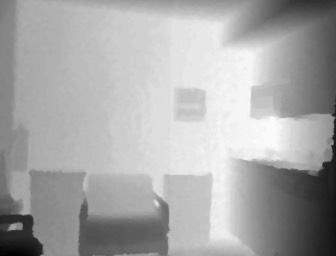}}\\
               \vspace{0.01\linewidth}
               \fbox{\includegraphics[width=1.00\linewidth]{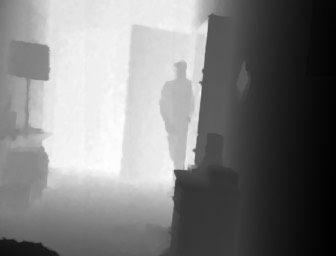}}\\
               \vspace{0.01\linewidth}
               \fbox{\includegraphics[width=1.00\linewidth]{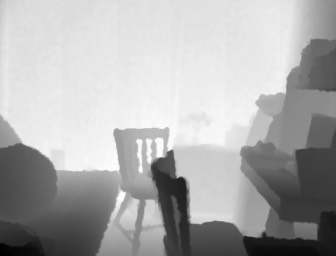}}\\
               \vspace{0.01\linewidth}
               \fbox{\includegraphics[width=1.00\linewidth]{NYU/nyudepth/9_depth.png}}\\
               \vspace{0.01\linewidth}
               \fbox{\includegraphics[width=1.00\linewidth]{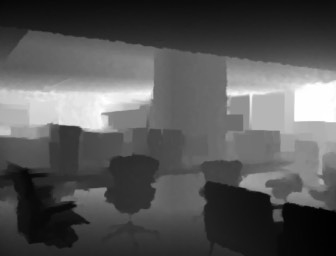}}\\
               \vspace{0.01\linewidth}
               \fbox{\includegraphics[width=1.00\linewidth]{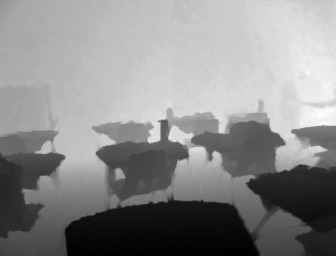}}\\
               \vspace{0.01\linewidth}
               \fbox{\includegraphics[width=1.00\linewidth]{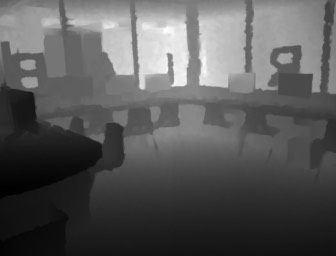}}\\
               \vspace{0.01\linewidth}
               \fbox{\includegraphics[width=1.00\linewidth]{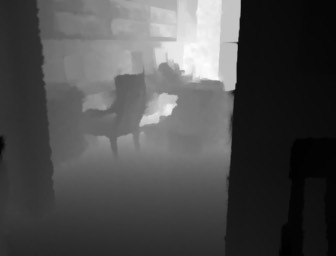}}\\
               \vspace{0.01\linewidth}
               \scriptsize{\textbf{\textsf{Ground Truth Depth}}}
            \end{center}
         \end{minipage}
         \end{center}
      \end{minipage}
      \hfill
      \begin{minipage}[t]{0.19\linewidth}
         \vspace{0pt}
         \begin{center}
         \begin{minipage}[t]{0.98\linewidth}
            \vspace{0pt}
            \begin{center}
               \fbox{\includegraphics[width=1.00\linewidth]{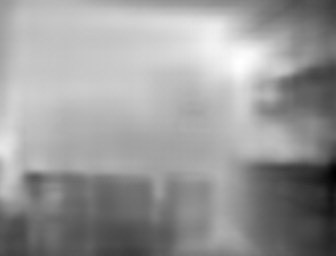}}\\
               \vspace{0.01\linewidth}
               \fbox{\includegraphics[width=1.00\linewidth]{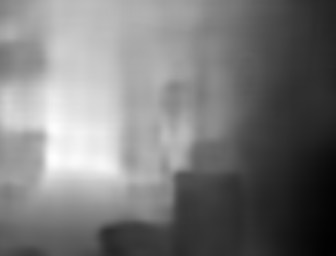}}\\
               \vspace{0.01\linewidth}
               \fbox{\includegraphics[width=1.00\linewidth]{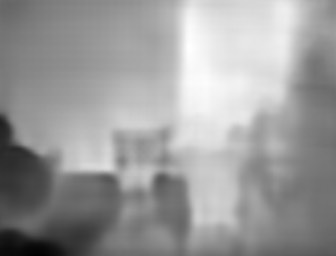}}\\
               \vspace{0.01\linewidth}
               \fbox{\includegraphics[width=1.00\linewidth]{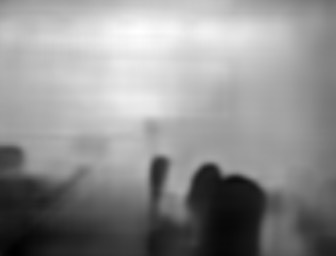}}\\
               \vspace{0.01\linewidth}
               \fbox{\includegraphics[width=1.00\linewidth]{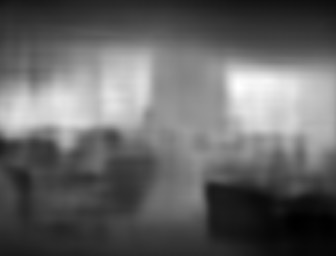}}\\
               \vspace{0.01\linewidth}
               \fbox{\includegraphics[width=1.00\linewidth]{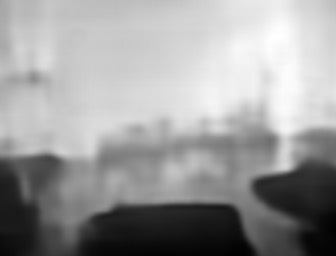}}\\
               \vspace{0.01\linewidth}
               \fbox{\includegraphics[width=1.00\linewidth]{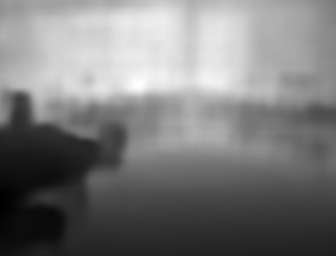}}\\
               \vspace{0.01\linewidth}
               \fbox{\includegraphics[width=1.00\linewidth]{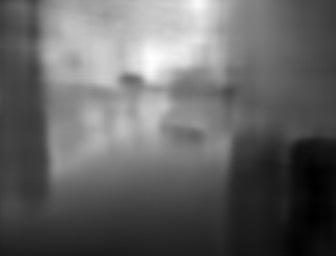}}\\
               \vspace{0.01\linewidth}
               \scriptsize{\textbf{\textsf{Image $\rightarrow$ Depth}}}
            \end{center}
         \end{minipage}
         \end{center}
      \end{minipage}
      \hfill
      	  \begin{minipage}[t]{0.19\linewidth}
         \vspace{0pt}
         \begin{center}
         \begin{minipage}[t]{0.98\linewidth}
            \vspace{0pt}
            \begin{center}
               \fbox{\includegraphics[width=1.00\linewidth]{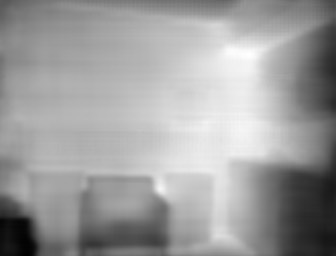}}\\
               \vspace{0.01\linewidth}
               \fbox{\includegraphics[width=1.00\linewidth]{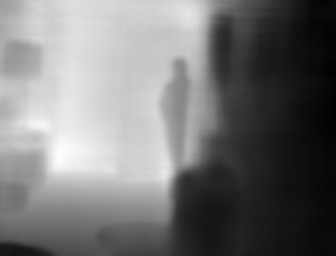}}\\
               \vspace{0.01\linewidth}
               \fbox{\includegraphics[width=1.00\linewidth]{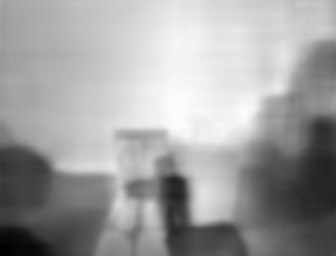}}\\
               \vspace{0.01\linewidth}
               \fbox{\includegraphics[width=1.00\linewidth]{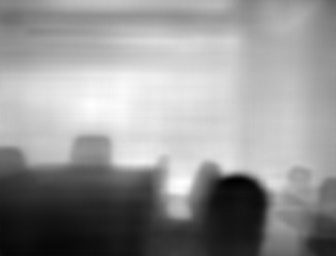}}\\
               \vspace{0.01\linewidth}
               \fbox{\includegraphics[width=1.00\linewidth]{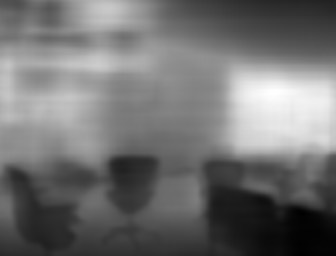}}\\
               \vspace{0.01\linewidth}
               \fbox{\includegraphics[width=1.00\linewidth]{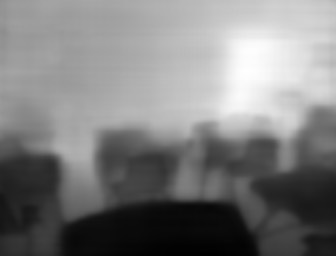}}\\
               \vspace{0.01\linewidth}
               \fbox{\includegraphics[width=1.00\linewidth]{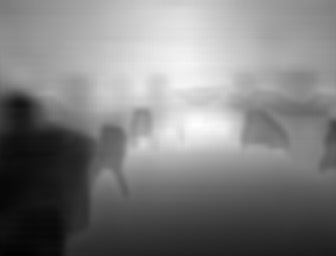}}\\
               \vspace{0.01\linewidth}
               \fbox{\includegraphics[width=1.00\linewidth]{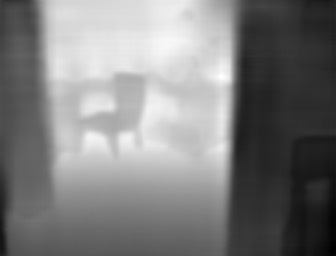}}\\
               \vspace{0.01\linewidth}
               \scriptsize{\textbf{\textsf{Segmentation $\rightarrow$ Depth}}}
            \end{center}
         \end{minipage}
         \end{center}
      \end{minipage}
   \end{center}
   \end{minipage}
   \end{center}
   \vspace{-0.01\linewidth}
   \caption{
      Depth predictions results on NYUv2. We show not only it is possible to predict depth maps from segmentation maps, but also these predicted depth maps are considerably sharper and more accurate than the predictions from RGB images.}
\end{figure}

\begin{table}[!th]
  \centering
  \footnotesize
{
\setlength{\tabcolsep}{8pt}
  \begin{tabular}{l|c|c|c|c|c|c|l}

Input&RMSE (lin.)& RMSE(log) & Abs Rel & Sqr Rel & $\delta < 1.25$ & $\delta < 1.25^2$ &  $\delta < 1.25^3$ \\
\hline
RGB & 0.628  &  0.227 & 0.1967 & 0.1749 & 0.718 & 0.918 & 0.976\\
Seg & 0.548  &  0.192 & 0.1557 & 0.1191 & 0.785& 0.953& 0.989\\
Seg+RGB & 0.524  &  0.182 & 0.1446 & 0.1052 & 0.807& 0.959& 0.991\\
\hline
  \end{tabular}}
  \caption{Depth prediction results of models with different inputs on NYUv2 test set.}
  \label{tab:segtodepth}
\end{table}

Table~\ref{tab:segtodepth} shows results of predicting depth maps from RGB and semantic segmentation inputs. Surprisingly, we get lower errors when using semantic segmentation as input. Since segmentation maps provide accurate object boundaries, depth predictions from segmentation maps in Figure 5.9 have sharper and more accurate boundaries around objects, while depth predictions from RGB images in some cases miss large part of objects. This observation indicates not only there is a strong correlation between semantic segmentation and monocular depth estimation, but also segmentation can be used as an effective intermediate representation for monocular depth estimation. The last row of Table~\ref{tab:segtodepth} shows results further improve when segmentation masks are concatenated with RGB channels and given to the input of the model. 

\subsection{Unified architecture for predicting segmentation and depth}\label{unifiedmodel}

Results in Table~\ref{tab:segtodepth} inspired us to use segmentation as an intermediate representation in our unified architecture for jointly predicting depth and segmentation. However, we used ground truth segmentation as an input to our model. Here we want to have a more practical approach and use predicted segmentation as an input. We break our depth and segmentation prediction model into two parts. The first part of the model takes an image as an input and predicts depth and segmentation. The second part of the model takes image, predicted depth and predicted segmentation as inputs and predicts depth and segmentation. In datasets like NYUv2 which offer both segmentation and depth ground truths, it is often the case that we don't have the same amount of annotations for both tasks. For example our training set has 60,000 images with depth ground truth but only 1449 images with both segmentation and depth ground truths. Given our unified model for depth and segmentation predictions the question is how do we deal with images where we do have depth maps but we don't have segmentation annotations. If we simply train the model on all training data, there are many training batches where we have no semantic segmentation supervision. This causes the model to focus less on semantic segmentation and give a higher weight to depth estimation. Ideally we want to benefit both tasks from joint training. To deal with this issue, our solution is simple: we make sure in each training batch we have at least one image for which both segmentation and depth maps are available. For the images with only depth ground truth and no segmentation annotations, we freeze segmentation head and only backpropagate through the depth head and the shared model. 

\subsubsection{Architecture details and experimental results} 

 We use zoom-out DenseNet-121 for the first part of the model. We set zoom-out spatial feature size to 1/4 of input image. For further details of zoom-out DenseNet architecture design refer to Section~\ref{sec:zoomfurther}. Once zoom-out features are computed we pass them through segmentation head and depth head. Each head consists of 3-layer CNN with $1\times1$ convolutions with 1000 hidden units in each layer. First part takes RGB image as input and predicts depth and semantic segmentation. Similarly to first part we use zoom-out DenseNet-121 for the second part. However, second part takes the output of the first part which are predicted depth and segmentation plus RGB image and predicts depth and segmentation. For each part in our unified model we have two loss functions one for depth head and one for segmentation head. In total we have 4 losses. We use L1 loss which measures mean absolute error between the predicted depth and the ground truth depth. We use cross-entropy loss for segmentation. Final loss is weighted sum of 4 loss terms. The average loss value is computed over couple of epochs for each of the loss terms. Then loss terms are weighted such that each loss term contributes equally to the final loss. We train our models for 40 epochs on both raw and labeled parts of the training set with learning rate of $10^{-4}$ using Adam update rule. Then we reduce the learning rate to $10^{-5}$ and train for 20 epochs only on the labeled part of the training set.   

Two-part architecture design allows us to predict depth and segmentation in unified model. Therefore, we utilize benefits of jointly training depth and segmentation. These benefits include a unified model with shared parameters between the two tasks and higher prediction accuracies comparing to having a separate model for each task. Furthermore, results in Table~\ref{tab:twostep} shows that significant boost in segmentation accuracy can be achieved from the same amount of segmentation annotations, but extra training images with just depth ground truth. Table~\ref{tab:twostep} shows the benefit of two-part model over the baseline. 

\paragraph{Baseline and two-part models} The difference between baseline model and two-part model is where we provide supervision. In two-part model we provide supervision for depth and segmentation predictions both at the end of the first part of the model and at the end of the second part of the model. The baseline model has the same architecture, but it has not been trained with extra supervision for the first step. Basically it is a single step model with the loss function at the very end of the network.

\paragraph{Baseline without segmentation/depth loss} Same as baseline model except we only estimate depth maps and ignore semantic segmentation task or vice versa. 

\paragraph{2x segmentation loss} In this setting we use the baseline model except segmentation cross entropy loss contributed 2 times more than depth L1 loss to the final loss. This experiment is done to make sure that the segmentation gain achieved by two-part model is due to model design and not the weight of segmentation loss with respect to the depth loss.

\begin{table}[!th]
  \centering
  \small
{
\setlength{\tabcolsep}{8pt}
  \begin{tabular}{l|c|l}

Model & Depth RMSE (lin.)& Seg mIoU \\
\hline
Baseline & 0.45  &  38\\
Baseline no seg loss & 0.49  &  -\\
Baseline no depth loss & - &  36\\
2x seg loss & 0.46 & 38\\
two-part model & 0.44  &  41\\
\hline
  \end{tabular}}
  \caption{Depth and semantic segmentation results of baseline and our two-part models based on zoom-out DenseNet-121 on NYUv2 test set.}
  \label{tab:twostep}
\end{table}

 Results in Table~\ref{tab:twostep} show: a) both segmentation and depth predictions can benefit from joint training b) segmentation can gain significant boost from extra depth ground truth data.

\section{Summary}
\label{sec:conclusion}

Our novel label embedding-based regularization method, when applied to training deep networks for
semantic segmentation, consistently improves their generalization performance.
The intuition behind our work, that additional supervisory signal can be
squeezed from highly detailed annotation, is supported by the types of errors
this regularizer corrects, as well as our efforts at introspection into our
learned label model.

We showed that it is possible to predict depth maps directly from semantic segmentation
maps and proposed a two-part architecture for joint prediction of depth and semantic segmentation which uses
semantic segmentation as an intermediate representation for depth estimation.
Our two-part architecture improves results over the baselines for both depth and semantic segmentation predictions.

\chapter{Conclusion}\label{sec:finalconclusions}

In this thesis, we presented zoom-out architecture for learning and extracting rich representations for structured visual prediction tasks. Our effective and simple architecture can turn any modern CNN backbone such as such as DenseNet, ResNet and NASNet into near state-of-the-art image labeling model, without the need to use complicated methods both in network design and training methodology.

We proposed a modified zoom-out architecture tailored for
classification tasks, to drive a weakly supervised process that automatically labels a sparse,
diverse training set of points likely to belong to classes of interest. We obtained competitive results on the VOC 2012 segmentation benchmark by training our model on the set of automatically generated sparse labels.

Our novel label embedded-based regularization method improves generalization performance of image labeling networks. Our results indicate that one should now reevaluate the relative utility
of different forms of annotation; our method makes detailed labeling more
useful than previously believed.  This observation may be especially important
for applications of computer vision, such as self-driving cars, that demand
detailed scene understanding, and for which large-scale dataset construction is
essential. Since they have access to large-scale detailed ground-truth labels, they can utilize our label embedding-based regularization technique to further improve the generalization performance of their deep models.

The success of our two-part model for joint training of segmentation and depth estimation suggests that when there is a strong correlation between two tasks, using the output of one task as an intermediate representation for the other task can lead two an improved performance in both tasks. A future work in this direction would be to extend this approach for more than two tasks, and have a unified architecture for multi-task learning.

\bibliography{segmentation}

\end{document}